\documentclass[10pt,twocolumn,letterpaper]{article}
\usepackage{iccv}
\usepackage{times}
\usepackage{epsfig}
\usepackage{graphicx}
\usepackage{amsmath}
\usepackage{amssymb}
\usepackage{times}
\usepackage{longtable}
\usepackage{amsmath}
\usepackage{amssymb}
\usepackage{algorithm}
\usepackage{algorithmic}
\usepackage{bbm}
\usepackage{array}
\usepackage{sidecap}

\usepackage[pagebackref=false,breaklinks=true,letterpaper=true,allcolors=blue,colorlinks,bookmarks=false]{hyperref}

\def\etal{et al.\ }
\def\eg{e.g.\ }
\def\etc{etc.\ }
\def\ie{i.e.\ }
\def\vs{vs.\ }
\def\cf{cf.\ }

\def\grad{\nabla}

\def\Lt{\tilde{L}}
\def\R{\mathbb{R}}
\def\X{\mathcal{X}}
\def\I{\mathcal{I}}
\def\F{\mathcal{F}}

\def\x{\textbf{x}}
\def\k{\textbf{k}}

\def\y{\textbf{y}}
\def\l{\boldsymbol{\ell}}

\iccvfinalcopy 

\def\iccvPaperID{2208}

\ificcvfinal\fi
\begin{document}

\title{
Scalable Nonlinear Embeddings for Semantic Category-based Image Retrieval
}
\author{Gaurav Sharma and Bernt Schiele \vspace{0.2em}\\
Max Planck Institute for Informatics, Germany\\
{\tt\small lastname@mpi-inf.mpg.de}
}

\maketitle
\thispagestyle{empty}


\begin{abstract}
We propose a novel algorithm for the task of supervised discriminative distance learning by
nonlinearly embedding vectors into a low dimensional Euclidean space. We work in the challenging
setting where supervision is with constraints on similar and dissimilar pairs while training.  The
proposed method is derived by an approximate kernelization of a linear Mahalanobis-like distance
metric learning algorithm and can also be seen as a kernel neural network. The number of model
parameters and test time evaluation complexity of the proposed method are $O(dD)$ where $D$ is the
dimensionality of the input features and $d$ is the dimension of the projection space--this is in
contrast to the usual kernelization methods as, unlike them, the complexity does not scale linearly
with the number of training examples. We propose a stochastic gradient based learning algorithm
which makes the method scalable (\wrt the number of training examples), while being nonlinear. We
train the method with up to half a million training pairs of 4096 dimensional CNN features. We give
empirical comparisons with relevant baselines on seven challenging datasets for the task of low
dimensional semantic category based image retrieval.
\end{abstract}

\section{Introduction}
\label{sec:intro}

Learning distance metrics for comparing multi-dimensional vectors is a fundamental problem. If a
perfect task adaptive distance metric is available, many important computer vision problems such as
image (object, scene etc.) classification and retrieval become trivial using nearest neighbor
search \cite{GoldbergerNIPS2004, WeinbergerJMLR2009}. Metric learning is also 
applicable to many other important computer vision tasks requiring vector comparisons
\cite{BedagkarIVC2014,GuillauminICCV2009,LiCVPR2012,LuPAMI2014,MignonCVPR2012,MensinkECCV2012}.

In the present paper, we are interested in learning a distance metric by embedding the vectors into
a low dimensional Euclidean space. We work with pairwise constraints of the form
$\{(\x_i,\x_j,y_{ij})\}$ where $\x_i,\x_j \in \R^D$ are the feature vectors and $y_{ij}= +1$ if they
are semantically similar and should have a small distance, and $y_{ij}=-1$ otherwise.
This is a practical setting as obtaining such side information is easier, \eg by getting user
feedback, than annotating all the vectors for their classes. Moreover, in such scenarios, as no
class specific models are learned, the distances and embeddings learned are generic.

\begin{figure}[t]
\centering
\includegraphics[width=\columnwidth, trim=0 385 360 0, clip]{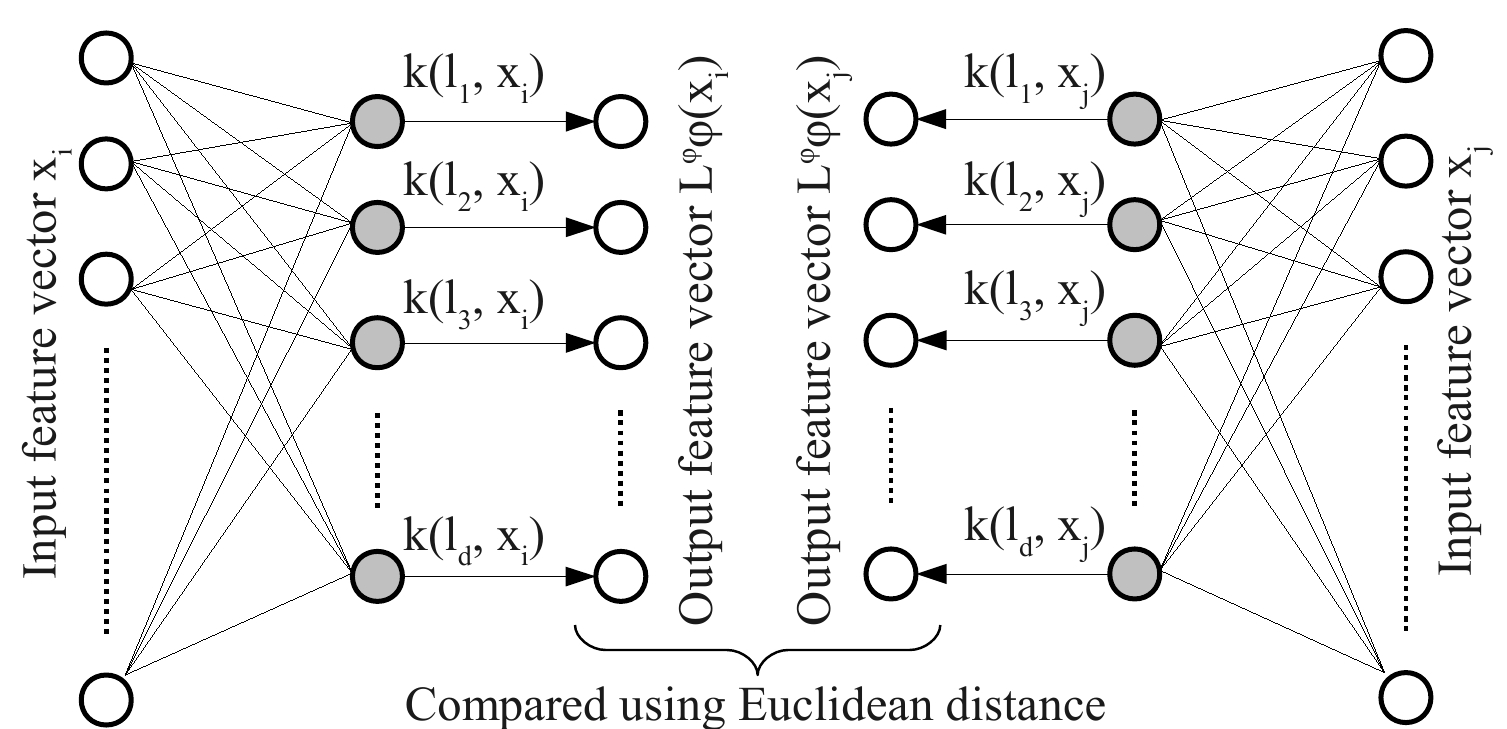} 
\caption{
The proposed method seen as a kernel neural network with one hidden layer (shaded). The input
features, for the test pair, are passed through the network respectively and then compared using
Euclidean distance.
}
\vspace{-1.2em}
\label{fig:illus}
\end{figure}

Metric learning has attracted substantial attention in the machine learning community
\cite{DavisICML2007, GlobersonNIPS2005, GoldbergerNIPS2004, WeinbergerJMLR2009} and has
specifically achieved much success in computer vision \eg for image auto-annotation  
\cite{GuillauminCVPR2009}, face verification \cite{MignonCVPR2012}, visual tracking
\cite{LiCVPR2012}, person reidentification \cite{BedagkarIVC2014} and nearest neighbor based image
classification \cite{MensinkECCV2012}. While initial work on metric learning was mostly for learning
a linear Mahalanobis-like distance, nonlinear metric learning has also been explored
\cite{DavisICML2007, HoiCVPR2006, MignonCVPR2012, SchultzNIPS2004}. Linear methods have been
extended to be nonlinear \eg by \emph{kernelization} \cite{DavisICML2007, HoiCVPR2006,
MignonCVPR2012} (we discuss in detail in \S\ref{sec:relwork}) and other nonlinear learning methods
have been proposed  \cite{ChopraCVPR2005, KedemNIPS2012, NorouziNIPS2012}. However, such
nonlinear methods have not been demonstrated to be scalable (to order of millions of training
points in spaces of order of thousands of dimensions). Similar to kernel support vector machines
(SVM), the complexity (number of model parameters and evaluation time) of kernelized
versions is often linear in the number of training examples\footnote{More precisely, for kernel SVMs, the
complexity is linear in the $N_{sv}$ (number of support vectors) whose expectation is bounded below
by $(\ell-1)E(p)$ where $E(p)$ is the expectation of the probability of error on a test vector and
$\ell$ is the number of training examples \cite{BrugesICML1996, Vapnik2000}. Hence, the complexity
can be expected to scale approximately linearly with the number of training examples.}. In view of
such undesirable scaling, we make the following contributions in this paper. (i)
Inspired by recent work on efficient nonlinear SVMs \cite{SharmaBMVC2013, SharmaHAL2014}, we propose
a novel metric learning method using kernels. The proposed method can also be seen as a kernel
neural network with one hidden layer (Fig.~\ref{fig:illus}), trained to optimize the verification
objective \ie bringing similar pairs close and pushing dissimilar ones far. We propose to use an
efficient stochastic gradient descent (SGD) algorithm for training the system. Use of SGD combined
with the fact that the complexity (number of model parameters and evaluation time) of the proposed
method does not depend on the number of training examples makes the method scalable \wrt the number
of training examples. While in the present paper we work with nonlinearity based on the popular
$\chi^2$ kernel \cite{VedaldiPAMI2012}, the method is generalizable to kernels whose derivatives can
be computed analytically. (ii) We show consistent improvement obtained by the method for the task of
semantic category based retrieval with seven challenging publicly available image datasets of
materials, birds, human attributes, scenes, flowers, objects and butterflies. Our experimental
results support the datasets. (iii) We also demonstrate scalability by training with order of millions of
training pairs of 4096 dimensional state-of-the-art CNN features
\cite{KrizhevskyNIPS2012,Vedaldi2014} and compare with five existing competitive baselines.

\section{Related work and background} 
\label{sec:relwork}

Metric learning has been an active topic of research (we encourage the interested reader to see
\cite{BelletArxiv2013, KulisFTML2012} for extensive surveys) with applications to face verification
\cite{LFWTech}, person reidentification \cite{BedagkarIVC2014}, image auto-annotation  
\cite{GuillauminCVPR2009}, visual tracking \cite{LiCVPR2012}, nearest neighbor based image
classification \cite{MensinkECCV2012} \etc in computer vision.
Starting from the seminal paper of Xing \etal \cite{XingNIPS2002}, many different approaches for
learning metrics have been proposed \eg \cite{DavisICML2007, GlobersonNIPS2005, GoldbergerNIPS2004,
HoiCVPR2006, MignonCVPR2012, SchultzNIPS2004, ShalevICML2004, TorresaniNIPS2006, TsangICANN2003,
WeinbergerNIPS2006}.

Different types of supervision has been used for learning metrics. While some methods require class
level supervision \cite{SalakhutdinovAISTATS2007}, others only require triplet
constraints, \ie $\{(\x_i,\x_j,\x_k)\}$, where $\x_i$ should be closer to $\x_j$ than to $\x_k$
\cite{WeinbergerNIPS2006}, and others still, only pairwise constraints, \ie $\{(\x_i,\x_j,y_{ij})\}$
where $y_{ij}=+1$ if $(\x_i,\x_j)$ are similar and $y_{ij}=-1$ if they are dissimilar
\cite{MignonCVPR2012}.

Most of the initial metric learning methods were linear, \eg the semidefinite programming formulation by
Xing \etal \cite{XingNIPS2002}, large margin formulation for $k$-NN classification by Weinberger
\etal \cite{WeinbergerNIPS2006}, `collapsing classes' formulation (make the distance between vectors
of same class zero and between those of different classes large) of Globerson and Roweis
\cite{GlobersonNIPS2005} and neighbourhood component analysis of Goldberger \etal
\cite{GoldbergerNIPS2004}. 

Towards scalability of metric learning methods, Jain \etal \cite{JainNIPS2008} proposed online
metric learning and more recently Simonyan \etal \cite{SimonyanBMVC2013}
proposed to use stochastic gradient descent for face verification problem.

Works also reported learning nonlinear metrics. Many linear metric learning approaches were shown to
be kernelizable \cite{DavisICML2007, HoiCVPR2006, MignonCVPR2012, SchultzNIPS2004,
ShalevICML2004, TorresaniNIPS2006, TsangICANN2003}. Tsang and Kwok \cite{TsangICANN2003} proposed a
metric learning problem similar to the $\nu$-SVM \cite{ScholkopfNC2000}, which they solved in the
dual allowing the use of kernels. Schultz and Joachims \cite{SchultzNIPS2004} proposed support
vector machine based algorithm which was accordingly kernelized while Chatpatanasiri \etal
\cite{ChatpatanasiriNC2010} proposed to use kernel PCA for nonlinear metric learning. Mignon and Jurie
\cite{MignonCVPR2012} proposed to use kernels for metric learning with sparse pairwise constraints
with a logistic loss based objective function--we are interested in a similar weakly supervised
setting and we discuss this in more detail in the next section (\S\ref{sec:background}). 

Other than nonlinearity via kernelization, alternate nonlinear forms of metrics have also been studied
\eg based on variations and adaptations of neural networks \cite{ChopraCVPR2005,
SalakhutdinovAISTATS2007}, boosting \cite{KedemNIPS2012}
binary codes with hamming distances \cite{NorouziNIPS2012}. Weinberger \etal
\cite{WeinbergerJMLR2009} also proposed to use local metric learning for introducing nonlinearity in
the learnt metric.

\subsection{Background: Supervised discriminative metric learning with pairwise constraints}
\label{sec:background}

Given a dataset $\X$ of positive and negative pairs of vectors  \ie 
$\X = \{(\x_i,\x_j, y_{ij})| (i,j) \in \I \subset \mathbb{N}^2 \}$,
with $\x_i\in\mathbb{R}^D$ and $y_{ij}\in\{-1,+1\}$ and $\I$ being an index set, the task is to learn a
distance function in $\mathbb{R}^D$. 

Many metric learning approaches learn, from $\X$, a Mahalanobis-like metric parametrized by matrix
$M\in\mathbb{R}^{D\times D}$, \ie 
\begin{equation} 
d^2(\x_i,\x_j) = (\x_i-\x_j)^{\top} M (\x_i-\x_j).
\end{equation}
$M$ is required to be symmetric positive semi-definite (PSD) matrix, for the distance to be a valid
metric, and hence can be factorized as
$M=\Lt^{\top}\Lt$,
with $\Lt\in\mathbb{R}^{d\times D}$ and $d\leq D$. The metric learning can then be seen as learning
a projection upon which the comparison is done using Euclidean distance in the resulting
space \ie
\begin{align}
\label{eqn:dist}
d^2(\x_i,\x_j) 
               & = \| \Lt\x_i - \Lt\x_j \|^2_2.
\end{align}

Learning such distance function has been achieved by using, among other methods, optimization 
of probabilistic objectives (based on likelihood) or loss functions based
on the max-margin principle \eg for the task of face verification in computer vision
\cite{GuillauminCVPR2009, MignonCVPR2012, SimonyanBMVC2013}. Here, we minimize the objective with
hinge loss, \ie 
\begin{equation}
\label{eqn:epmlopti}
\min_{\Lt,b} F = \sum_{\X} \max \left(0, 1 - y_{ij} \{b - d^2(\x_i, \x_j)\} \right),
\end{equation}
which aims to learn $\Lt$ such that the positive pairs are at distances less than $b-1$, to each
other, while the negatives are at distances greater than $b+1$, with $b$ being the bias parameter
\ie a threshold on the distance between two vectors to decide if they are same or not. Explicit
regularization is absent as often $d \ll D$ \ie the rank $d$ of the learnt metric $M=\Lt^{\top}\Lt$
is fixed to be small.

While the distance function learned as above is linear, the problem may be complex and require
nonlinear distance function. A popular way of learning a nonlinear distance function is by
kernelizing the metric, as inspired by the traditional kernel based methods \eg KPCA and KLDA;
invoke representer theorem like condition and write the rows of $\Lt$ as linear combinations of the
input vectors \ie $\Lt=AX^T$ (where $X$ is the matrix of all vectors $\x_i$ as columns). Noticing
that the distance function in Eq.~\ref{eqn:dist} depends only on the dot products of the vectors,
allows nonlinearizing the algorithm as follows. Mapping the vectors with a non-linear \emph{feature
map} $\phi:\mathbb{R}^D \rightarrow \F$ and then using the \emph{kernel trick} \ie $k(\x_i,\x_j) =
\langle \phi(\x_i), \phi(\x_j)\rangle_\F$, we can proceed as follows,
\begin{equation}
d^2(\x_i,\x_j) = \|AX_\phi^T\phi(\x_i) - AX_\phi^T\phi(\x_j)\| = \| A (\k_i - \k_j) \|^2,
\end{equation}
where, 
$X_\phi = [\phi(\x_1), \phi(\x_2),\ldots,\phi(\x_n)]$
is the matrix of $\phi$ mapped vectors and 
\begin{equation}
\k_t = X_\phi^T \x_t = [k(\x_1, \x_t), k(\x_2, \x_t), \ldots, k(\x_n, \x_t)]^T
\end{equation} 
is the t$^{th}$ column of the kernel matrix. Such reasoning was used by Mignon and Jurie
\cite{MignonCVPR2012} recently. While this is a successful way of
nonlinearizing the algorithm, it is costly and not scalable as training requires the whole kernel
matrix. 

\section{Proposed method}
\label{sec:nonlinml}
We now give the details of the proposed nonlinear embeddings by approximate kernelization of
Mahalanobis-like distance metric learning.

Continuing from the discussion in the previous section \ref{sec:background}, we note that the rows of
$\Lt$ can be thought of as a basis set (albeit not necessarily orthogonal) on which the test vectors are
projected by taking dot products. After projection the comparison is simply done using the
Euclidean distance. Now consider again a nonlinear feature map $\phi:\mathbb{R}^D \rightarrow \F$,
and the corresponding kernel function $k(\x_i,\x_j) = \langle \phi(\x_i), \phi(\x_j) \rangle_\F$; we
can view the projection, with matrix $L^\phi$ in the $\phi$ mapped space $\F$, as 
\begin{equation}
L^\phi \phi(\x) = \left[\langle \l^\phi_1, \phi(\x) \rangle_\F, 
                                           \ldots, 
                                           \langle \l^\phi_d, \phi(\x) \rangle_\F \right],
\end{equation}
where $\l^\phi_t \in \F$ are the rows of $L^\phi$.
Instead of writing each $\l^\phi$ as linear combinations of $\{ \phi(\x_i) \}$, as done
traditionally, we take an alternate route and make an approximating assumption as follows. 
We recall the concept of pre-image, in $\R^D$, of a vector in $\F$ \ie $\x\in\R^D$ corresponding to a
feature space vector $\x^\phi \in \F$ such that $\phi(\x)=\x^\phi$, which has been studied in the
past in the context of different kernel methods \cite{BrugesICML1996,KwokNN2004}. We assume that
there exists $\l_t\in\R^D$ such that either $\phi(\l_t) = \l^\phi_t$ \ie $\l_t \in \R^D$ is the
pre-image of $\l^\phi_t \in \F$ or, if such pre-image doesn't exist, then $\phi(\l_t) \approx
\l^\phi_t$ \ie $\l_t \in \R^D$ is an approximation for the pre-image of $\phi(\l_t)$. 

Once we have $\phi(\l_t) = \l^\phi_t$, we can then write 
\begin{align}
L^\phi \phi(\x) & = \left[\langle \phi(\l_1), \phi(\x) \rangle_\F,  \nonumber
                                             \ldots, 
                                             \langle \phi(\l_d), \phi(\x) \rangle_\F \right] \\
                & = \left[k(\l_1,\x), k(\l_2,\x),\ldots, k(\l_d,\x) \right].
\end{align}
Intuitively, what we did here is that instead of a linear projection of the test vector on the rows
of $\Lt$ as in the linear case (\S\ref{sec:background} above), we now do a `nonlinear projection' on the
rows of  $L = [\l_1^T, \l_2^T, \ldots, \l_d^T]^T$. This can also be seen in the
similar spirit as dimensionality reduction using nonlinear kernel methods \eg in kernel PCA the
principal components come out to be linear combinations of
$\phi$ mapped input vectors \ie $\sum_i \gamma_i \phi(\x_i)$, and the test vectors are
projected nonlinearly to these principal components. Similarly, our method can be seen as a
way of doing nonlinear dimensionality reduction with, as we will detail in the following,
discriminative supervised learning using similar and dissimilar pairs annotations.

\begin{algorithm}[t]
\begin{algorithmic}[1]
\STATE \emph{Given}: Training set ($\X$), margin ($m$), learning rate ($r$) 
\STATE \emph{Initialize}: $b=1$, $L \leftarrow$ random$(-0.5,0.5)$
\FORALL{$i = 1,\ldots,$\texttt{niters} }
    \STATE Randomly sample a training pair $(\x_i,\x_j, y_{ij})\in \X$ 
    \STATE Compute $d_\phi^2(\x_i,\x_j)$ using Eq.~\ref{eqn:nonlindist}
    \IF{$y_{ij}(b - d_\phi^2(\x_i,\x_j)) < m$}
        \FORALL{$t = 1,\ldots,d$}
            \STATE // ref.\ Eq.~\ref{eqn:grad}
            \STATE $\l_t \leftarrow \l_t - r y_{ij} (k^t_i - k^t_j) (\grad_{\l_t}k^t_i
            - \grad_{\l_t}k^t_j) $
        \ENDFOR
    \ENDIF
\ENDFOR
\caption{SGD for proposed nonlinear metric learning}
\label{algo:ml}
\end{algorithmic}
\end{algorithm}

Following our assumption, the distance function computed in the feature space becomes,
\begin{align}
d_\phi^2 (\x_i, \x_j)  & = \| L^\phi\phi(\x_i) - L^\phi\phi(\x_j) \|^2 \nonumber \\
\label{eqn:nonlindist} & = \sum_{t=1}^d (k(\l_t,\x_i) - k(\l_t, \x_j))^2.
\end{align}
With this formulation, the parameters to be learned are the elements of the matrix $L$ \ie $\l_t \in
\R^D \forall t=1,\ldots,d$. Note that this matrix is different from the $\Lt$ matrix in the linear
distance function case above in \S\ref{sec:background}. We propose to learn $L$ with a standard max
margin hinge loss based objective. The optimization problem thus takes the form,
\begin{equation}
\label{eqn:opti}
\min_{L} F = \sum_{\X} \max \left(0, m - y_{ij} \{b - d_\phi^2(\x_i, \x_j)\} \right),
\end{equation}
where the fixed margin of $1$ is replaced by a free parameter $m$. This is important as depending on
the kernel used the distances may be bounded from above \eg with histogram based kernels commonly
used in computer vision, say $\chi^2$ kernel, the maximum distance between two vectors (histograms)
is bounded by zero from below and unity from above. Hence, clearly the bias $b$ (recall that this is
like a threshold on the distances, to decide if a pair is same or not) has to be less than unity and
consequently the margin has to be less than unity as well. As we will show later in experiments, the
method is not very sensitive to these parameters and we fixed it once for different datasets.

Finally, we substitute for the distance function in feature space using Eq.~\ref{eqn:nonlindist}
and propose to optimize the objective efficiently using SGD.
At each iteration we sample an annotated training pair and update the $L$ matrix based on the
sub-gradients of the objective function. While the objective function is nonlinear
we find that the optima obtained with our SGD implementation perform consistently and well in
practice, without per instance/database tuning.

Alg.~\ref{algo:ml} gives the pseudo code of the procedure used for generating the experimental
results in this paper. The sub-gradients required for the SGD iterations in the algorithm are
calculable analytically and are as follows, 
\begin{align}
\label{eqn:grad}
&\grad_{\l_t} F(\{(\x_i,\x_j,y_{ij})\};L,b) \nonumber \\
&    = \left\{     
        \begin{array}{l}
            0  \text{ \; if \; } y_{ij} \{b - d_\phi^2(\x_i, \x_j)\} \geq m \\
            2 y_{ij} (k^t_i - k^t_j) (\grad_{\l_t}k^t_i - \grad_{\l_t}k^t_j)  \; \text{otherwise,}\\
        \end{array} 
      \right.
\end{align}
where $k^t_i = k(\l_t, \x_i)$ and $\grad_b F(\{(\x_i,\x_j,y_{ij})\};L,b)  = 0$ if $y_{ij} \{b
- d_\phi^2(\x_i, \x_j)\} \geq 1$ and $-y_{ij}$, otherwise.  As a last detail, we use the shifted
$\chi^2$ kernel, shown to be quite effective for image classification \etc \cite{VedaldiPAMI2012},
given by 
\begin{equation}
k(\x,\y) = \sum_{c=1}^D \frac{2x_c y_c}{|x_c|+|y_c|}.
\end{equation}
The gradient for the $\chi^2$ kernel is also analytically calculable and is given by,
\begin{equation}
\nabla_{\l} k (\l,\x) = \frac{2x_d|x_d|}{(|x_d| + |\ell_d|)^2}.
\end{equation}

The complexity of the stochastic updates is $O(dD)$\footnote{While there are kernels whose
computation complexity is not linear in $D$ \eg kernel based on the earth movers distance (EMD)
\cite{RubnerIJCV2000},  we work here only with those with linear complexity.} (kernel evaluations
with $\l_t \forall t=1,\ldots,d$), while the complexity of similar updates with the traditional
parametrization will be $O(ND)$ where $N$ is the number of distinct training vectors, as the $\l_t$
are linear combinations of $\phi(\x_i)$. Since $d$ is fixed and small ($d \ll N$) the updates are
relatively inexpensive and the algorithm is, thus, scalable to order of millions of training points. 
Also, since the learning is using a SGD based algorithm which takes the training pairs sequentially,
it can also be applied in an online setting where the training examples are only available with time.

Another important salient feature of the algorithm is that it can directly work with high
dimensional vectors and does not require them to be compressed (preprocessed) first with unsupervised
methods, \eg PCA \cite{WeinbergerJMLR2009} or KPCA \cite{ChatpatanasiriNC2010}. This potentially
allows the method to exploit the full representative power of the high dimensional space of the
vector for the current discrimination task, which might be otherwise lost in the case of
unsupervised dimensionality reduction as a preprocessing.

\section{Experimental results}
\label{sec:exp}
\begin{figure}
\centering
\includegraphics[width=0.51\linewidth,trim= 0 0 20 25,clip]{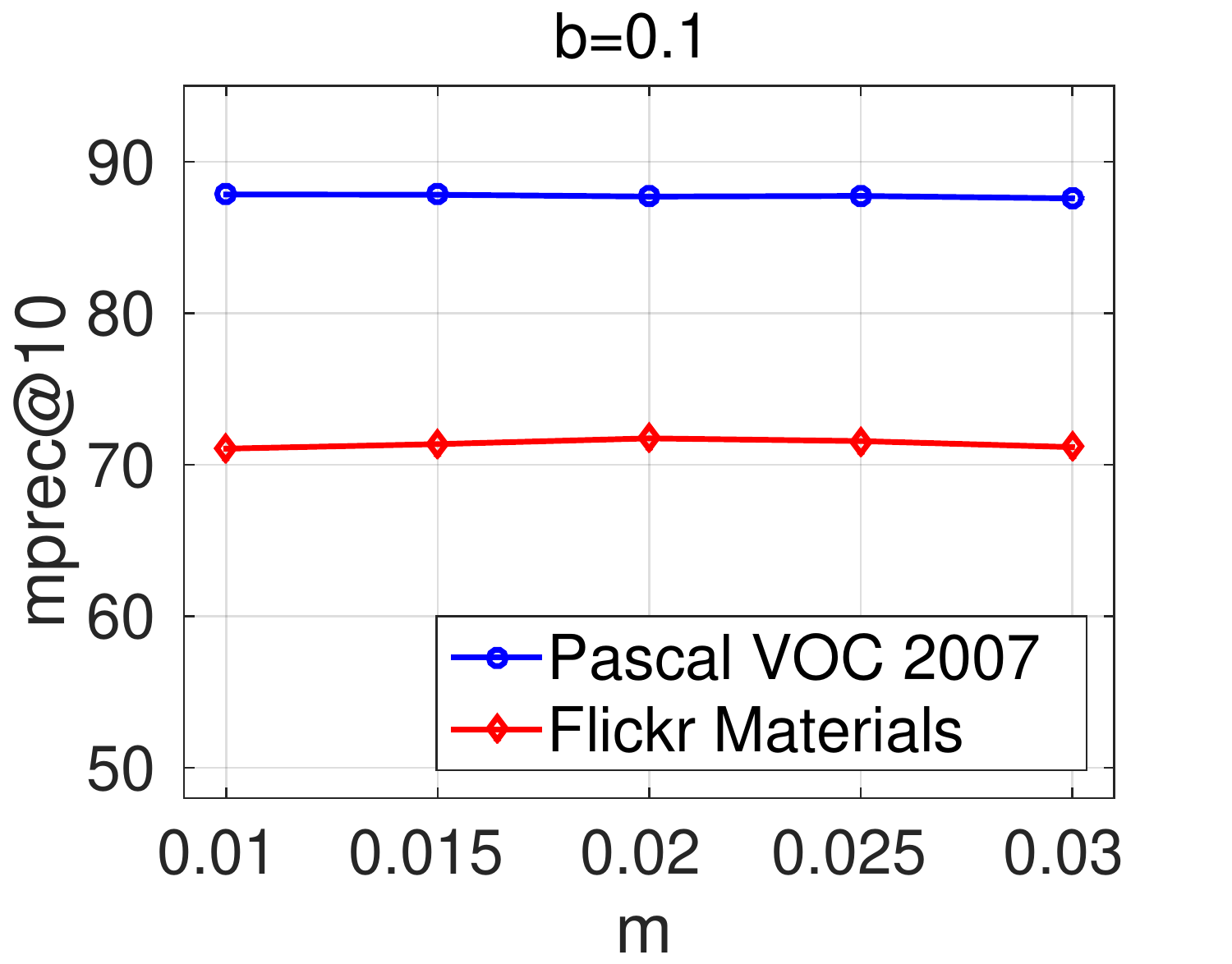} 
\includegraphics[width=0.47\linewidth,trim=56 0 0 25,clip]{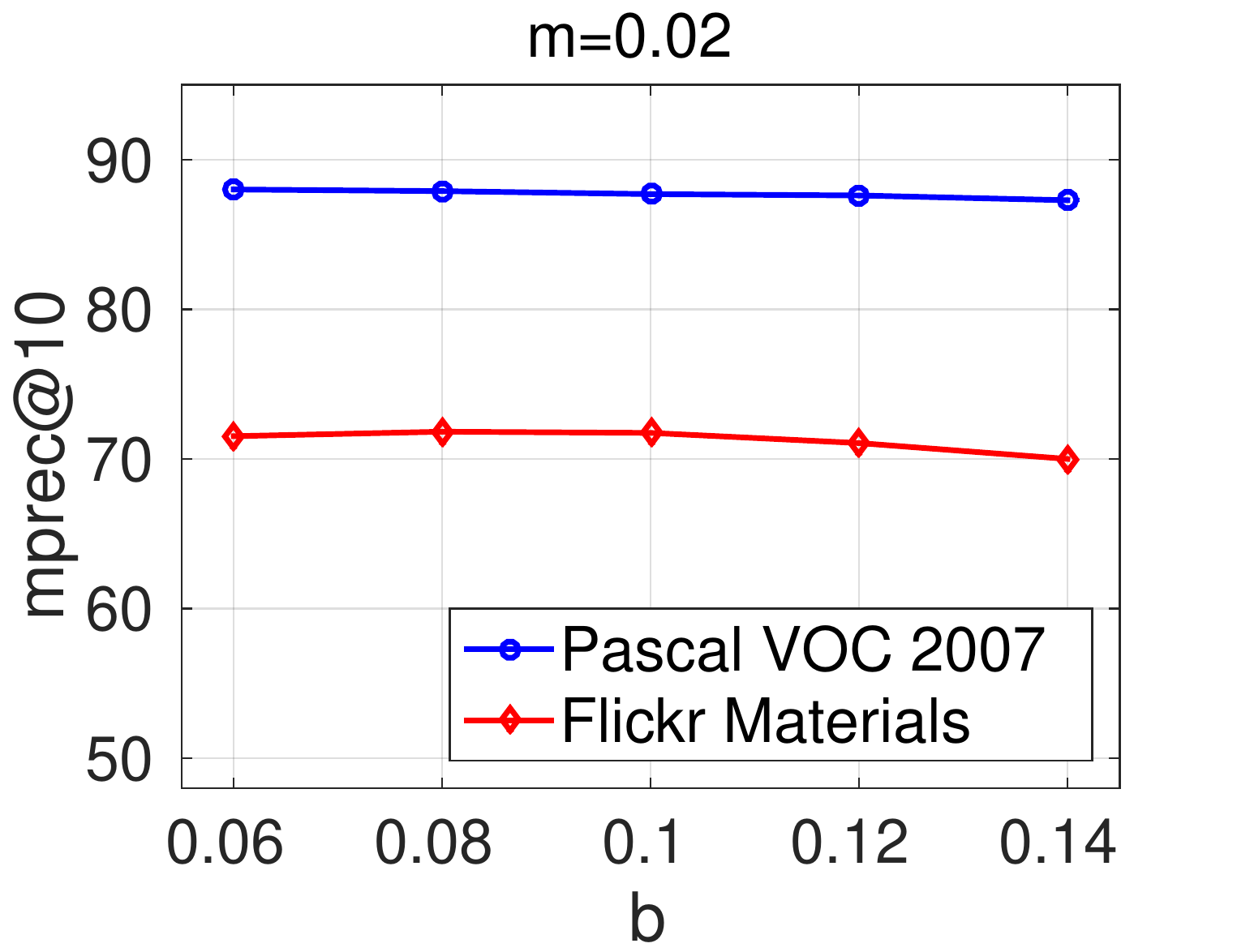} 
\caption{
Performance with varying margin $m$ (fixed $b=0.1$) and bias $b$ (fixed $m=0.02$), for the proposed
method ($d=64$).
} 
\vspace{-1em}
\label{fig:perf_m_b}
\end{figure}

\noindent
\textbf{Datasets.} 
We report experiments with seven publicly available datasets: Pascal VOC 2007
\cite{Everingham2007}, Scene-15 \cite{LazebnikCVPR2006}, Flickr Materials (FMD) \cite{SharanJV2009},
Oxford 102-Flowers \cite{NilsbackICVGIP2008}, Leeds Butterflies \cite{WangBMVC2009}, Caltech UCSD
Birds 200-2011 (CUB) \cite{WahCUB_200_2011} and Human Attributes (HAT) \cite{SharmaBMVC2011}.
Tab.~\ref{tab:datasets} gives the statistics, \ie number of classes and number of training,
validation and testing images, for the seven datasets. We use the provided
\texttt{train+val} sets for training and \texttt{test} set for testing where available. Otherwise,
for Scene-15 we randomly take 100 images per class for training and rest for testing and for Leeds
Butterflies we take the first 20 images of each class for testing and rest for training. We use only
the classes with less than 1000 positive images in the HAT dataset, as including the images with
high number of positives was giving saturated results, while respecting the \texttt{train/val/test}
split provided.
\vspace{0.6em} \\
\textbf{Image features.} 
CNN features have been quite successful in image classification after the seminal work of Krizhevsky
\etal \cite{KrizhevskyNIPS2012}, have been competitive for many different computer vision tasks
\cite{RazavianCVPRW2014}. Thus, we use CNN features using the \texttt{MatConvNet}
\cite{Vedaldi2014} library, with the 16 layer model
\cite{SimonyanICLR2015} which is pre-trained on the ImageNet dataset \cite{DengCVPR2009}. We use
the outputs of the last fully connected layer after linear rectification \ie our features are
non-negative. To validate the baseline implementation we used the extracted 4096 dimensional CNN
features with linear SVM to perform the classification task for the different datasets.
Tab.~\ref{tab:datasets} gives the performance and those of the best method on the dataset.  We see
that the features used here perform competitively to existing methods. Hence, we conclude that the
features we use in the experiments are relevant and comparable to the state-of-the-art.
\begin{table*}
\centering
\newcolumntype{C}{>{\centering\arraybackslash}p{5.5em}}
\newcolumntype{D}{>{\centering\arraybackslash}p{7.0em}}
\begin{tabular}{|r|C|C|C||C|D|}
\hline
 & \multicolumn{3}{c||}{Statistics} & \multicolumn{2}{c|}{Performances} \\
Dataset & \#classes & \#train+val & \#test & Present & Existing \\
\hline 
\hline
          Scene-15$^1$ \cite{LazebnikCVPR2006}   &  15 & 1500 & 2985 & 90.3 & \ \  91.6      \ \ \  \cite{ZhouNIPS2014}       \\
  Flickr Materials$^1$ \cite{SharanJV2009}       &  10 &  500 &  500 & 79.6 & \ \  82.8\  \  \ \ \ \cite{CimpoiArxiv2014} \ \ \\
 Leeds Butterflies$^1$ \cite{WangBMVC2009}       &  10 &  632 &  200 & 99.0 & \ \  96.4      \ \ \  \cite{LiECCV2014}         \\
   Pascal VOC 2007$^2$ \cite{Everingham2007}     &  20 & 5011 & 4952 & 86.1 & \ \  89.7      \ \ \  \cite{SimonyanICLR2015}  \\
  Human Attributes$^2$ \cite{SharmaBMVC2011}     &  27 & 7000 & 2344 & 55.4 & \ \  59.7      \ \ \  \cite{SharmaCVPR2013}     \\
Oxford 102-Flowers$^1$ \cite{NilsbackICVGIP2008} & 102 & 2020 & 6149 & 84.3 & \ \  86.8      \ \ \  \cite{RazavianCVPRW2014}  \\
Caltech UCSD Birds$^1$ \cite{WahCUB_200_2011}    & 200 & 5994 & 5794 & 63.1 & \ \  69.1\  \  \ \ \ \cite{CimpoiArxiv2014} \ \ \\
\hline
\end{tabular}
\caption{
Statistics of the datasets used for the experiments, along with the performances ($^1$mean class
accuracy or $^2$mean average precision) of the features used (with SVM) \vs existing methods. The
performances serve to demonstrate that the features we use are competitive.
}
\label{tab:datasets}
\vspace{-1em}
\end{table*}
\vspace{0.6em} \\
\textbf{Baselines.} 
We report results with five baselines. In all cases the final vectors are compared using Euclidean
distance. As a reference, we take the full ($\ell_2$ normalized) 4096 dimensional CNN features
(denoted `No proj' in the figures) with Euclidean distance.  Such a system was recently shown to be
competitive for instance retrieval \cite{BabenkoECCV2014}. As the first baseline, we do Principal
Component Analysis (PCA) based dimensionality reduction. As the second baseline, we learn a metric
using the Neighborhood Component Analysis (NCA) \cite{GoldbergerNIPS2004}. As the third and fourth
baseline, we learn a metric using the Large Margin Nearest Neighbor (LMNN) \cite{WeinbergerJMLR2009}
algorithm with (i) vectors reduced to dimension $d$ with PCA, and learning a square
projection metric, denoted LMNN(s), and (ii) vectors reduced to 256 dimensions (98\%
variance on average) with PCA (for efficiency) and then learning a rectangular projection matrix, denoted
LMNN(r). HAT and VOC 2007 datasets have some images which have multiple labels; for training NCA and
LMNN baselines such images were removed from the training set as these algorithm use a multi-class
supervision. Publicly available
code\footnote{\url{http://www.cse.wustl.edu/~kilian/code/lmnn/lmnn.html}} is used for NCA and LMNN
algorithms. As the final baseline we use the linear metric learning (ML) algorithm as described in
\S\ref{sec:background} with similar objective function (modulo
nonlinearity) and the same training data as the proposed method.
\vspace{0.6em} \\
\textbf{Evaluation.} We report results for the retrieval setting. We use the \texttt{train+val} sets of
the datasets for training our baselines and the proposed nonlinear method. We use each \texttt{test}
image as a query and the rest of the \texttt{test} images as the gallery and report the
mean precision@$K$ (mprec@$K$) \ie average of precision@$K$ is computed for queries of a given class, 
averaged over the classes. The CNN features for the test images are transformed by the
respective methods and are compared with Euclidean distance.
\vspace{0.6em} \\
\textbf{Implementation details.} 
We fix the number of iterations to one million. We sample (up to) 500,000 pairs of similar and
dissimilar vectors from the \texttt{train+val} set for training both the baseline linear metric
learning and the proposed nonlinear metric learning (NML). The baseline methods LMNN and NCA
do not use similar pairwise constraints but constraints derived using class labels. The vectors were
$\ell_2$ normalized for all baselines\footnote{We also tried $\ell_1$ normalization with $\ell_1$
and $\chi^2$ distances for the reference 4096 dimensional vectors, the results were similar (see
supplementary material).} and were $\ell_1$ normalized for the proposed method (as the $\chi^2$
kernel is based on histograms). The bias and margin for the linear metric learning baseline were
fixed to $b=1$ and $m=0.2$ while those for the proposed nonlinear method were fixed to $b=0.1$ and
$m=0.02$. These values were chosen based on preliminary experiments on the VOC 2007 validation set
and were kept \emph{constant for all experiments} reported \ie no dataset specific tuning was
done. The algorithm is not very sensitive to the choice of $m$ and $b$,
Fig.~\ref{fig:perf_m_b} show the performances for a range of $m$ and $b$ values (with the other
fixed resp.) on the Flickr Materials \cite{SharanJV2009} and Pascal VOC 2007 \cite{Everingham2007}
datasets. The bias and margin need to be kept sufficiently low. The reason for low values of bias,
and resp.\ margin, for the nonlinear method is that since the vectors are $\ell_1$ normalized
histograms, the maximum score they can give with the $\chi^2$ kernel is 1 (which only happens with
the vector itself) and hence the scale of the distances is expected to be less than 1 for the
proposed nonlinear method. 

\begin{figure*}
\centering
\includegraphics[width=0.245\textwidth,trim=20 180 190 155,clip]{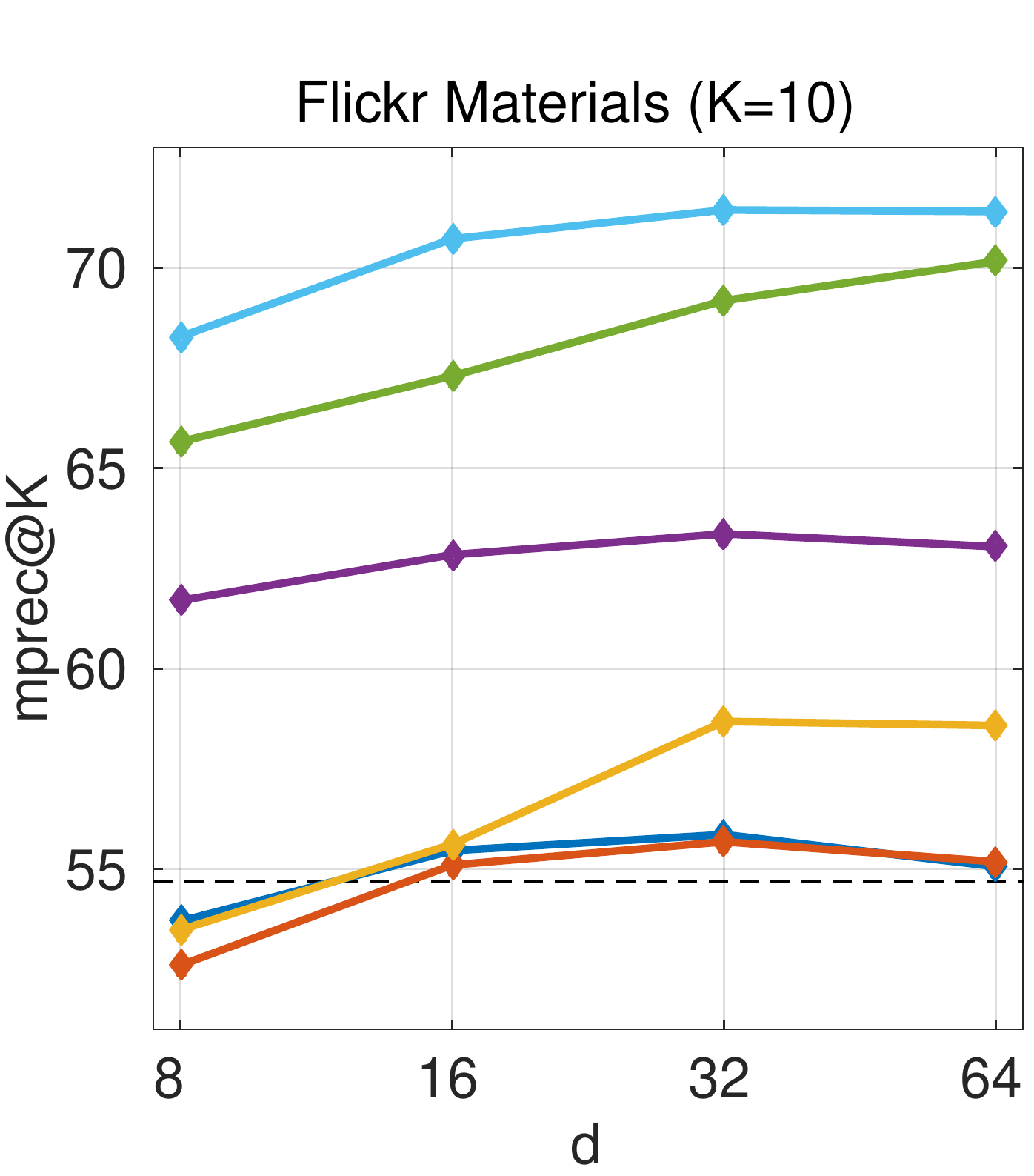}   
\includegraphics[width=0.245\textwidth,trim=20 180 190 155,clip]{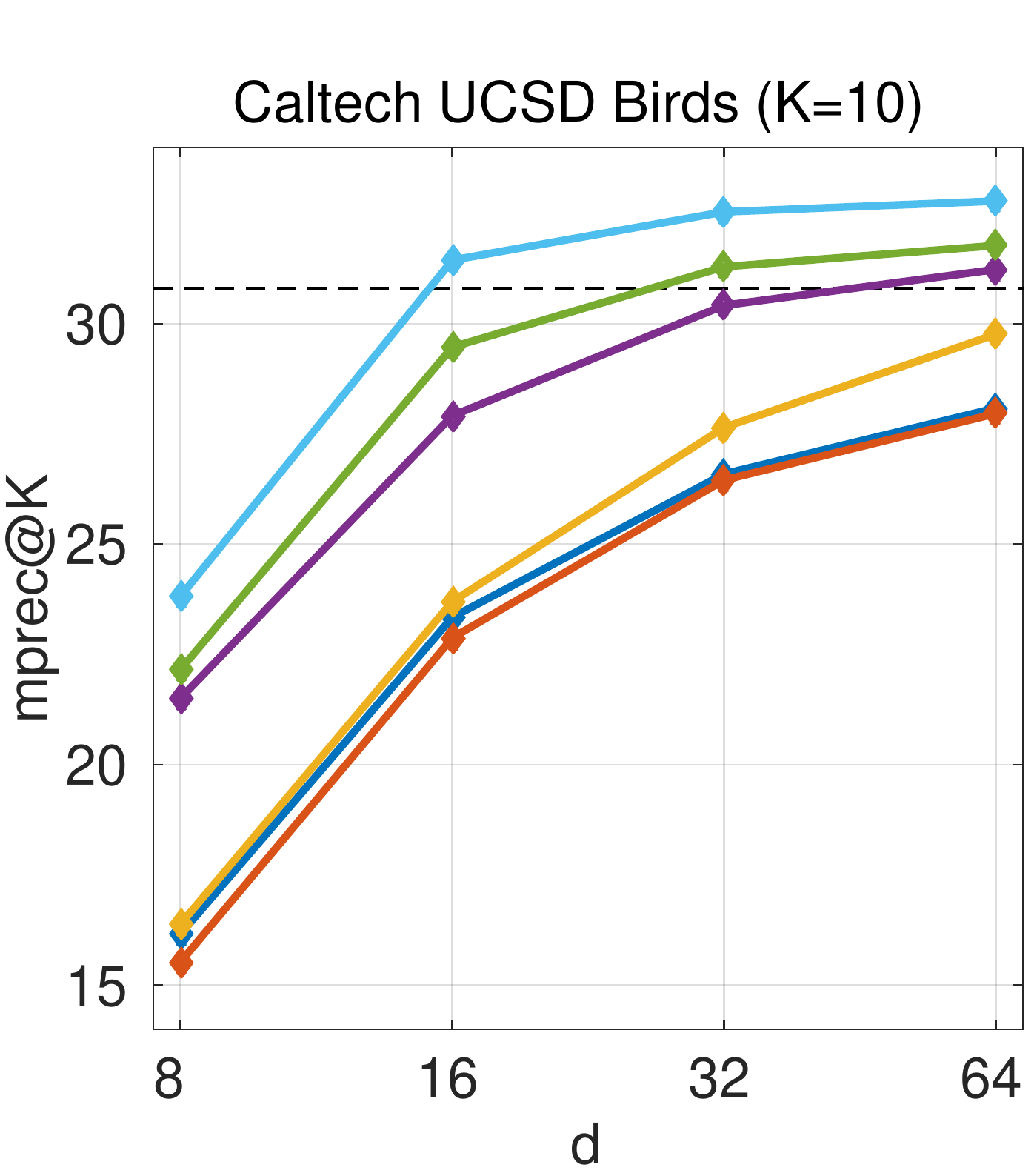}   
\includegraphics[width=0.245\textwidth,trim=20 180 190 155,clip]{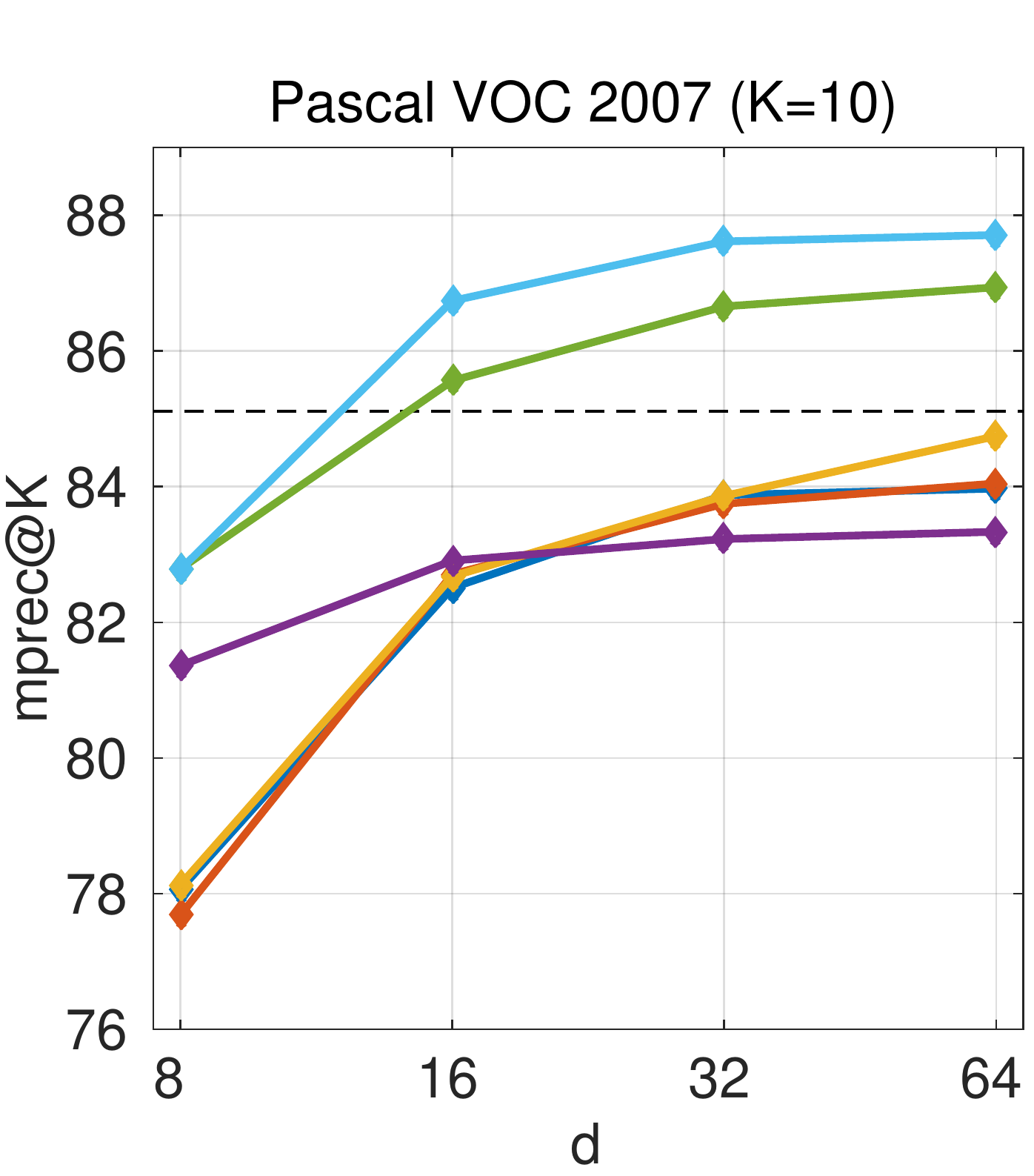}   
\includegraphics[width=0.245\textwidth,trim=20 180 190 155,clip]{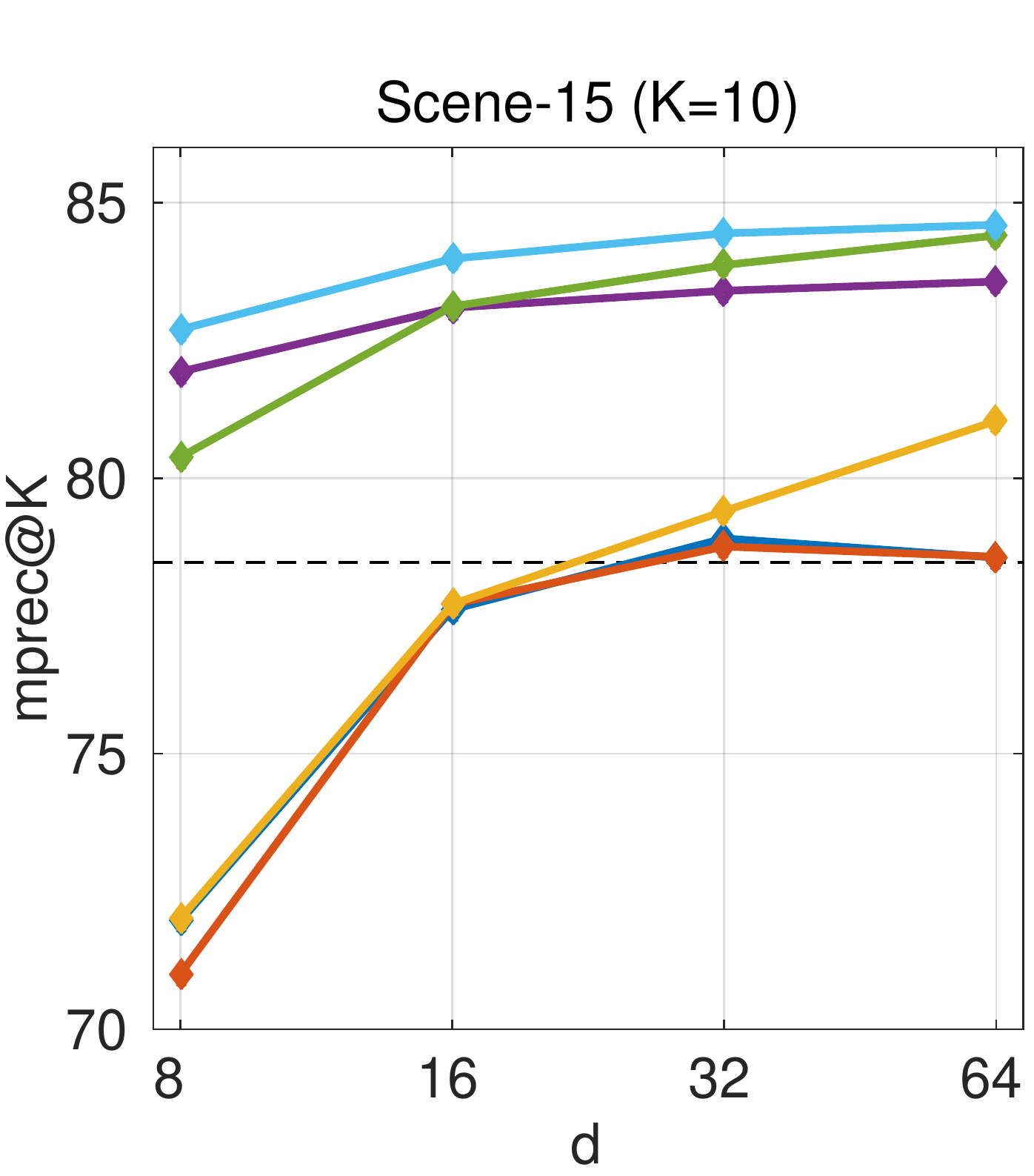}   \\ 
\hspace{2em}
\includegraphics[width=0.245\textwidth,trim=20 180 190 155,clip]{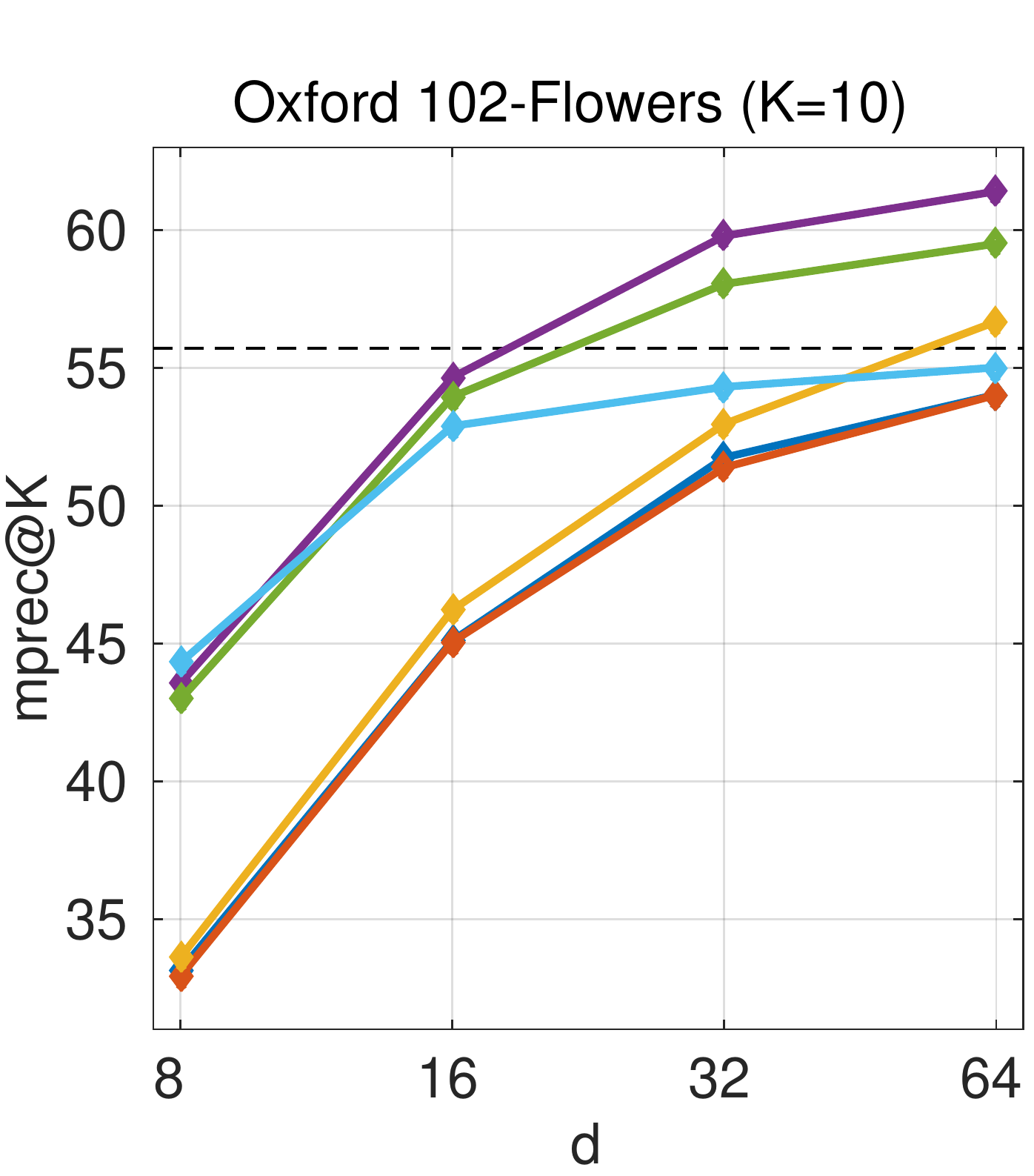}   
\hspace{4px}
\includegraphics[width=0.245\textwidth,trim=20 180 190 155,clip]{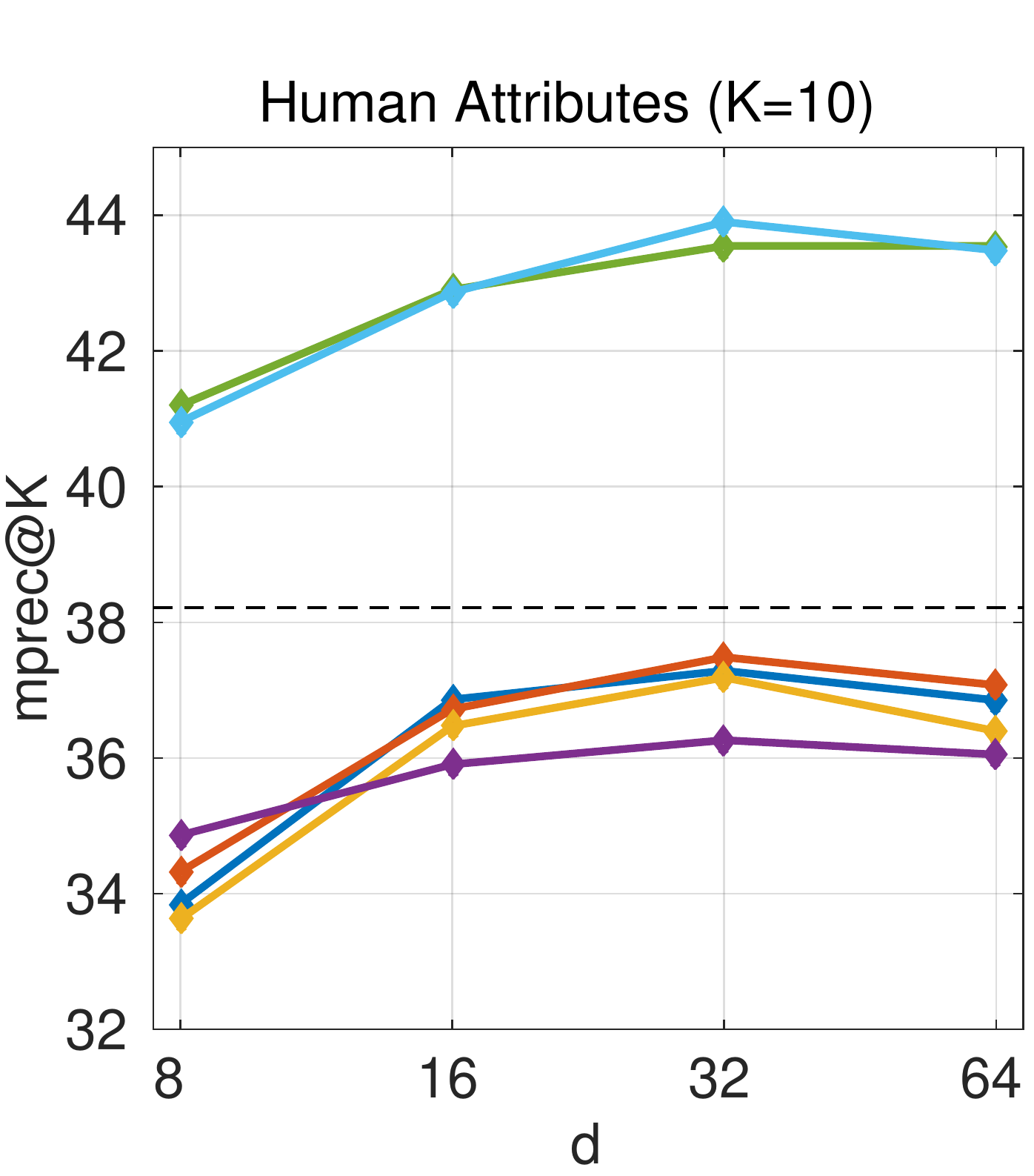}
\includegraphics[width=0.335\textwidth,trim=20 180  50 155,clip]{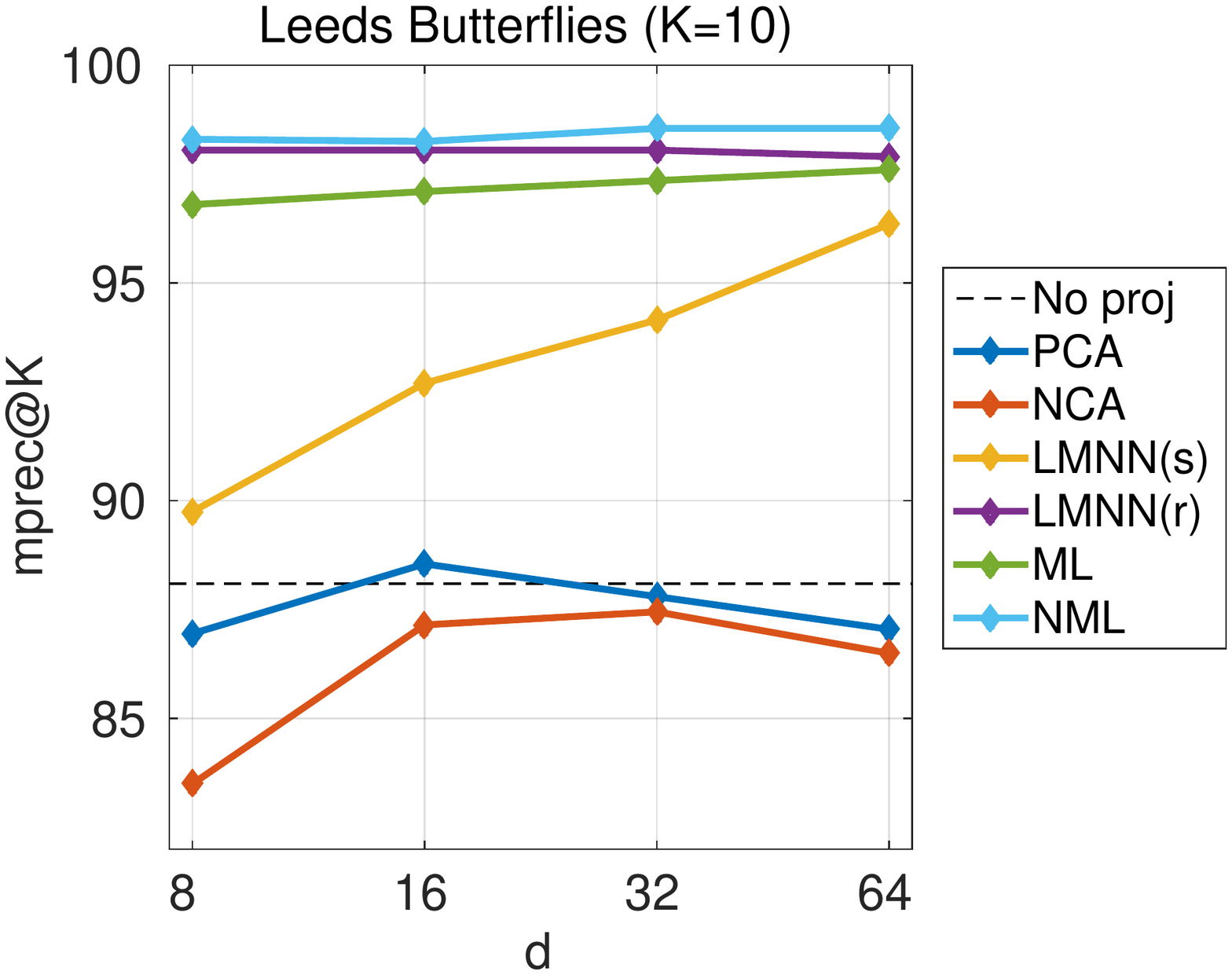}   
\caption{
Comparisons of methods on the different datasets for number of top retrievals $K=10$, and projection
dimension $d \in \{8,16,32,64\}$ (see \S\ref{sec:exp_quant}).  
}
\vspace{-1em}
\label{fig:prec_d}
\end{figure*}

\begin{figure*}
\centering
\includegraphics[width=0.245\textwidth,trim=20 180 190 155,clip]{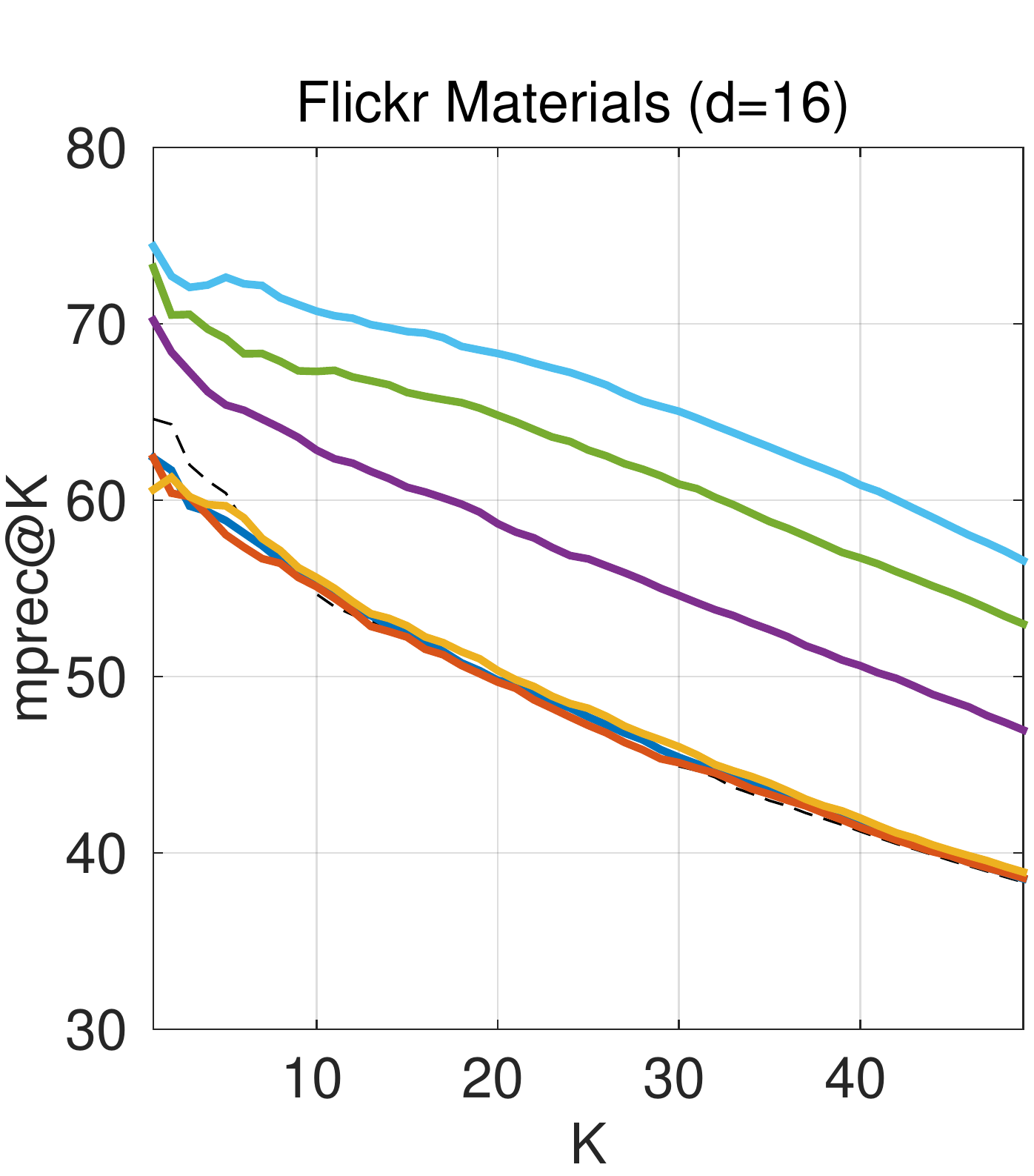}   
\includegraphics[width=0.245\textwidth,trim=20 180 190 155,clip]{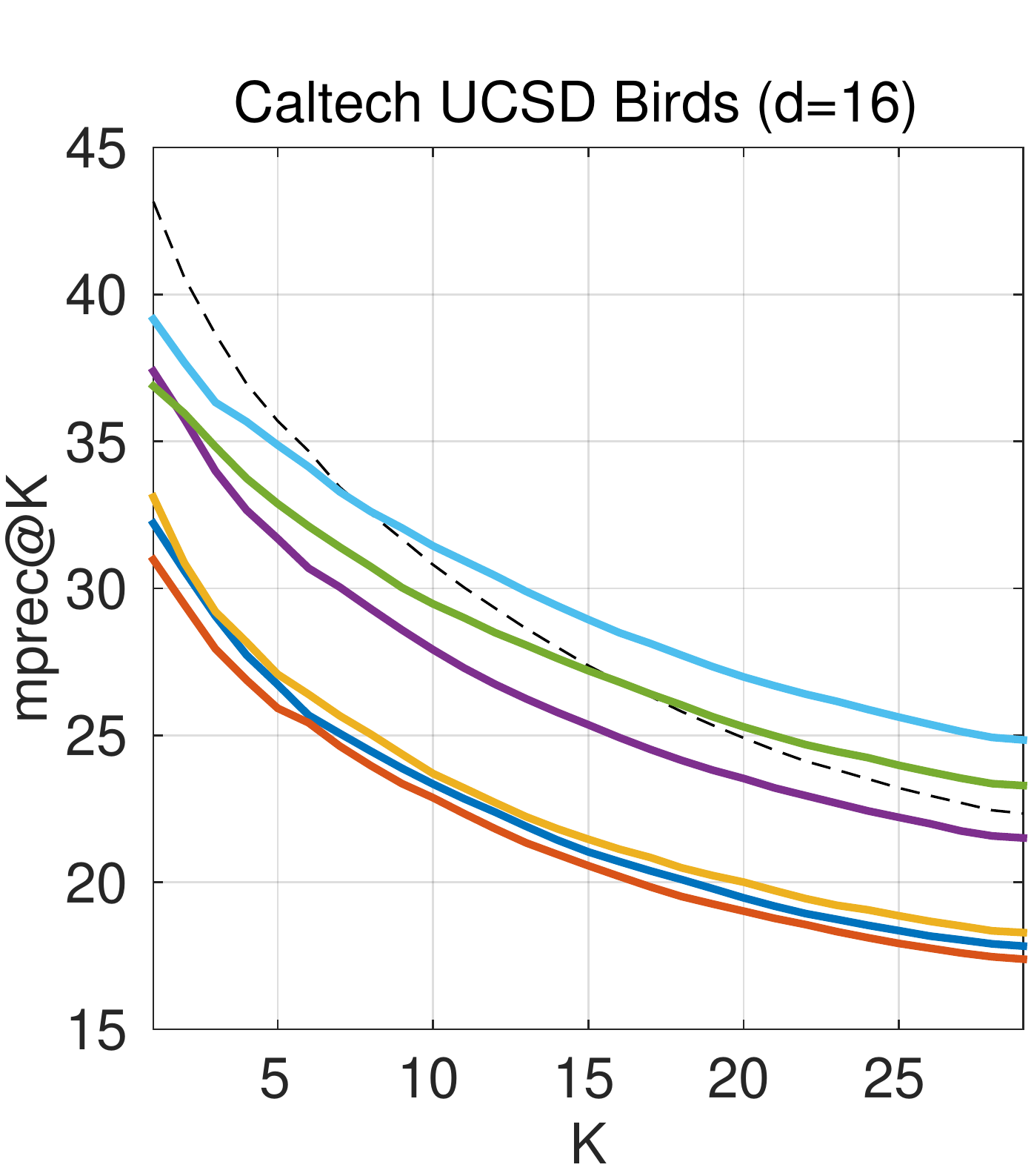}   
\includegraphics[width=0.245\textwidth,trim=20 180 190 155,clip]{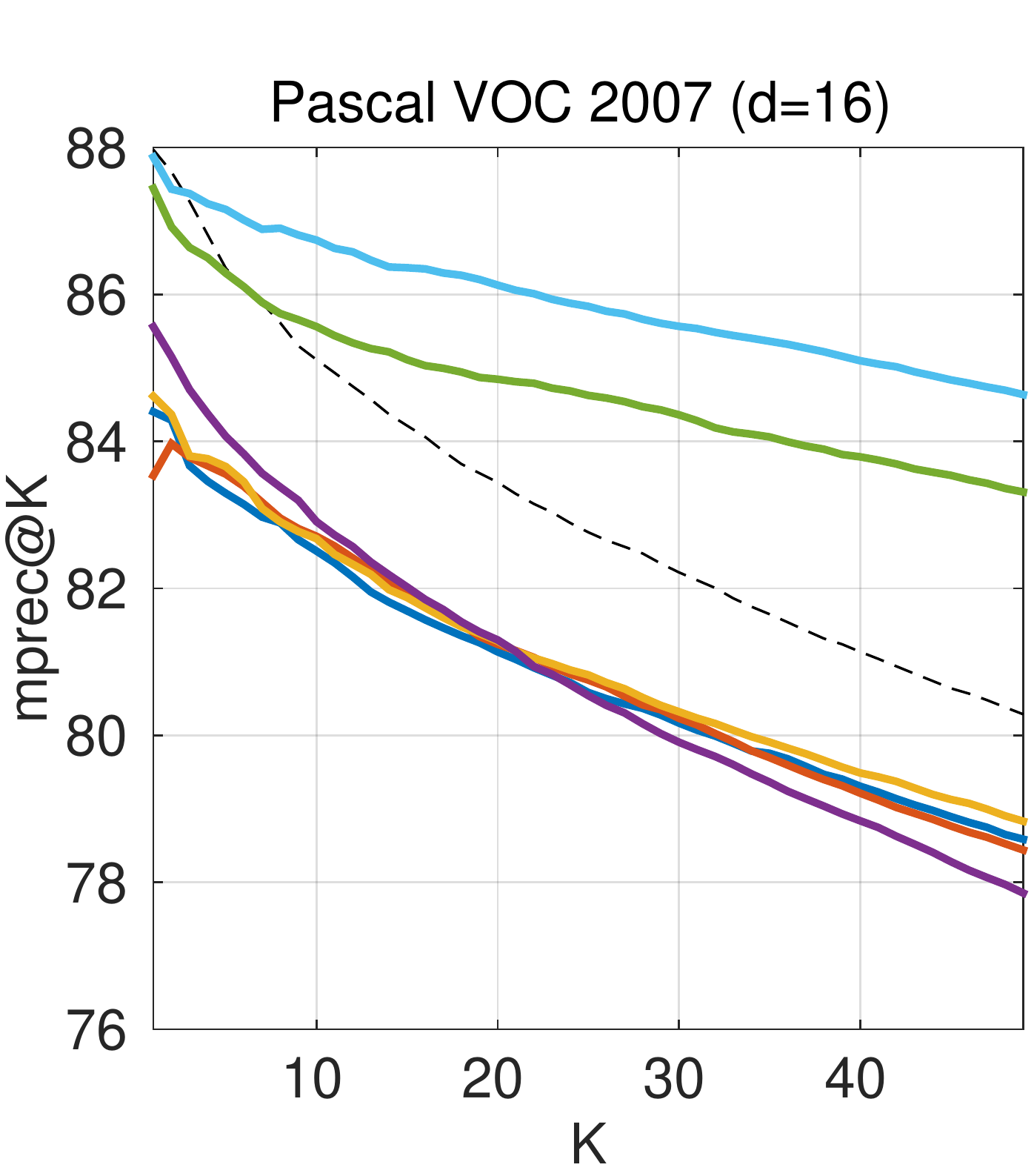}   
\includegraphics[width=0.245\textwidth,trim=20 180 190 155,clip]{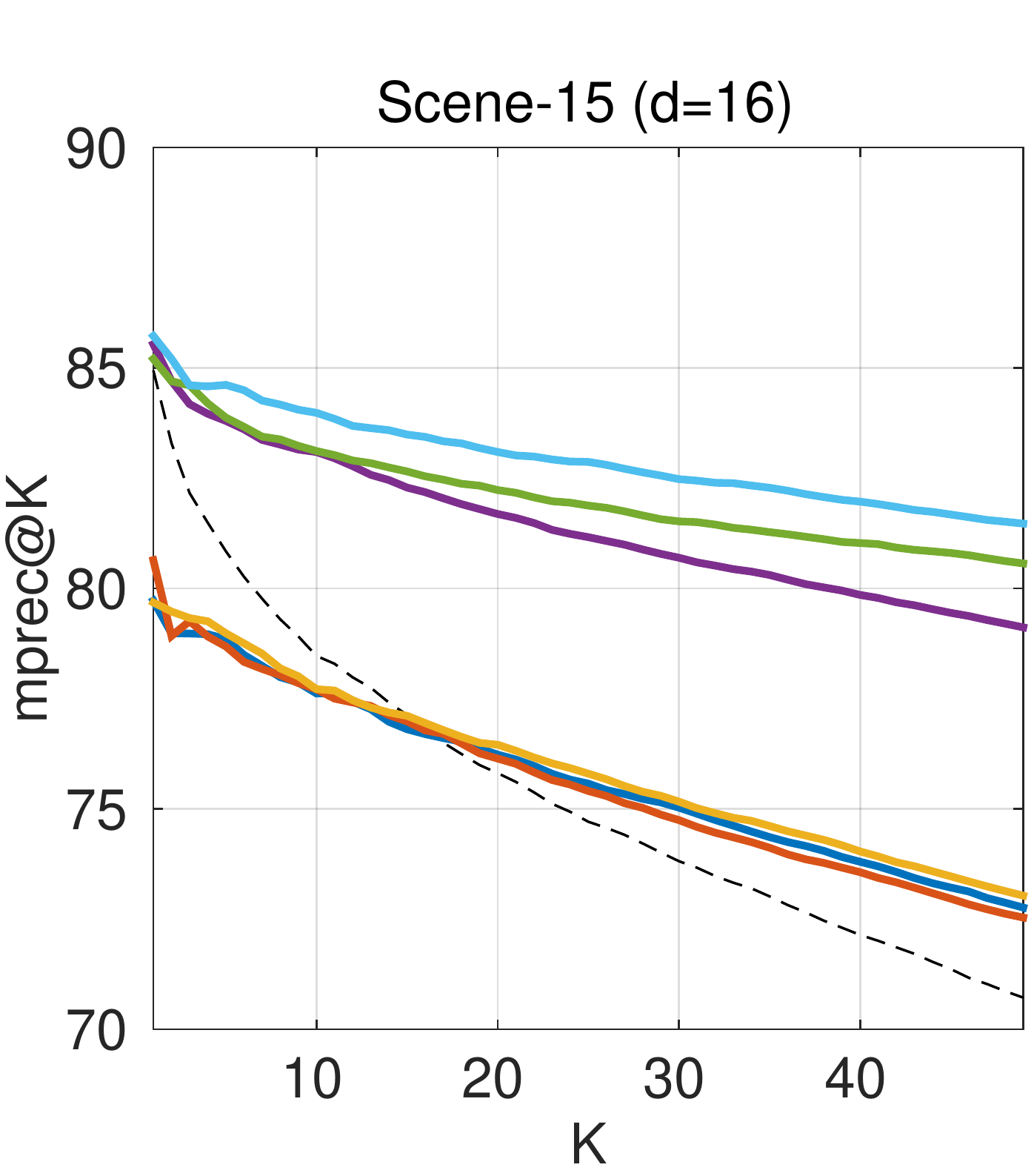}   \\
\hspace{2em}
\includegraphics[width=0.245\textwidth,trim=20 180 190 155,clip]{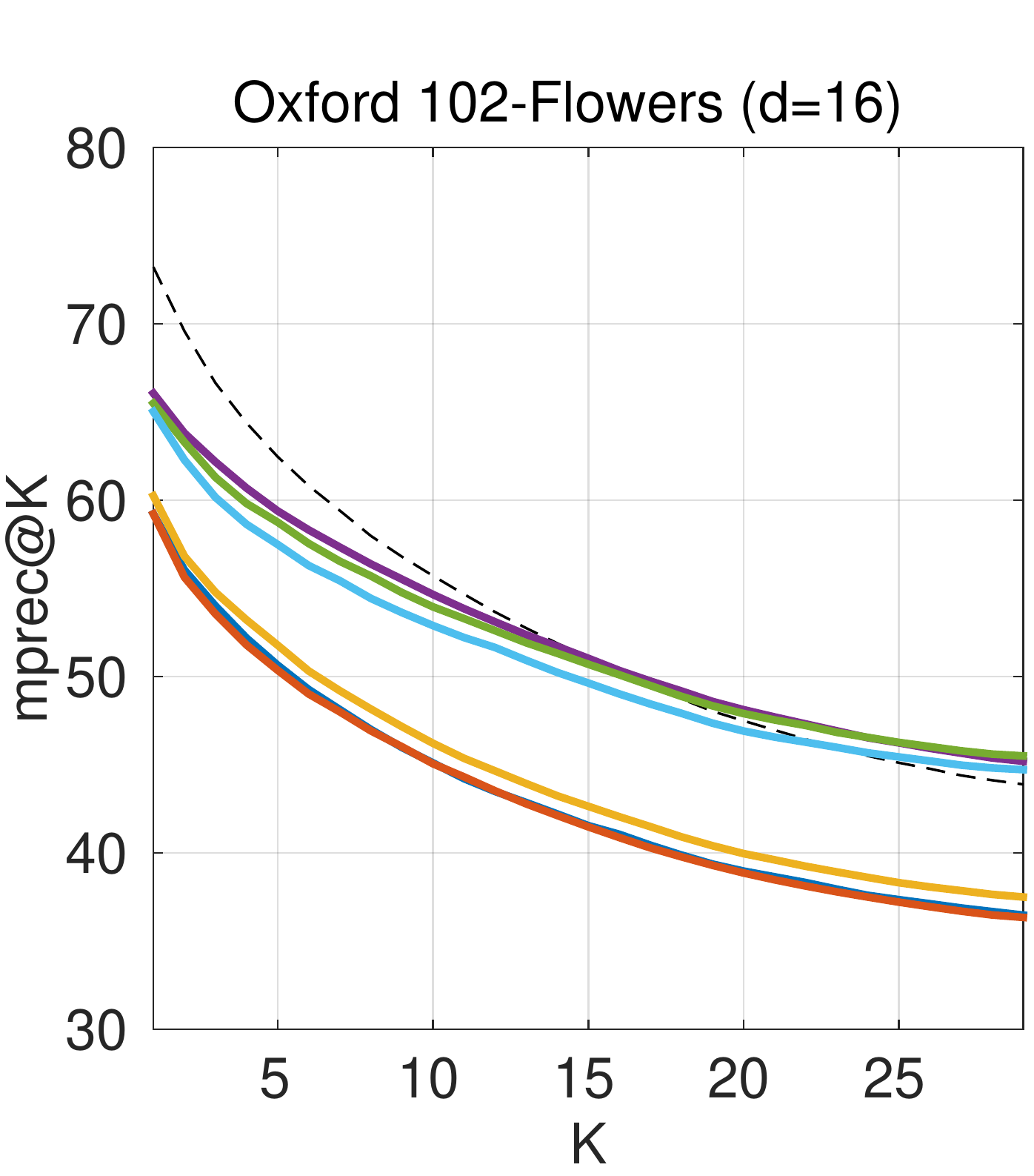}   
\hspace{4px}
\includegraphics[width=0.245\textwidth,trim=20 180 190 155,clip]{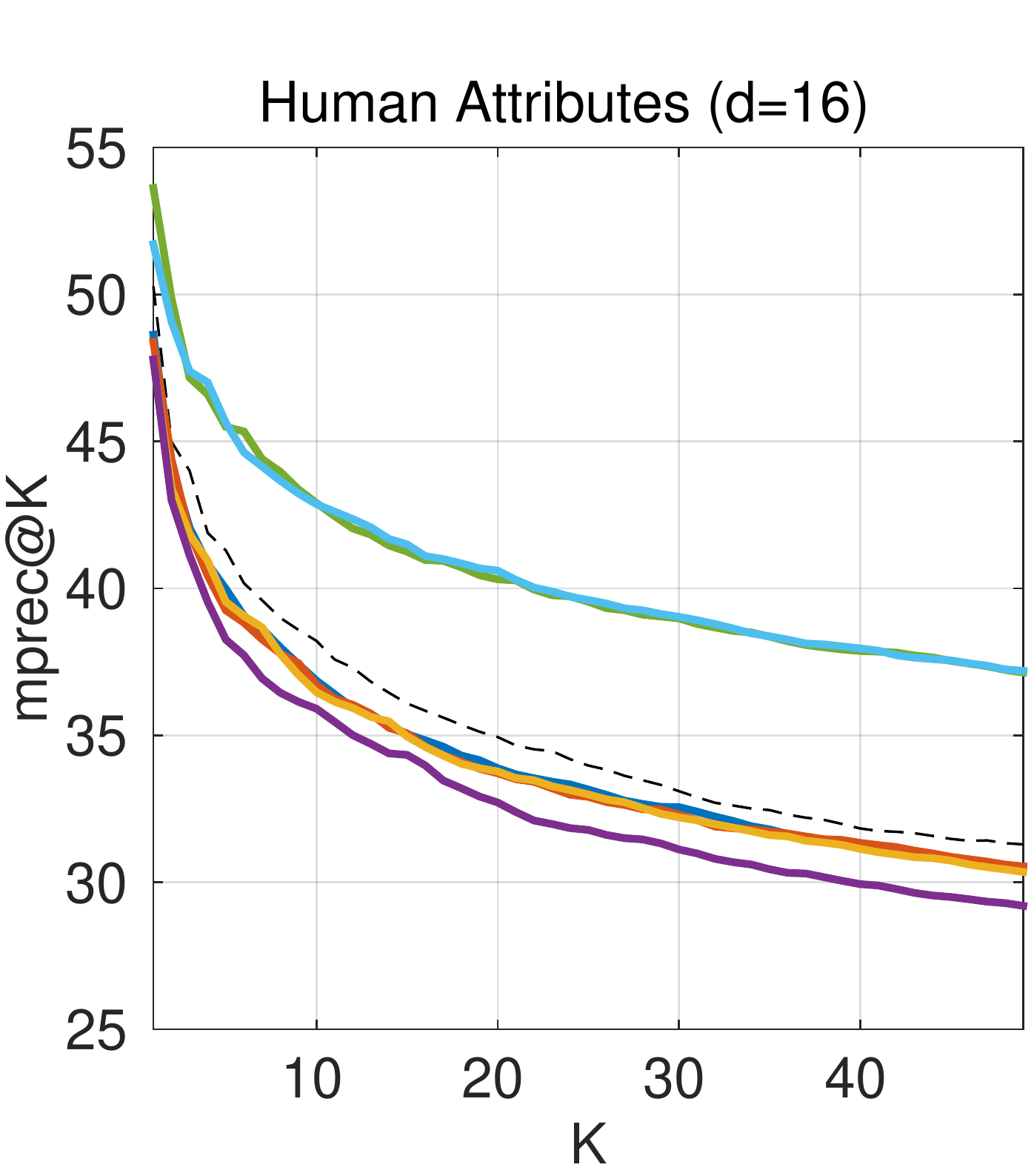}   
\includegraphics[width=0.335\textwidth,trim=20 180  50 155,clip]{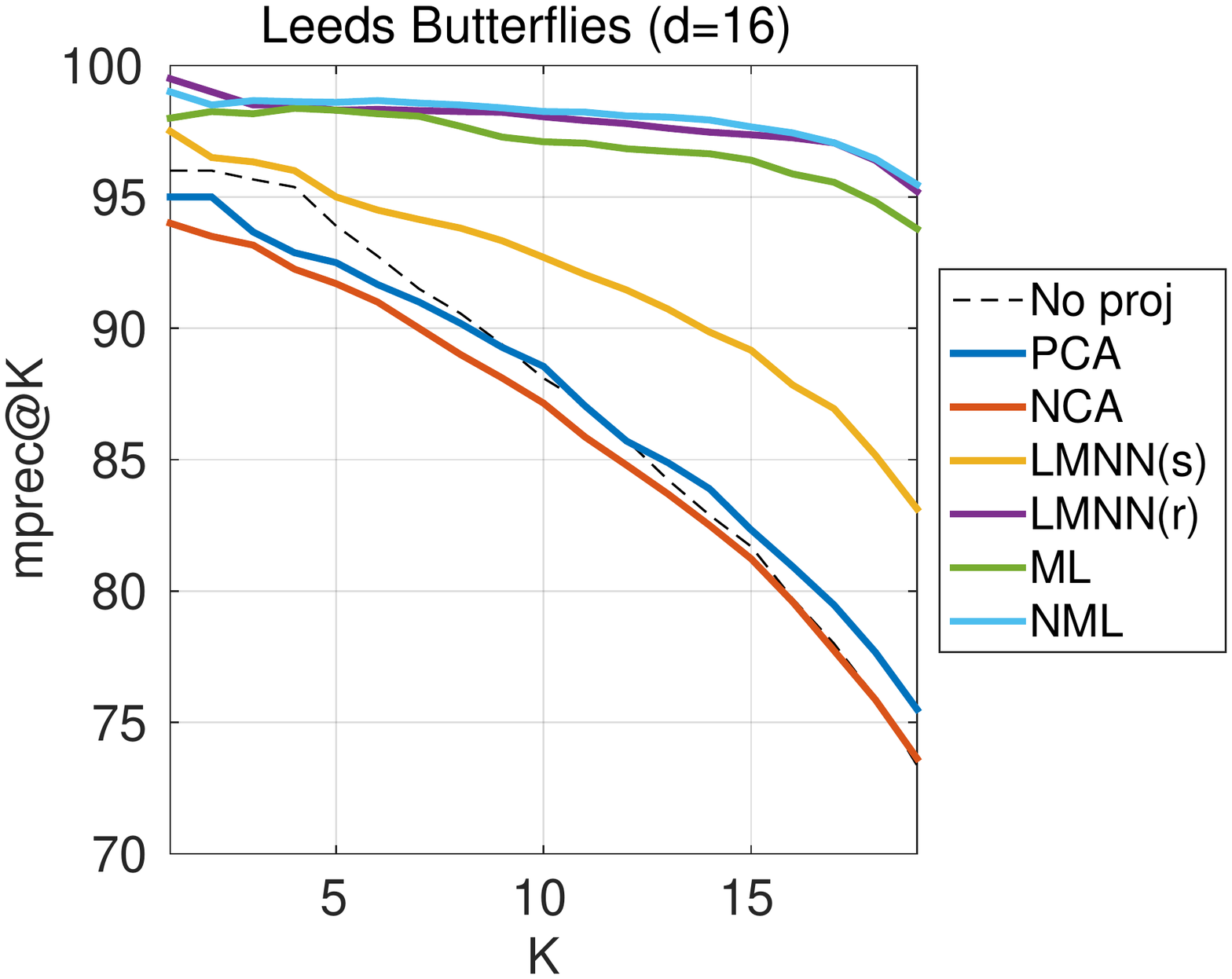}   
\caption{
Comparisons of methods on the different datasets for number of top retrievals $K \in
[1,\max(50,n^+-1)]$ and projection dimension $d=16$ (see \S\ref{sec:exp_quant}).  
}
\label{fig:prec_k}
\end{figure*}

\subsection{Quantitative results}
\label{sec:exp_quant}
We present some results for typical setting and analyse them.
We refer the reader to the supplementary material for additional results.

Fig.~\ref{fig:prec_d} shows the performance of the different methods for different projection
dimensionality $d \in \{8,16,32,64\}$, with a fixed $K=10$, \ie the number of top images retrieved.
Fig.~\ref{fig:prec_k} shows the performance for different $K \in [1,50]$ with a fixed $d=16$. The
Birds and Butterflies datasets have $n^+=30$ and $n^+=20$ images per class and hence we only report
for $K$ up to $29$ and $19$ for them, respectively. 

We observe, from Fig.~\ref{fig:prec_d}, for $K=10$ \ie precision for the top 10 retrieved images,
that among the baselines, supervised LMNN is generally better than supervised NCA and unsupervised 
PCA while the ML baseline is the most competitive. The proposed NML is generally better or at least as
good as the baselines, notably the most competitive ML baseline. It is outperformed by the baselines
only on the Flowers dataset at relatively higher dimensional (\eg $d=32,64$) projections. The
improvements are consistent over Materials, Birds, Objects and Scenes datasets for
all $d\in\{8,16,32,64\}$. For the Human Attributes, NML does not improve the baseline ML, but is
essentially similar to it \ie does not deteriorate either. On the Flowers dataset NML is 
outperformed by the baselines at higher dimensions. Further, from Fig.~\ref{fig:prec_k} we note a
similar general trend, for a fixed projection dimension $d=16$ while observing the precision at
varying number of $K$ top retrievals. The proposed NML consistently performs better on the
Materials, Birds, Objects and Scenes datasets while on the Human Attributes and
Butterflies datasets it performs similar, compared to the baselines. 

In addition to the results shown here, in general NML was found to be better or at least similar to
the baseline ML methods (see supplementary material). The few cases where it is not better are
those at high projection dimensions $d$, which could be explained by the higher need of nonlinearity
of the model at lower dimensions, where an equivalent linear model with similar number of parameters
may not suffice, while at higher dimensions, the larger number of parameters for the linear method
are sufficient for the task.  Thus, we conclude that except in a single case, \ie the Flowers
dataset at relatively higher dimensional projections, NML improves over the reported baselines
demonstrating the benefit of the proposed scalable nonlinear embedding and efficient SGD based
learning thereof. 

Tab.~\ref{tab:kml_vs_nml} gives typical performances of the kernelized version
(\S\ref{sec:background}) of baseline ML (kML) \vs the proposed NML, on the Materials dataset. NML 
obtains similar performances while being $O(d)$ at test time \cf $O(N)$ for kML.

It is also interesting to note the performance of the method after compression compared to using
the full 4096 dimensional CNN feature without any compression. In both Fig.~\ref{fig:prec_d} and
Fig.~\ref{fig:prec_k} the dotted black line shows this performance as a reference. We see that more
often than not, the discriminatively learnt NML projection improves the performance over the full
features. For the Materials, Human Attributes, Scenes, and Butterflies datasets this improvement is
significant. In the case of Birds and VOC 2007 Objects the performance drops below the
reference for low dimension projections while gradually improving at higher dimensional projections.
Only in the case of the Flowers dataset, NML projection does not improve this reference but is
roughly equal to it at higher dimensions, \ie $d = 64$. Thus, we conclude that the proposed NML is
also beneficial to improve performance of the reference system with full feature vector (4096
dimensional) without compression.

\begin{SCtable}[1.6]
\small
\centering
\begin{tabular}{|c|c|c|}
\hline
d & kML & NML \\
\hline 
\hline 
8  & 69.6 & 68.3 \\
16 & 70.8 & 70.7 \\
32 & 70.9 & 71.5 \\
64 & 71.4 & 71.4 \\
\hline
\end{tabular}
\vspace{-1em}
\caption{
Performance (mprec@10) of  the kernelized metric learning \vs proposed nonlinear metric
learning, on the Flickr Materials dataset, for different projection dimensions $d$.
\vspace{-1em}
}
\label{tab:kml_vs_nml}
\end{SCtable}

\sidecaptionvpos{figure}{c}
\begin{figure*}
\centering
\begin{minipage}{0.84\textwidth}
\def\len{0.1 \linewidth}
\def\lent{0.89 \linewidth}
\raisebox{4ex}{
\includegraphics[width=\len,trim=0 105 935 -20,clip]{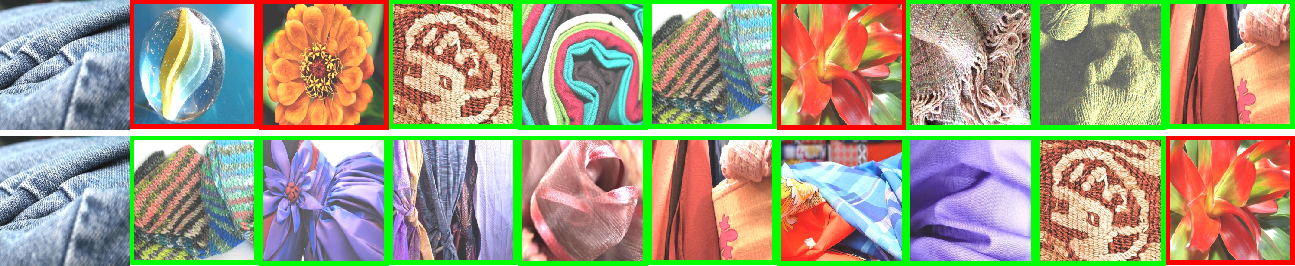}   
}
\includegraphics[width=\lent,trim=104 0 0 0,clip]{materials1}   
\\
\raisebox{4ex}{
\includegraphics[width=\len,trim=0 105 935 -20,clip]{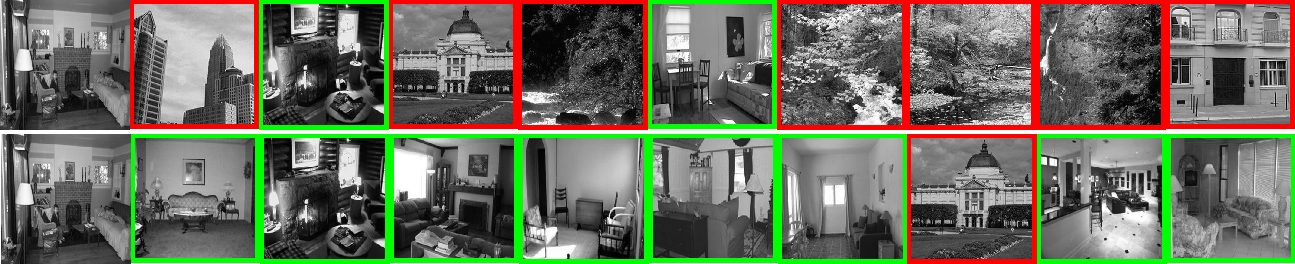}   
}
\includegraphics[width=\lent,trim=104 0 0 0,clip]{scenes2}   
\\
\raisebox{4ex}{
\includegraphics[width=\len,trim=0 105 935 -20,clip]{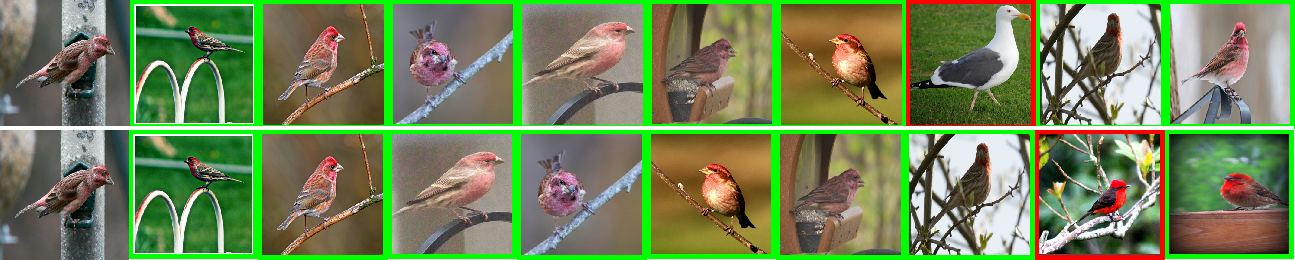}   
}
\includegraphics[width=\lent,trim=104 0 0 0,clip]{birds4}   
\\
\raisebox{4ex}{
\includegraphics[width=\len,trim=10 115 945 15,clip]{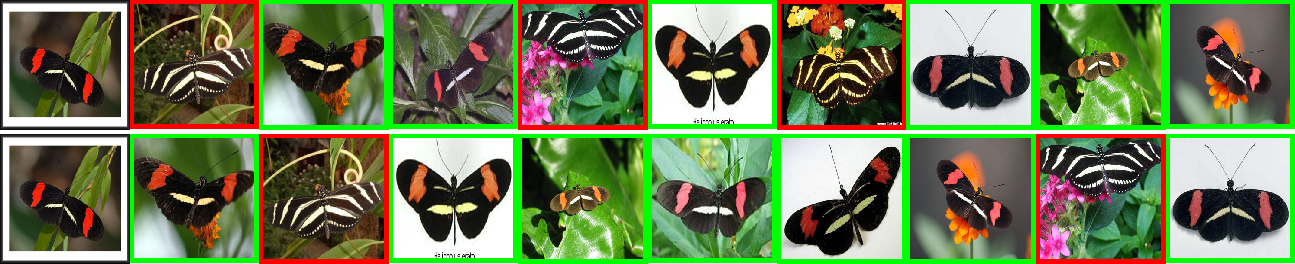}   
}
\includegraphics[width=\lent,trim=104 0 0 0,clip]{butterfly1}   
\vspace{0.1em} 
\end{minipage}
\caption{
Example retrievals for a given query image (leftmost single image), using the baseline linear metric
learning \vs the proposed nonlinear embeddings (first and second rows of each pair resp.). Both
methods project to just 8 dimensions \ie 32 bytes per image. The correctly (incorrectly) retrieved
images are highlighted using green (red) bounding boxes. 
}
\label{fig:vis}
\vspace{-0.5em}
\end{figure*}

\subsection{Qualitative results}
We show some example retrievals in Fig.~\ref{fig:vis} using (i) the baseline linear metric learning
based projections and (ii) the proposed nonlinear embeddings. For each pair of rows, the first one
corresponds to method (i) and second one to (ii). The first image is the query and the rest of the
images are retrievals by the respective methods, sorted by their scores. The vectors after
projections for both the methods are 8 dimensional floating point numbers \ie it takes 32 bytes to
store one image. The goal is to retrieve images of the same semantic category as the example query
\eg in the first example, images of `fabrics' are retrieved. Similarly other retrievals are for
specific types of birds, scenes, butterflies etc. Note the variations in the appearance of the
retrieval results, which can be attributed to the combination of the strong CNN based appearance
features and discriminatively learned embeddings.

\subsection{Computation times}
The training time (with 500,000 training pairs) is about an hour, on a single core of 3 Ghz CPU
running linux, for proposed method while the linear ML with exactly the same training data takes
around five minutes ($d=8$ for both). For testing (on 500 test images), ML takes 52 ms while the
proposed method takes 258 ms, in a similar setup. These times exclude the feature extraction time.
Both implementations are in MATLAB and not optimized for performance, however, we suspect ML has an
advantage as it depends only on matrix operations which use optimized low-level
libraries.

\section{Discussion and conclusion}
\label{sec:discuss}
We presented a novel method for supervised discriminative distance learning.
The method performs nonlinear embedding of the high dimensional vectors into a Euclidean space where
the vectors are compared using standard $\ell_2$ distance. The method is derived from an approximate
kernelization of Mahalanobis-like distance metric learning. We proposed an efficient and scalable
stochastic gradient based learning method and reported results for the task of semantic category
based image retrieval on seven publicly available challenging image datasets of materials, birds,
human attributes, scenes, flowers, objects and butterflies. Our experimental results support the
conclusion that the proposed method is consistently better than standard competitive baselines
specially in the case of low dimensional projections. We also made the observation that doing such
supervised discriminative dimensionality reduction also improves the performance of the reference
system which uses the full high dimensional feature without any compression.

The proposed method used a classic kernel ($\chi^2$ kernel) for the nonlinearity, exploring
alternate, even data driven/learned kernels within a similar framework seems an interesting future
direction to us.

\small{ 
\bibliographystyle{ieee}
\bibliography{biblio}
}

\newpage

\appendix

\section*{\Large Appendix}

\section{Normalization of the feature vectors}
Tab.~\ref{tab:baseline} gives the performance of the reference system which uses the full 4096
dimensional CNN feature, without any dimension reduction with any method. It compares different
normalizations ($\ell_1$ and $\ell_2$) coupled with appropriate distances ($\ell_1$, $\ell_2$ and
$\chi^2$).  Since the proposed method works with histograms we $\ell_1$ normalize the features, to
interpret them as histograms.  While for other methods/baselines, we use $\ell_2$ normalization. As
Tab.~\ref{tab:baseline} shows, different normalization coupled with appropriate different distance
measure achieve essentially similar results. 

\begin{table}[h]
\centering
\begin{tabular}{r|c|c|c}
\hline
Dataset & $\ell_2+\ell_2$ & $\ell_1+\ell_1$ & $\ell_1+\chi^2$ \\
\hline \hline
Scene-15 \cite{LazebnikCVPR2006}            & 85.0 & 84.5 & 84.6 \\
Flickr Materials \cite{SharanJV2009}        & 64.6 & 65.2 & 64.8 \\
Leeds Butterflies \cite{WangBMVC2009}       & 96.0 & 96.0 & 96.0 \\
Pascal VOC 2007 \cite{Everingham2007}       & 88.0 & 87.8 & 87.9 \\
Human Attributes \cite{SharmaBMVC2011}      & 50.3 & 50.2 & 49.9 \\
Oxford 102-Flowers \cite{NilsbackICVGIP2008}& 73.2 & 73.6 & 73.9 \\
Caltech UCSD Birds \cite{WahCUB_200_2011}   & 43.2 & 43.0 & 43.5 \\
\hline
\end{tabular}
\caption{
Performance (mprec@$K$ for $K=1$) for different normalization and distances on the CNN features for the
reference system using full 4096 dimensional CNN feature. $\ell_2 + \ell_2 (\ell_1+\ell_1)$ means the features were
$\ell_2 (\ell_1)$ normalized and compared with $\ell_2 (\ell_2)$ distance while $\ell_1+\chi^2$
means they were $\ell_1$ normalized and compared with $\chi^2$ distance.
}
\label{tab:baseline}
\end{table}

\section{Additional results for different parameter settings}
\begin{figure*}
\centering
\includegraphics[width=0.245\textwidth,trim=20 180 190 155,clip]{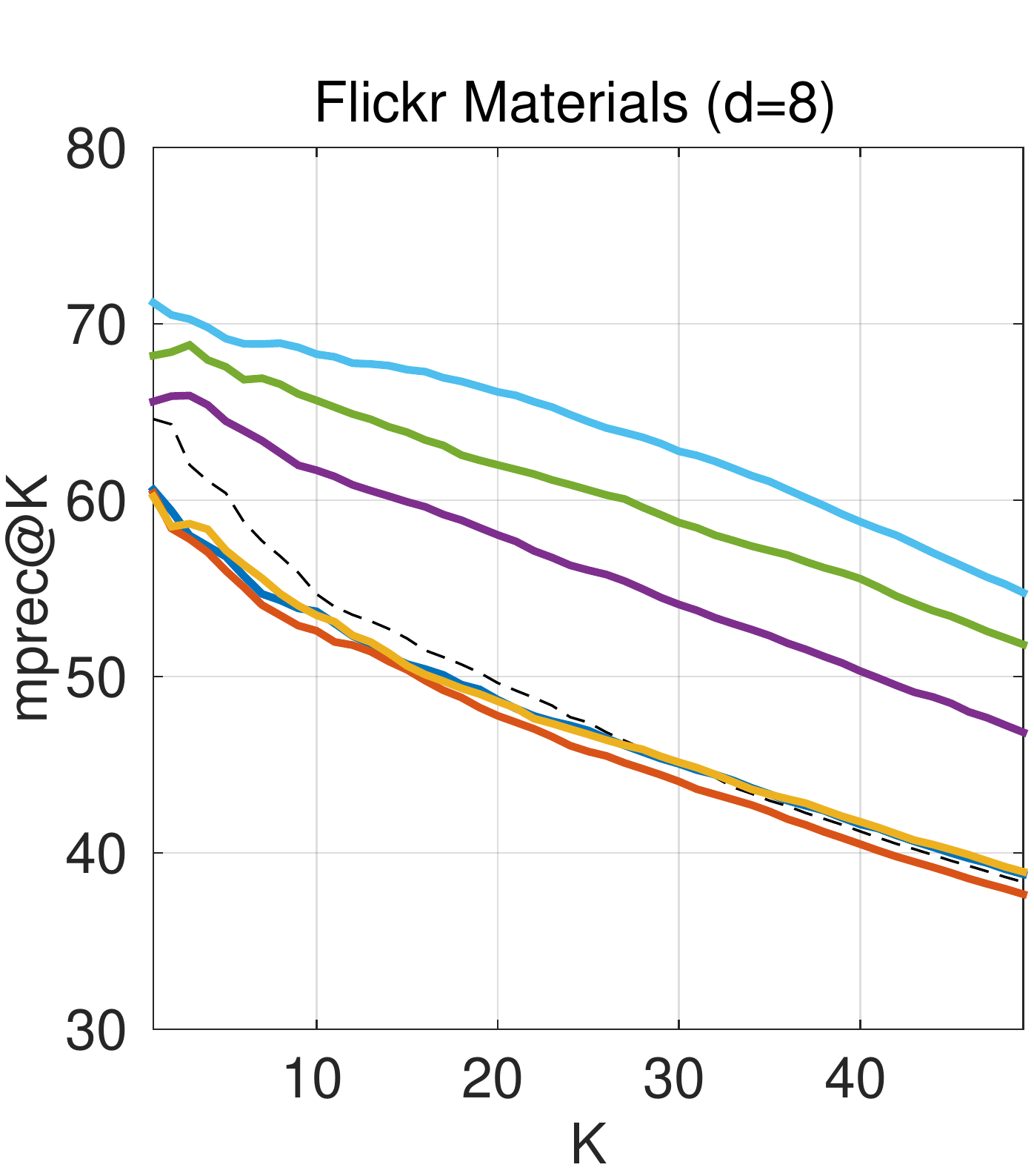}   
\includegraphics[width=0.245\textwidth,trim=20 180 190 155,clip]{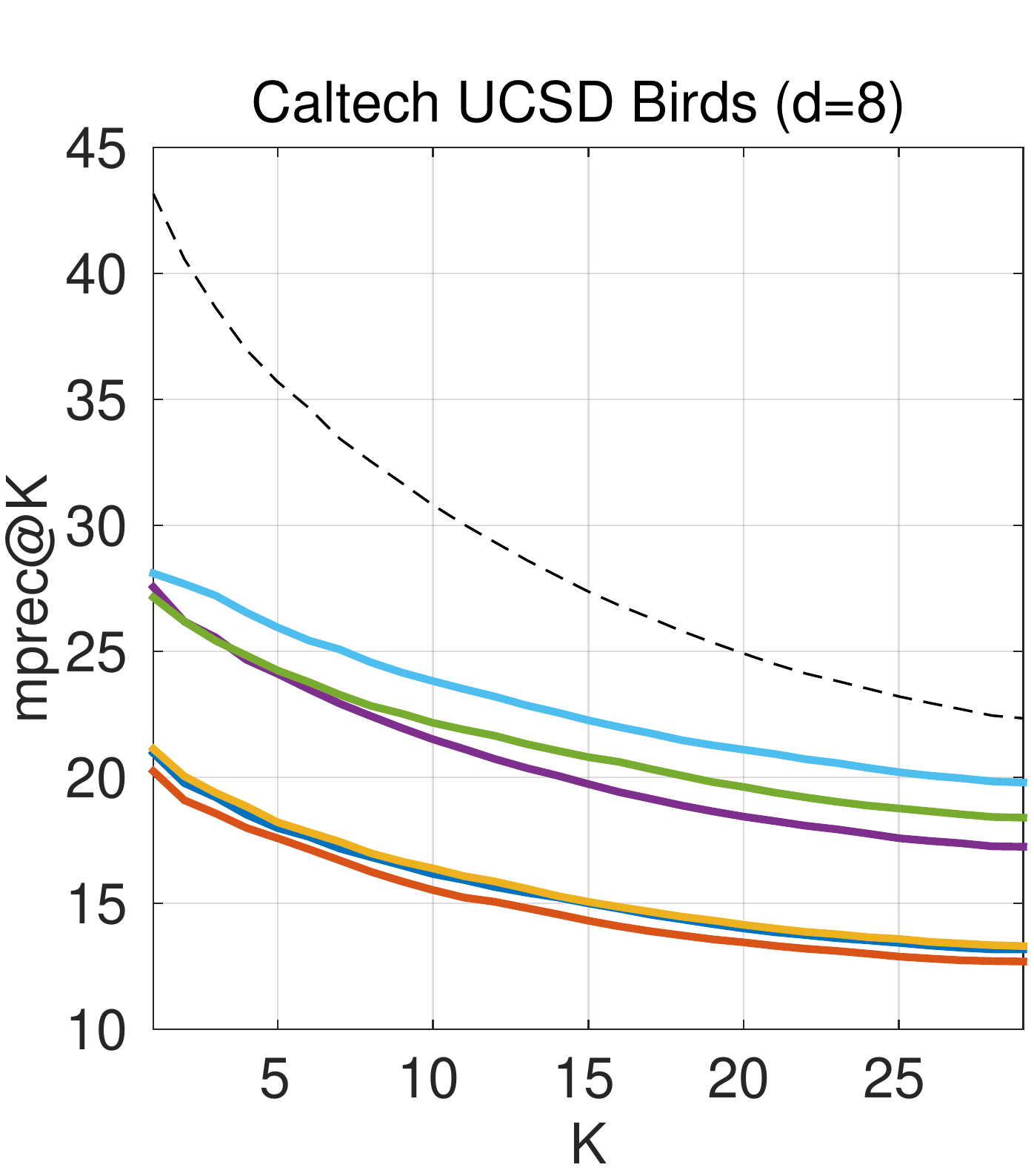}   
\includegraphics[width=0.245\textwidth,trim=20 180 190 155,clip]{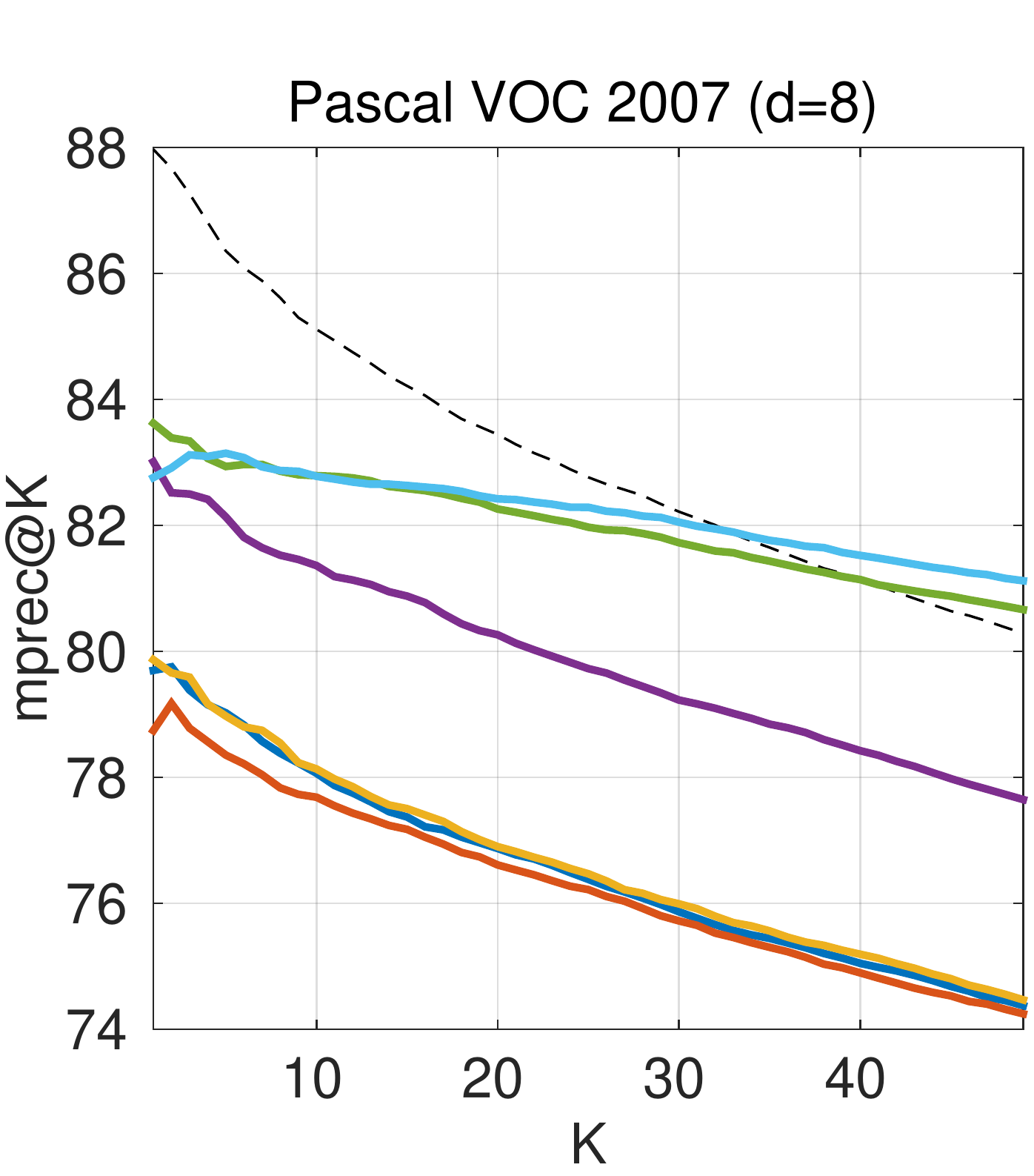}   
\includegraphics[width=0.245\textwidth,trim=20 180 190 155,clip]{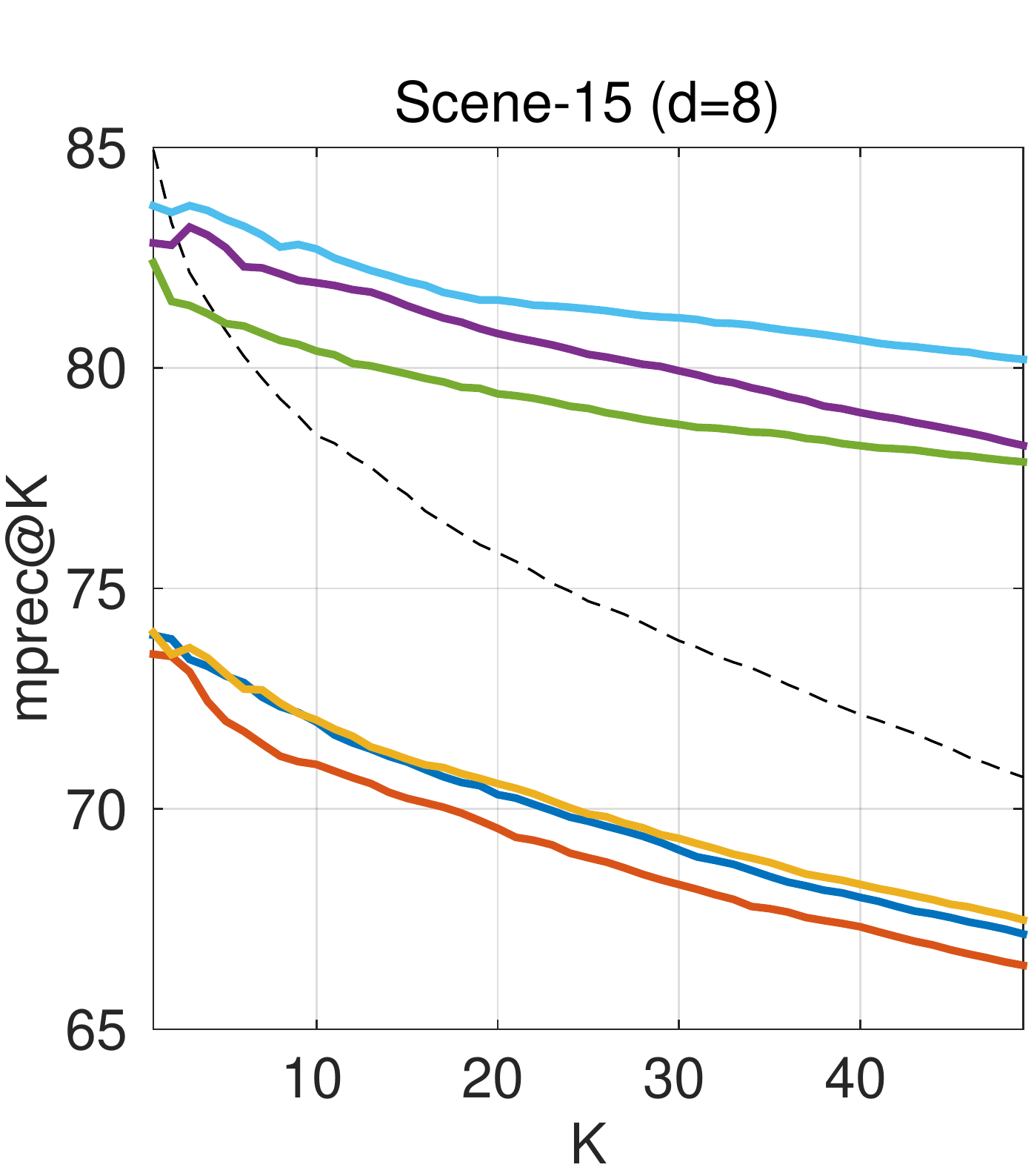}   
\includegraphics[width=0.245\textwidth,trim=20 180 190 155,clip]{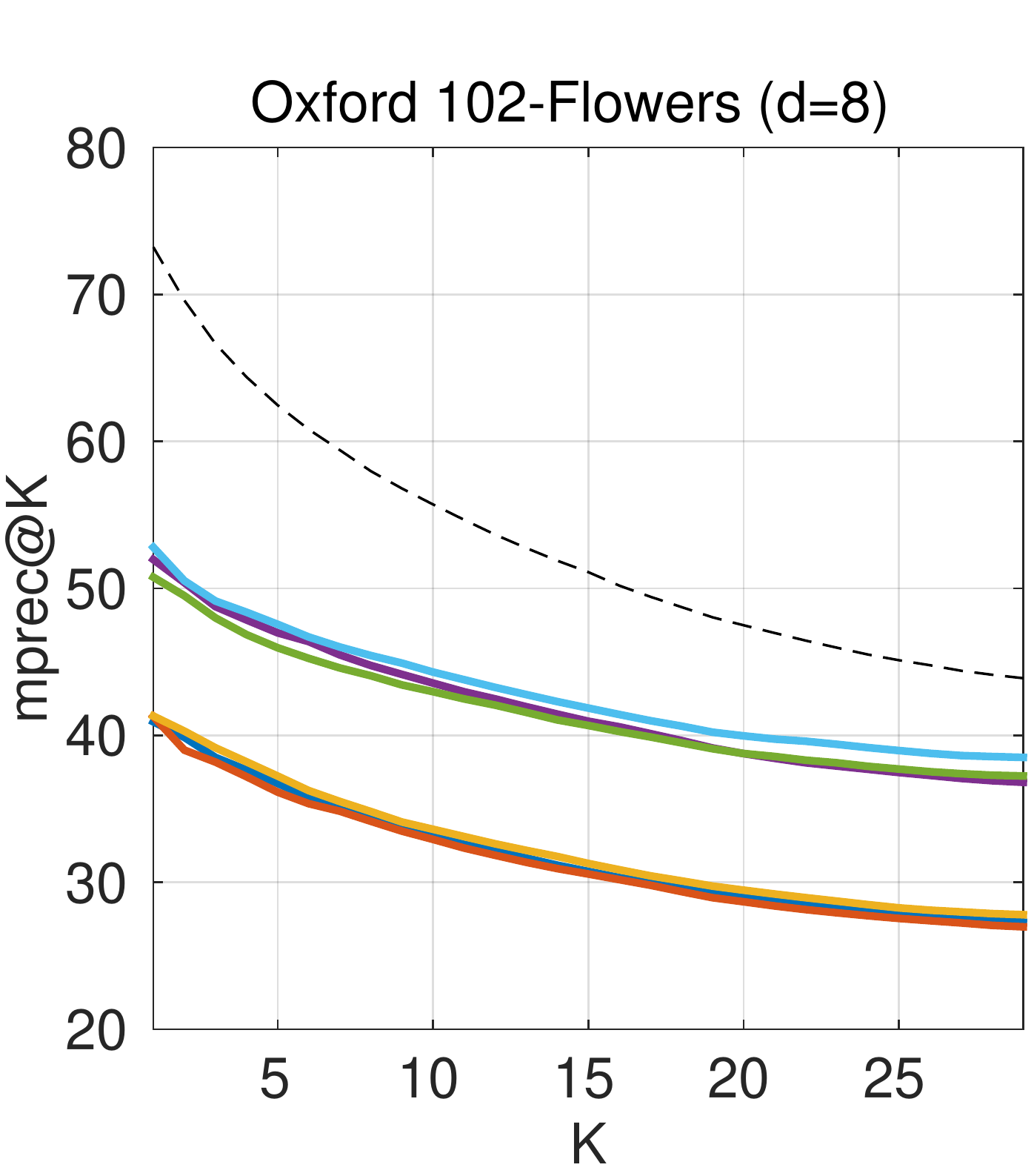}   
\includegraphics[width=0.245\textwidth,trim=20 180 190 155,clip]{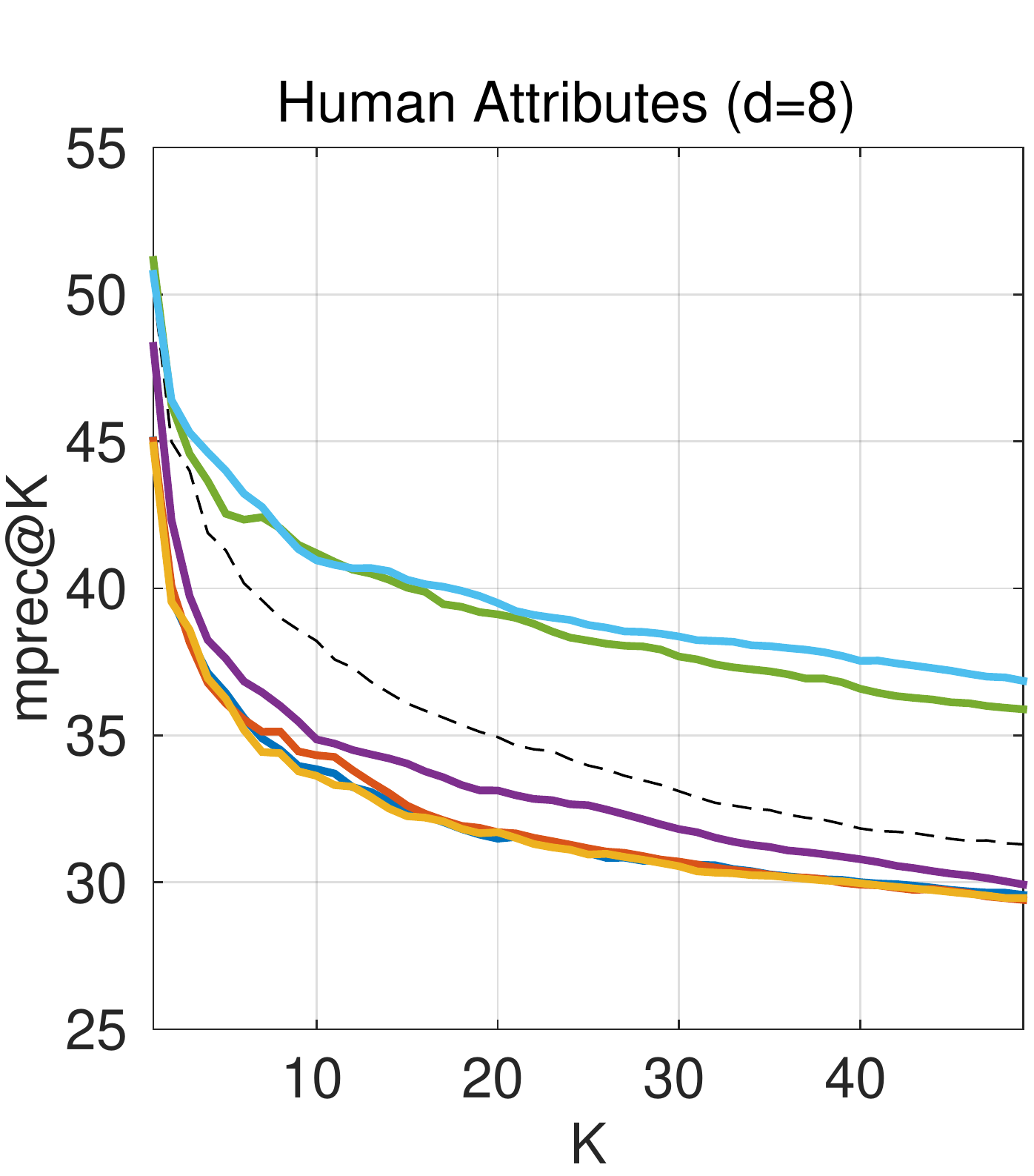}   
\includegraphics[width=0.335\textwidth,trim=20 180  50 155,clip]{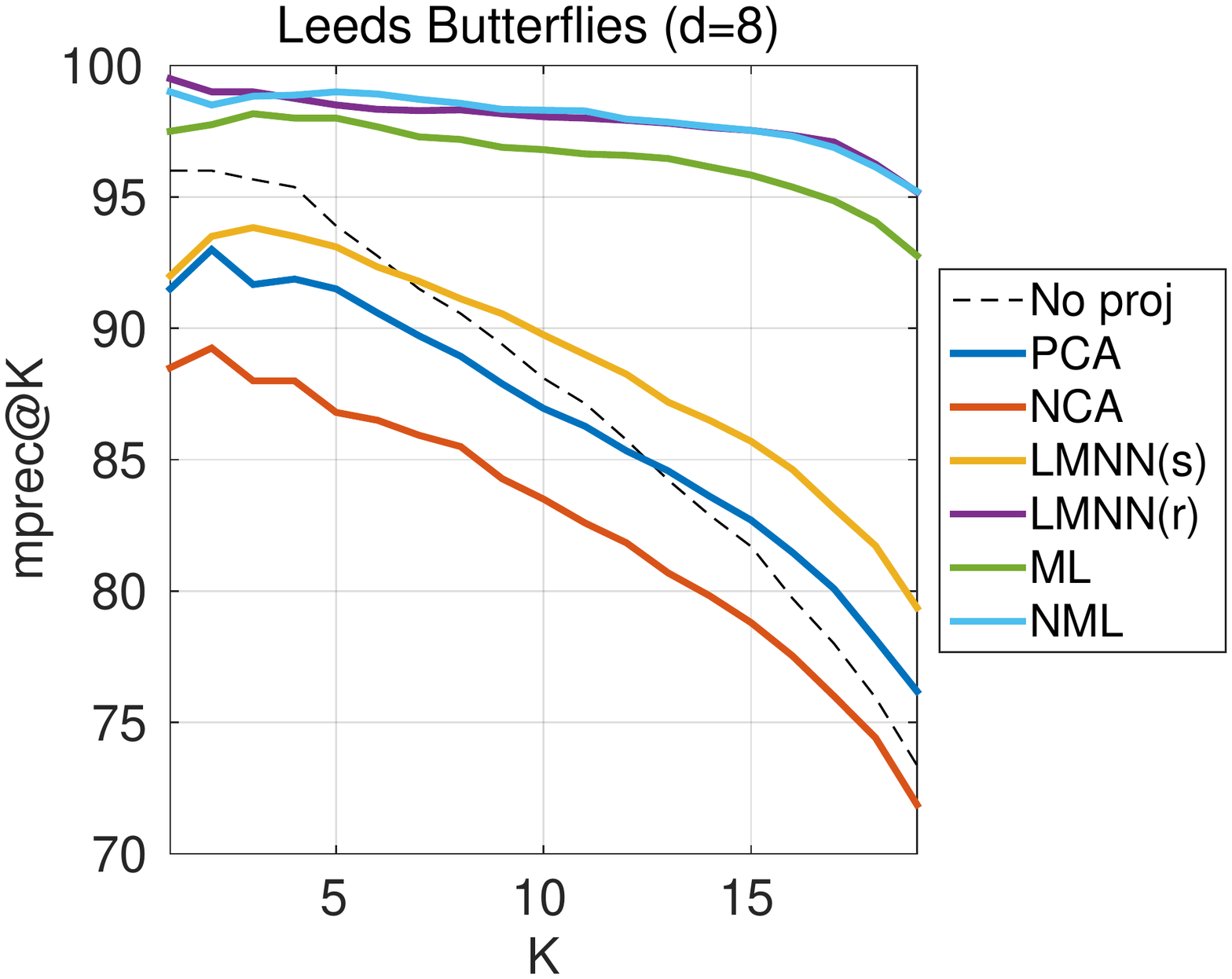}   
\caption{
Comparisons of methods on the different datasets for $K \in [1,\max(50,n^+-1)]$ and $d=8$.
}
\label{supp_fig:prec_k_d8}
\end{figure*}

\begin{figure*}
\centering
\includegraphics[width=0.245\textwidth,trim=20 180 190 155,clip]{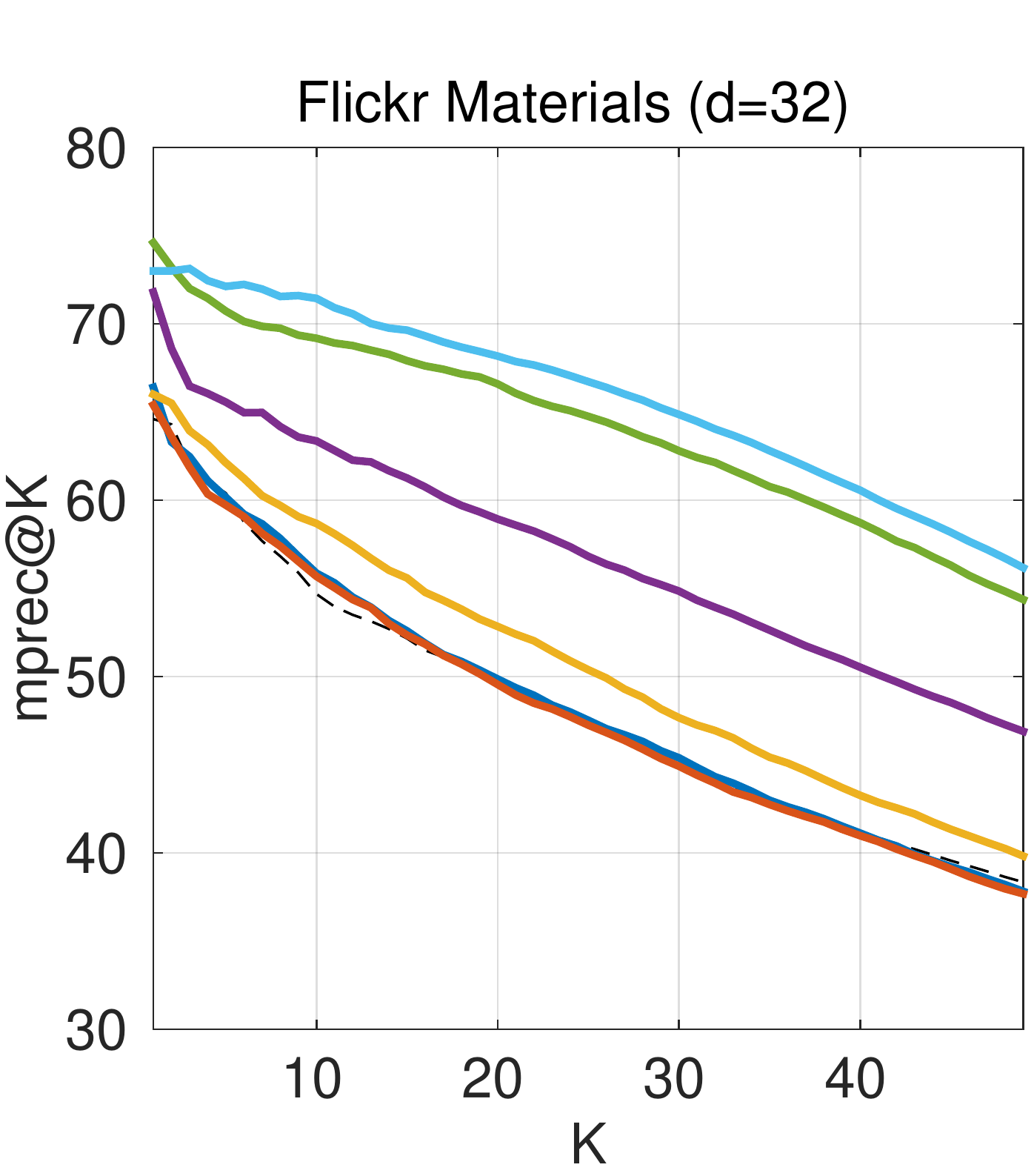}   
\includegraphics[width=0.245\textwidth,trim=20 180 190 155,clip]{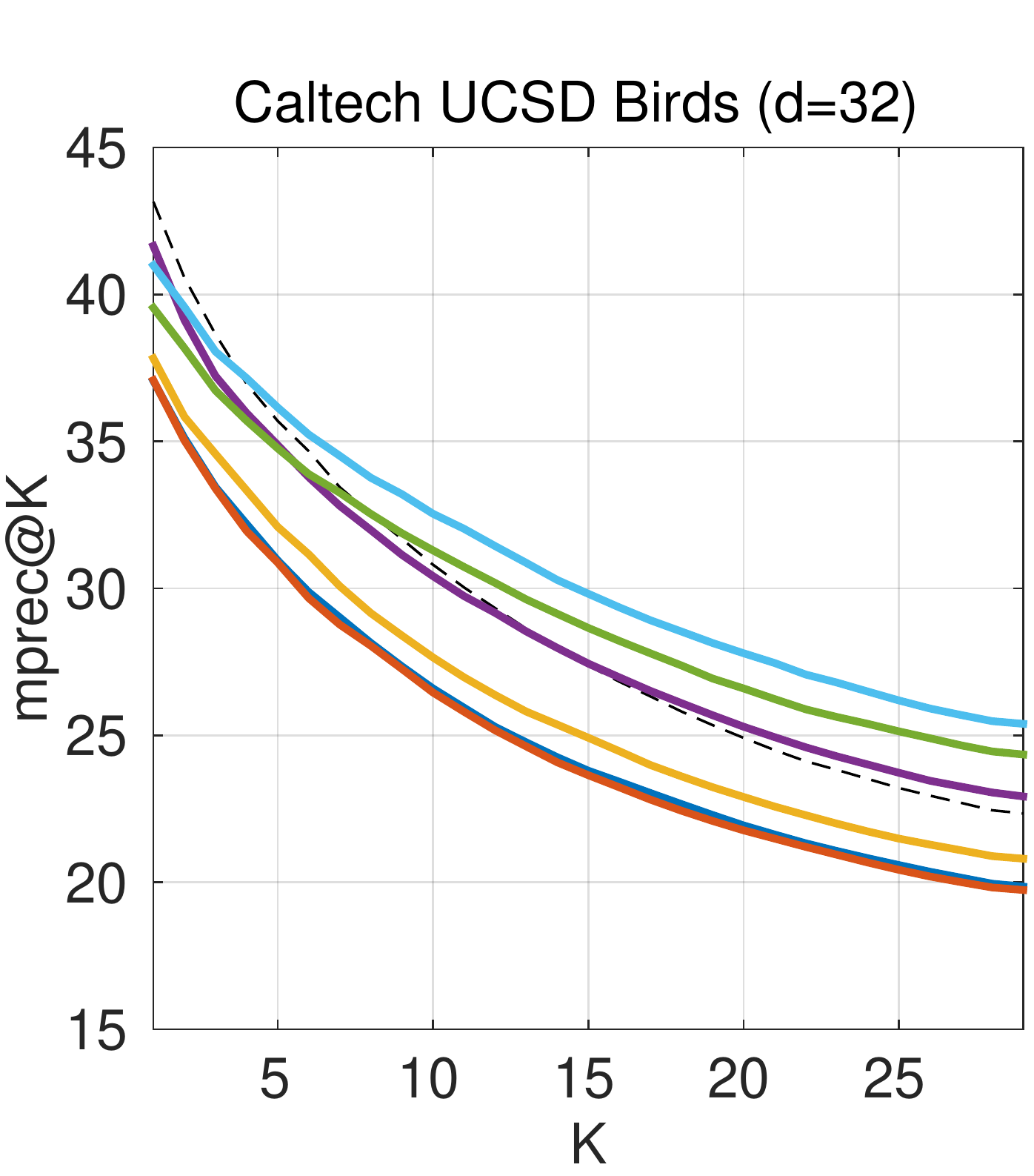}   
\includegraphics[width=0.245\textwidth,trim=20 180 190 155,clip]{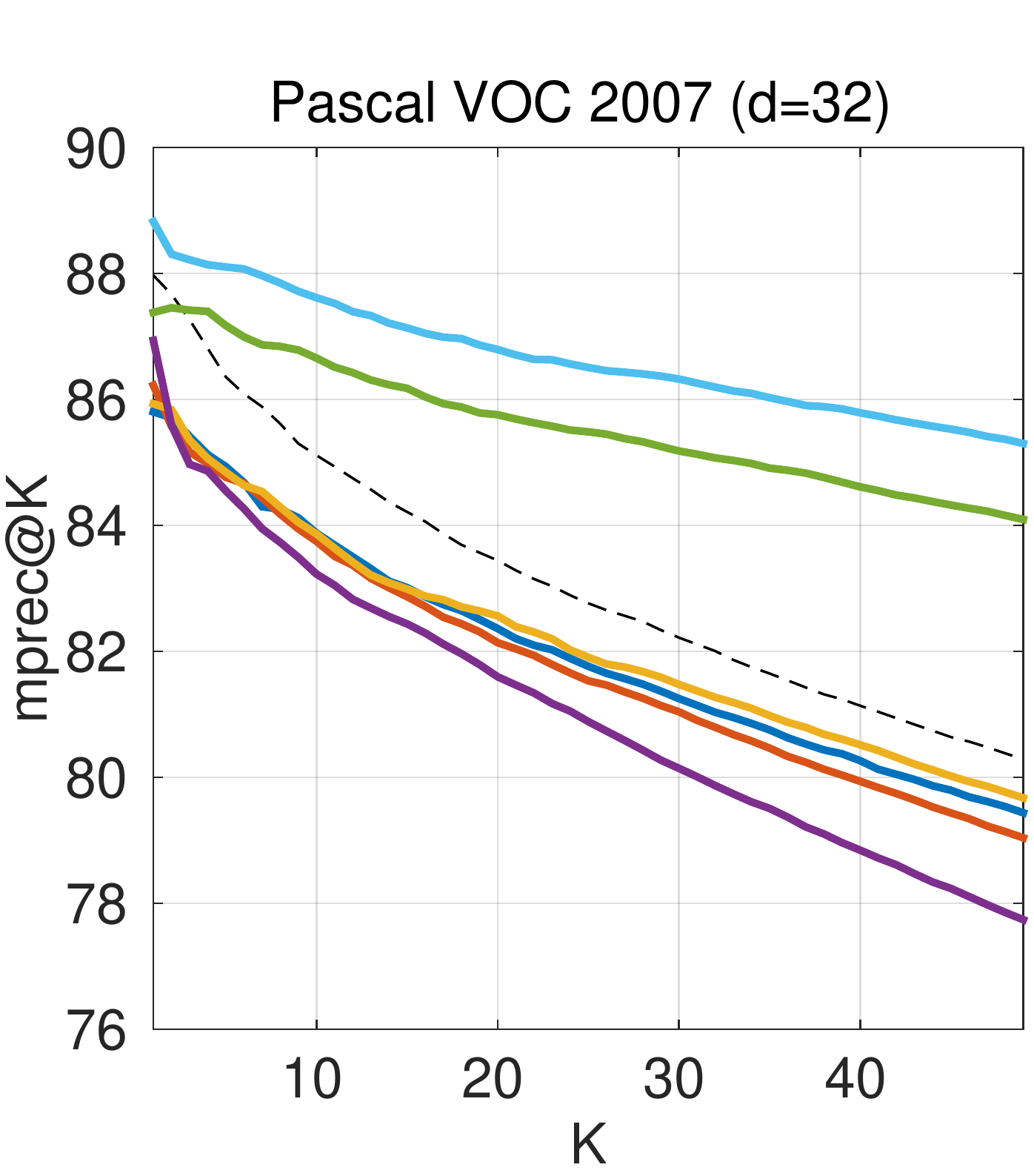}   
\includegraphics[width=0.245\textwidth,trim=20 180 190 155,clip]{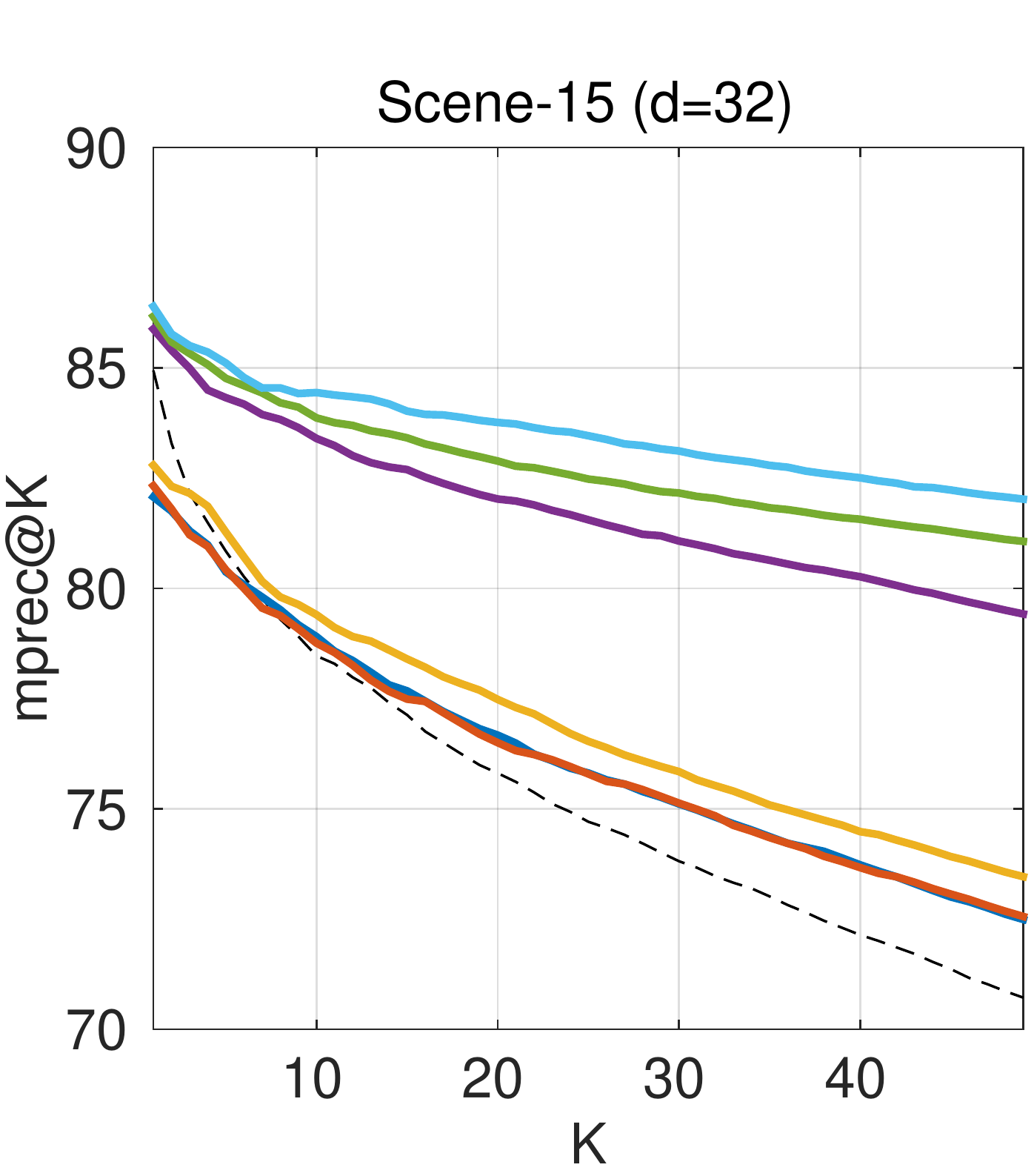}   
\includegraphics[width=0.245\textwidth,trim=20 180 190 155,clip]{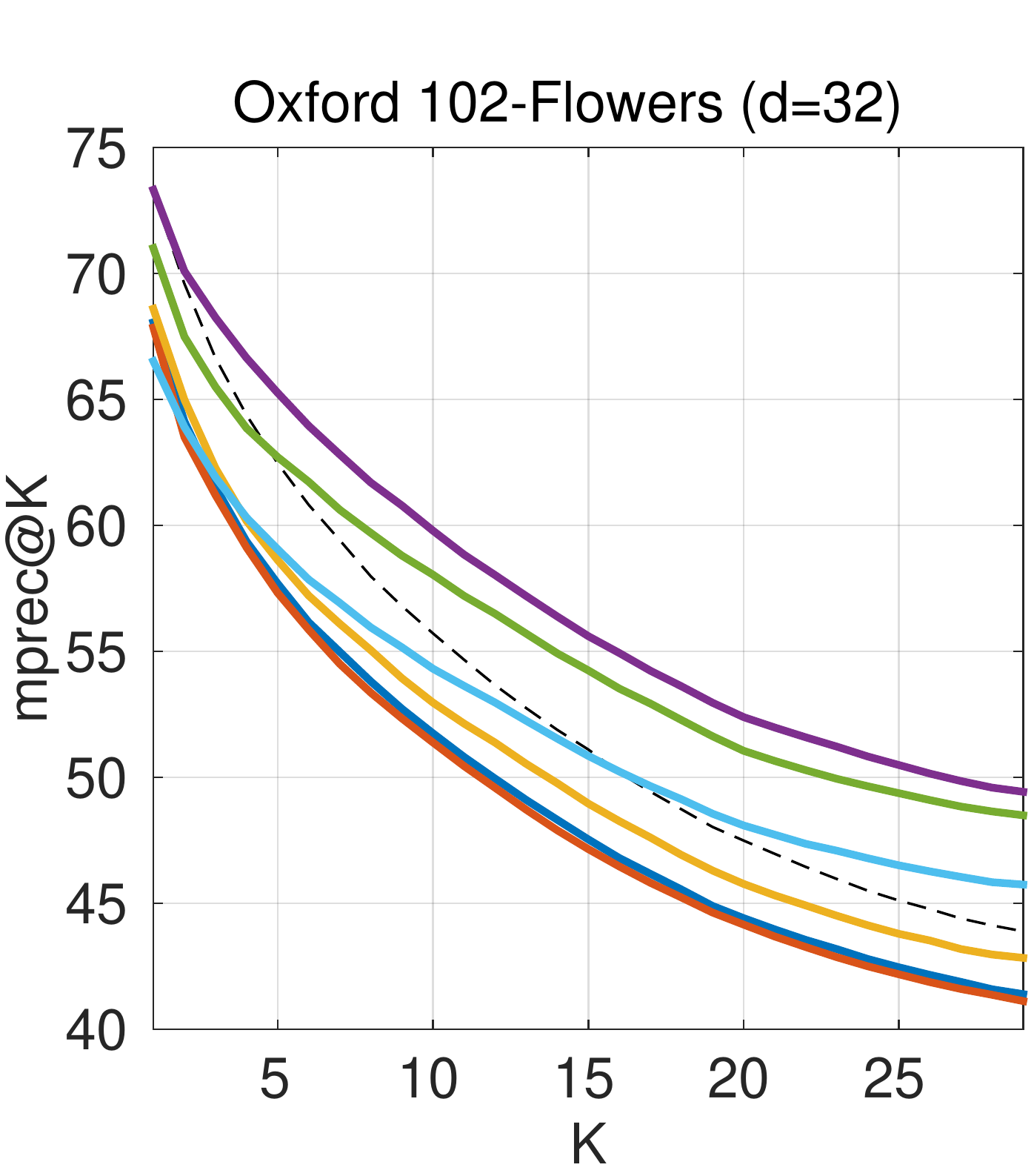}   
\includegraphics[width=0.245\textwidth,trim=20 180 190 155,clip]{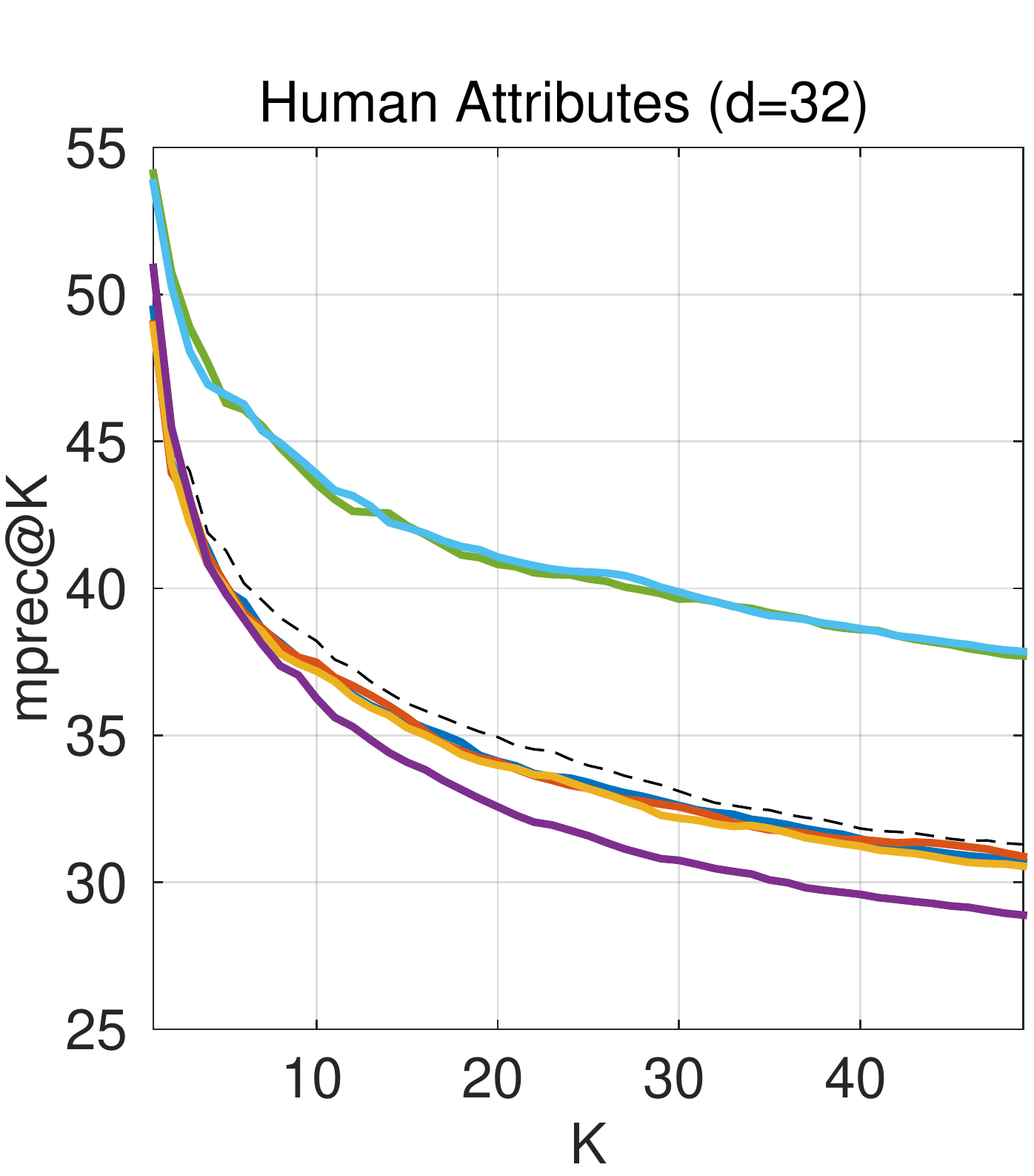}   
\includegraphics[width=0.335\textwidth,trim=20 180  50 155,clip]{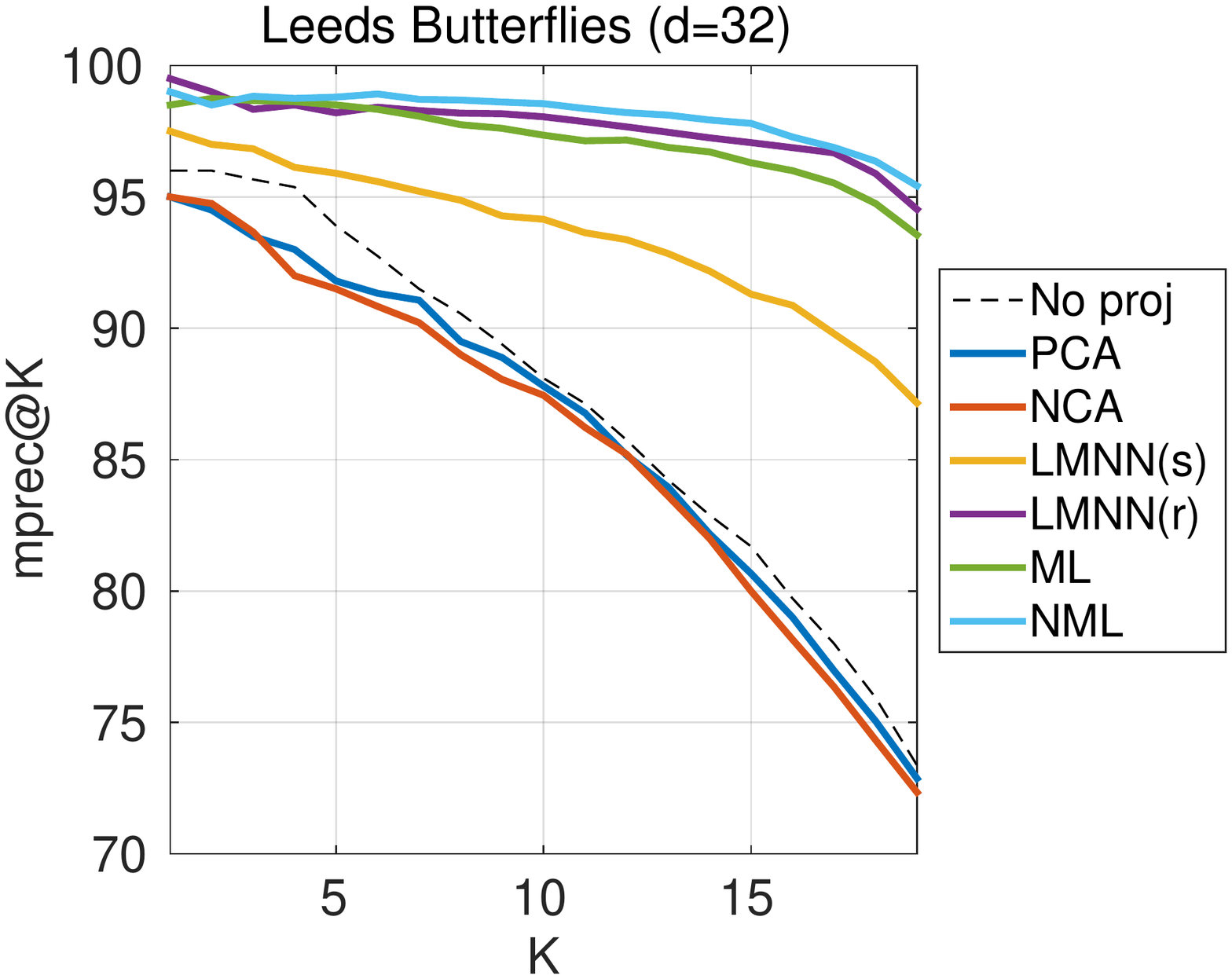}   
\caption{
Comparisons of methods on the different datasets for $K \in [1,\max(50,n^+-1)]$ and $d=32$.
}
\label{supp_fig:prec_k_d32}
\end{figure*}

\begin{figure*}
\centering
\includegraphics[width=0.245\textwidth,trim=20 180 190 155,clip]{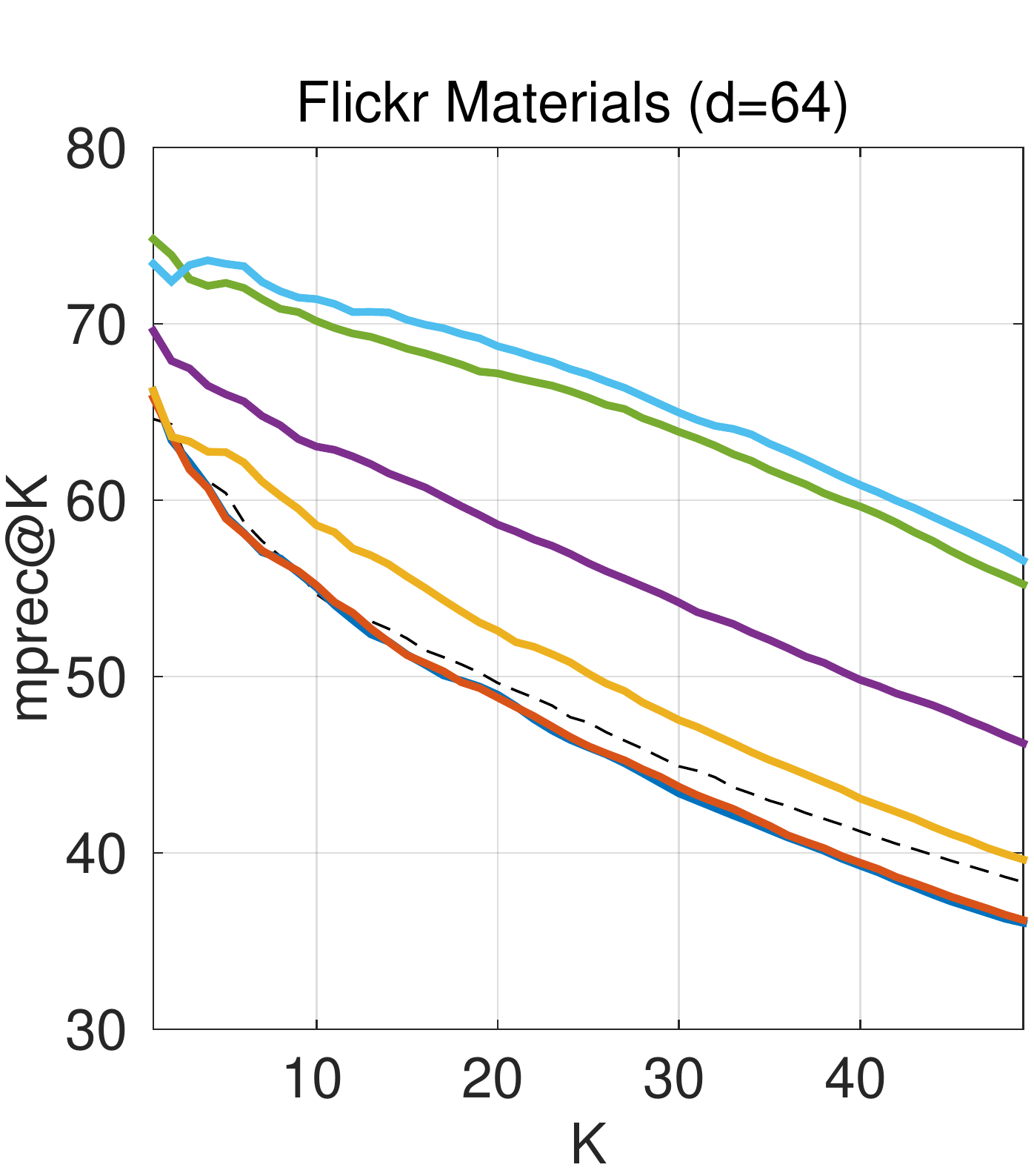}   
\includegraphics[width=0.245\textwidth,trim=20 180 190 155,clip]{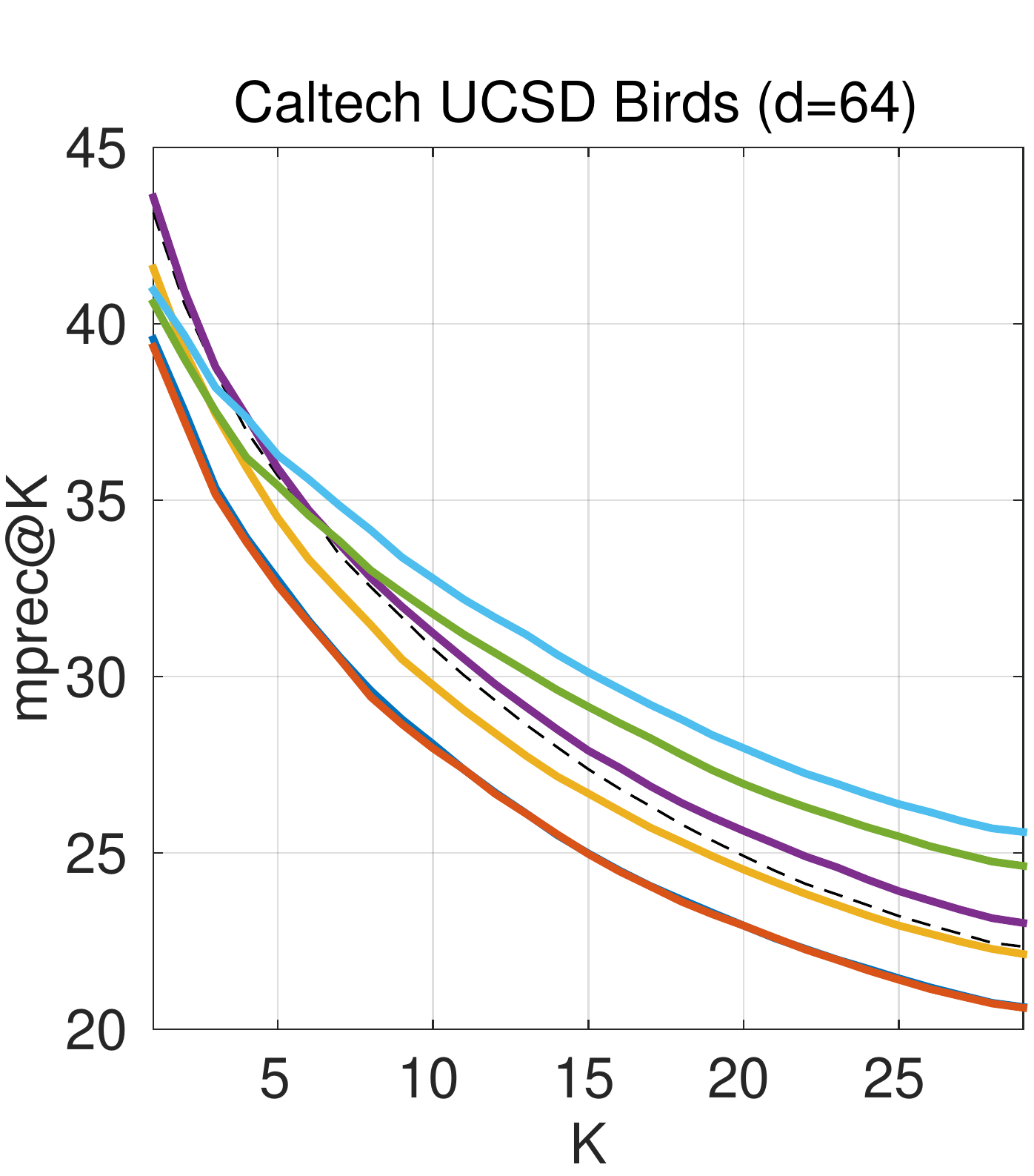}   
\includegraphics[width=0.245\textwidth,trim=20 180 190 155,clip]{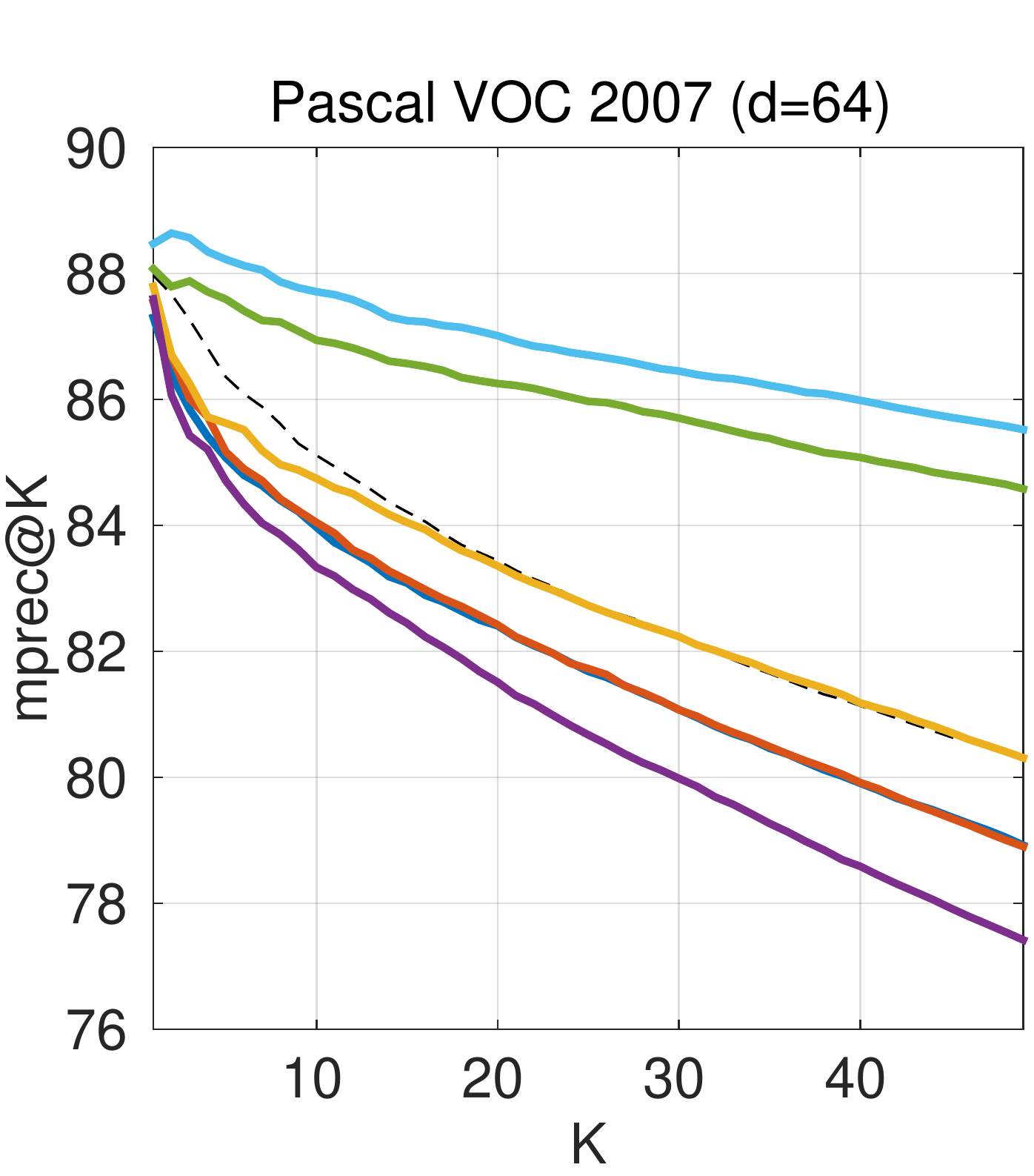}   
\includegraphics[width=0.245\textwidth,trim=20 180 190 155,clip]{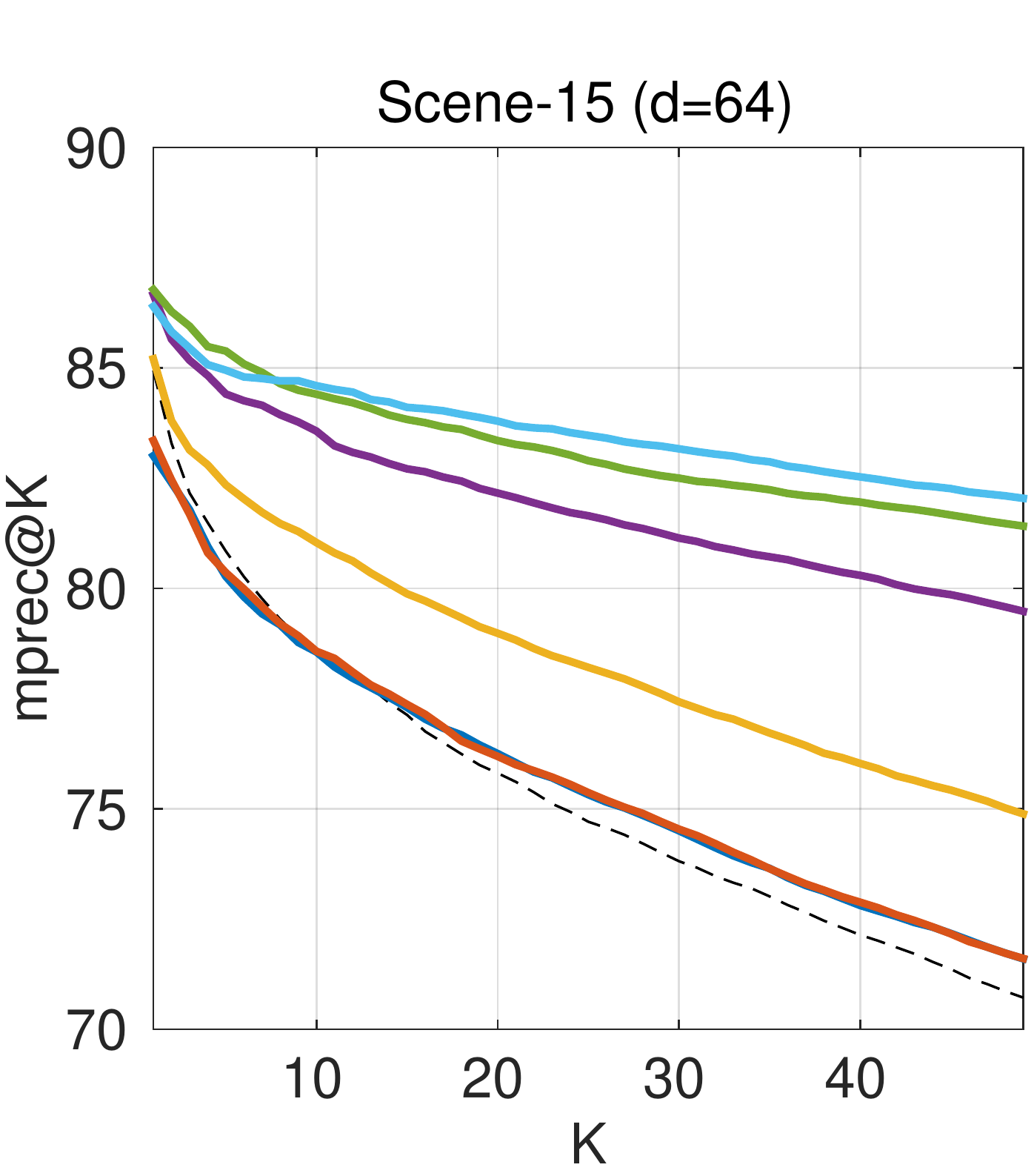}   
\includegraphics[width=0.245\textwidth,trim=20 180 190 155,clip]{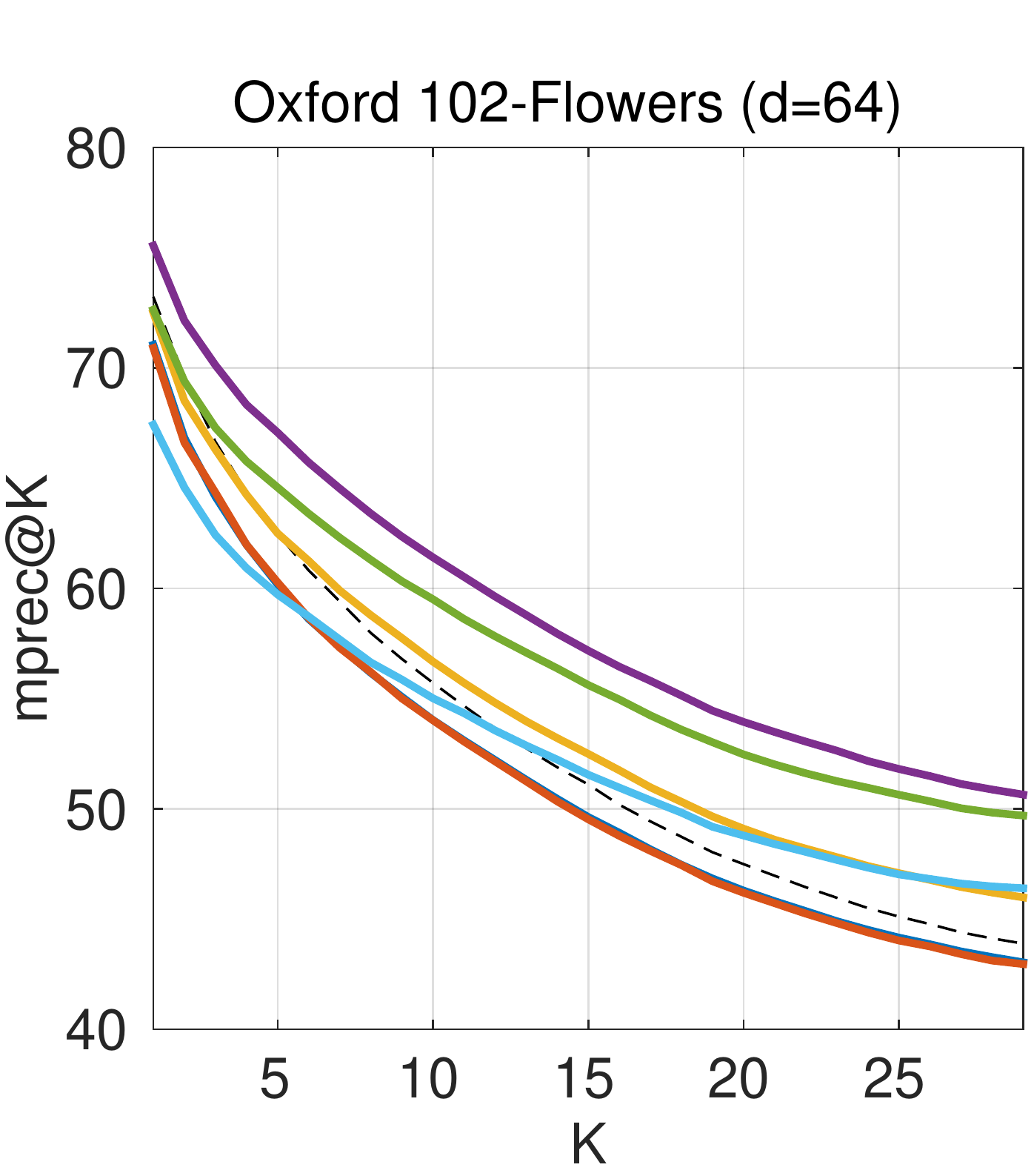}   
\includegraphics[width=0.245\textwidth,trim=20 180 190 155,clip]{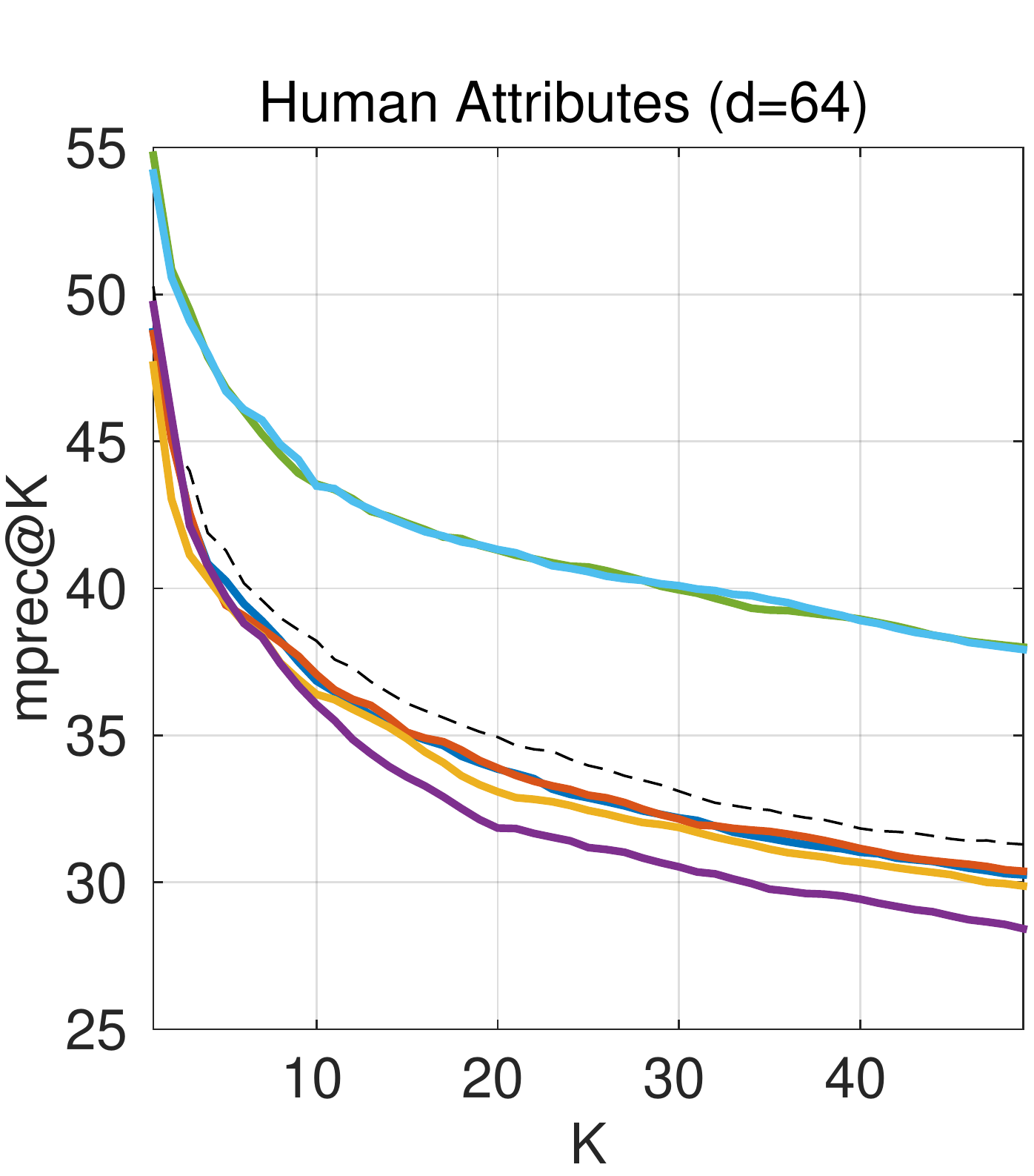}   
\includegraphics[width=0.335\textwidth,trim=20 180  50 155,clip]{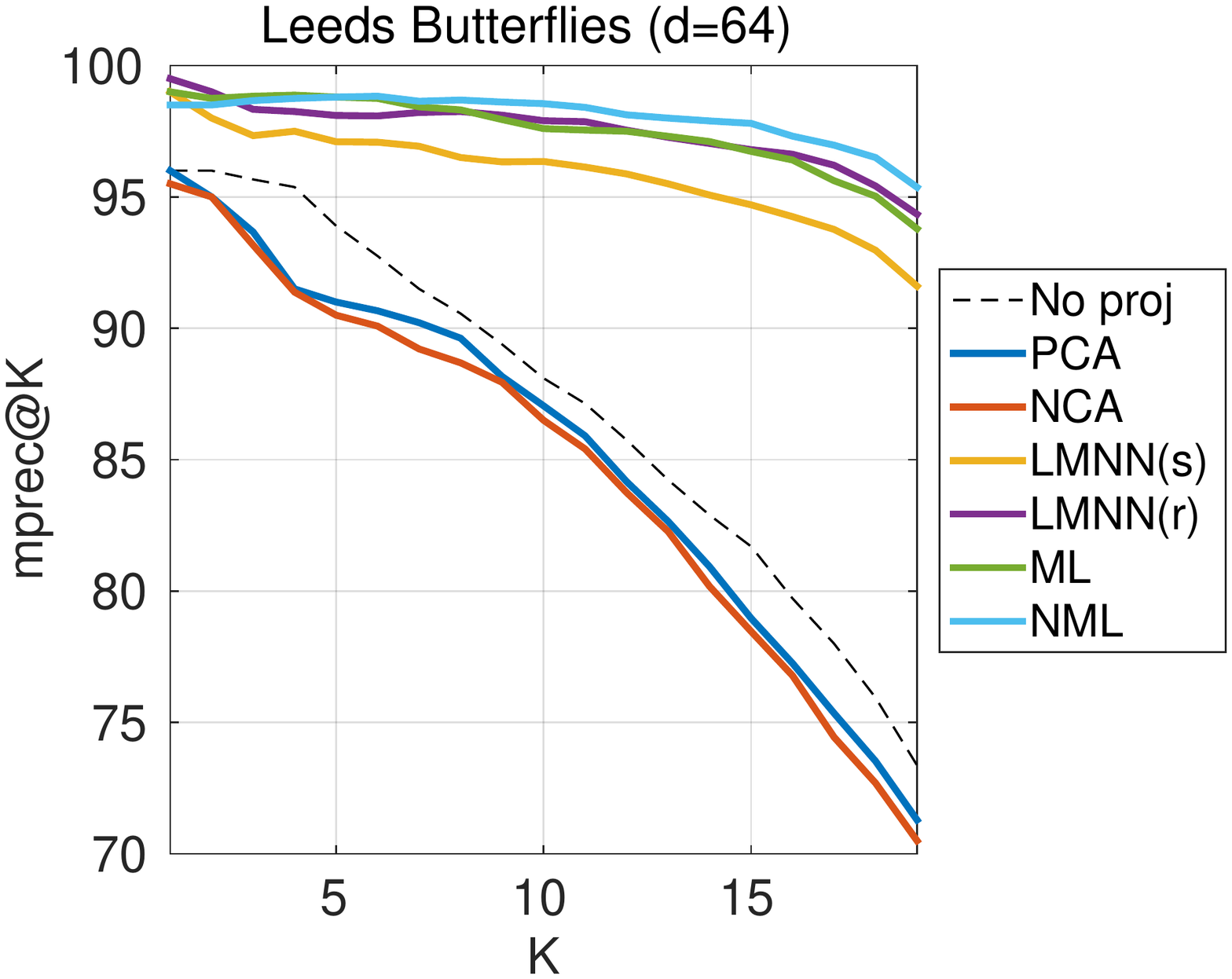}   
\caption{
Comparisons of methods on the different datasets for $K \in [1,\max(50,n^+-1)]$ and $d=64$.
}
\label{supp_fig:prec_k_d64}
\end{figure*}

\begin{figure*}
\centering
\includegraphics[width=0.245\textwidth,trim=20 180 190 155,clip]{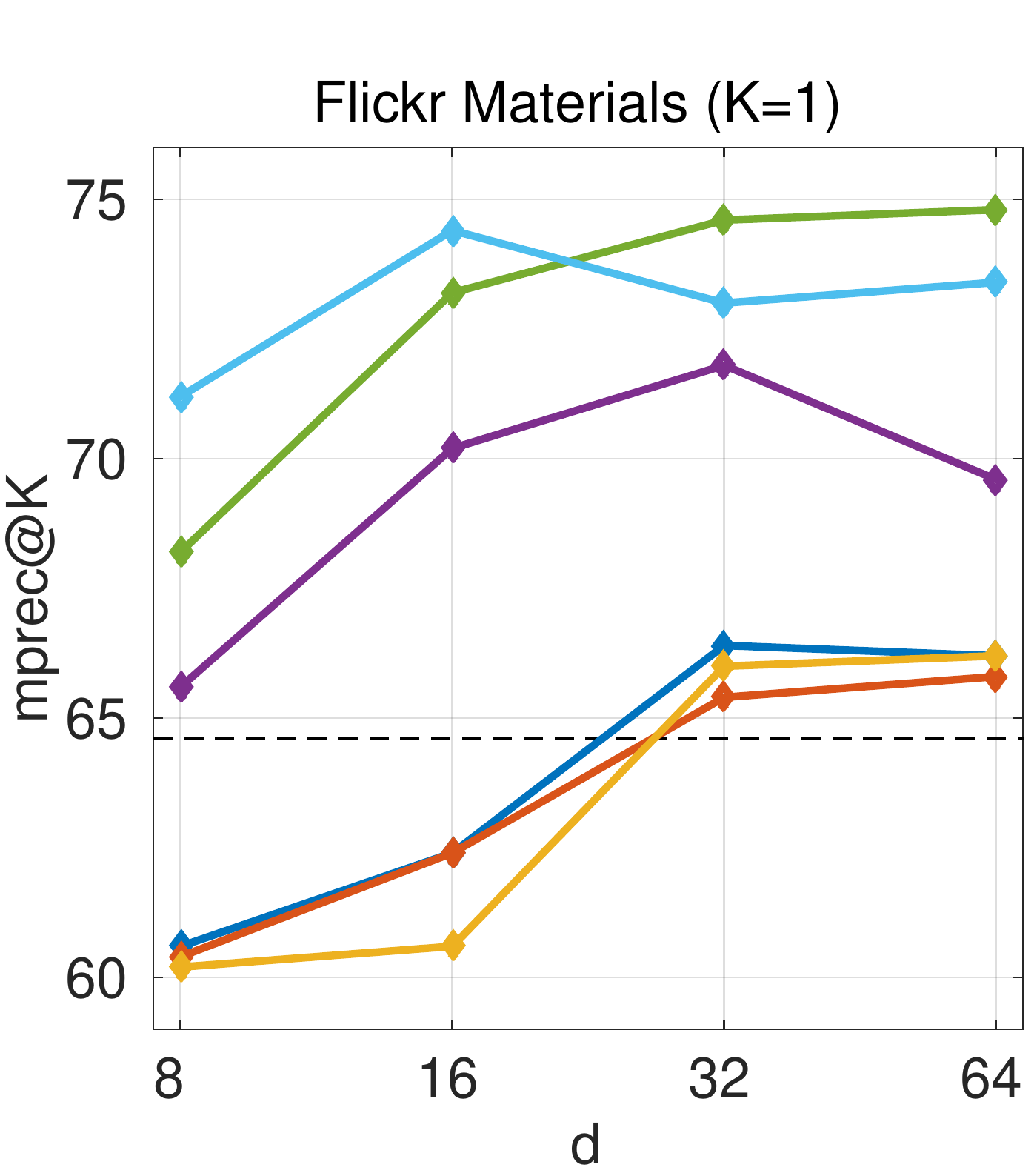}   
\includegraphics[width=0.245\textwidth,trim=20 180 190 155,clip]{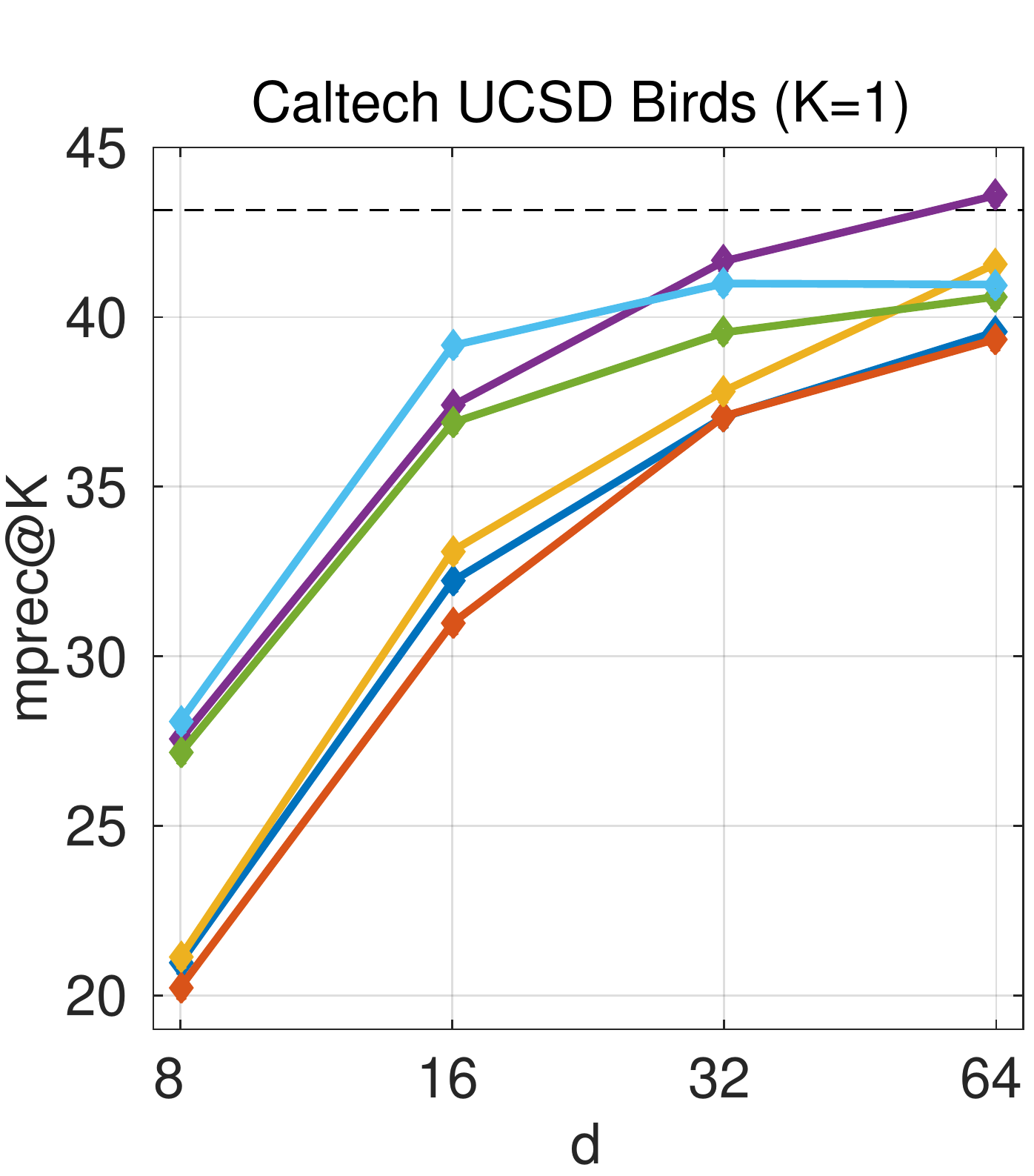}   
\includegraphics[width=0.245\textwidth,trim=20 180 190 155,clip]{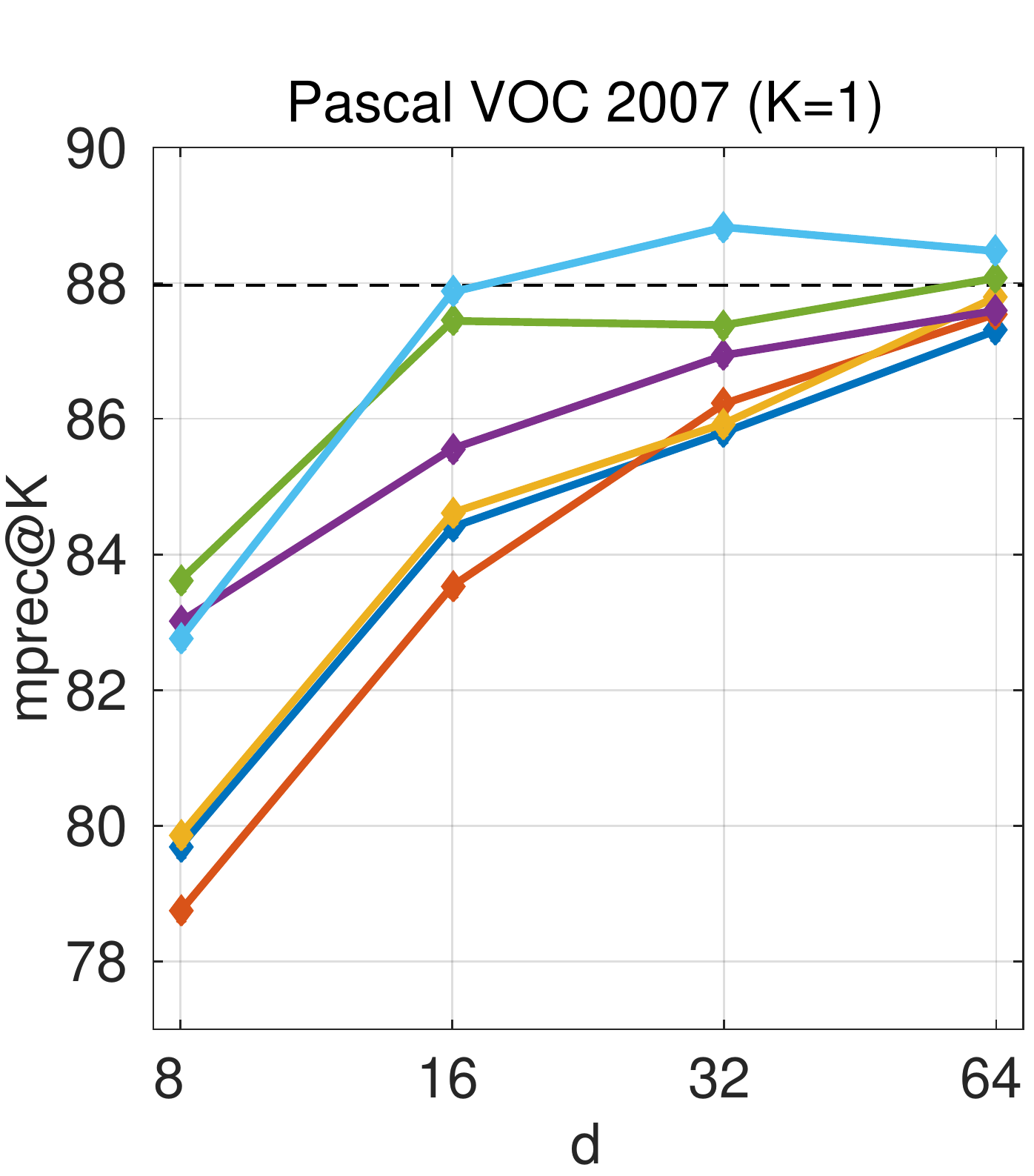}   
\includegraphics[width=0.245\textwidth,trim=20 180 190 155,clip]{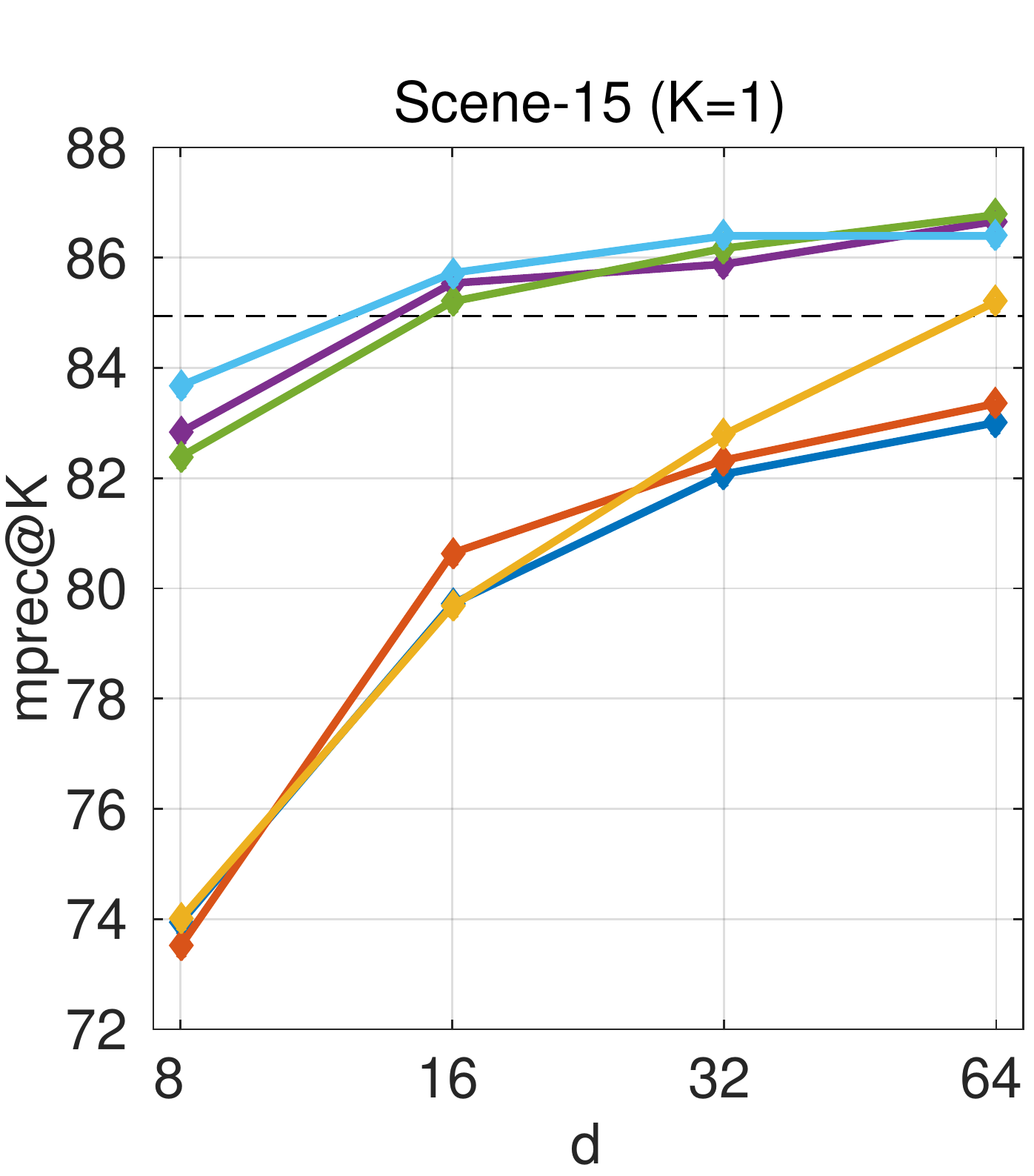}   
\includegraphics[width=0.245\textwidth,trim=20 180 190 155,clip]{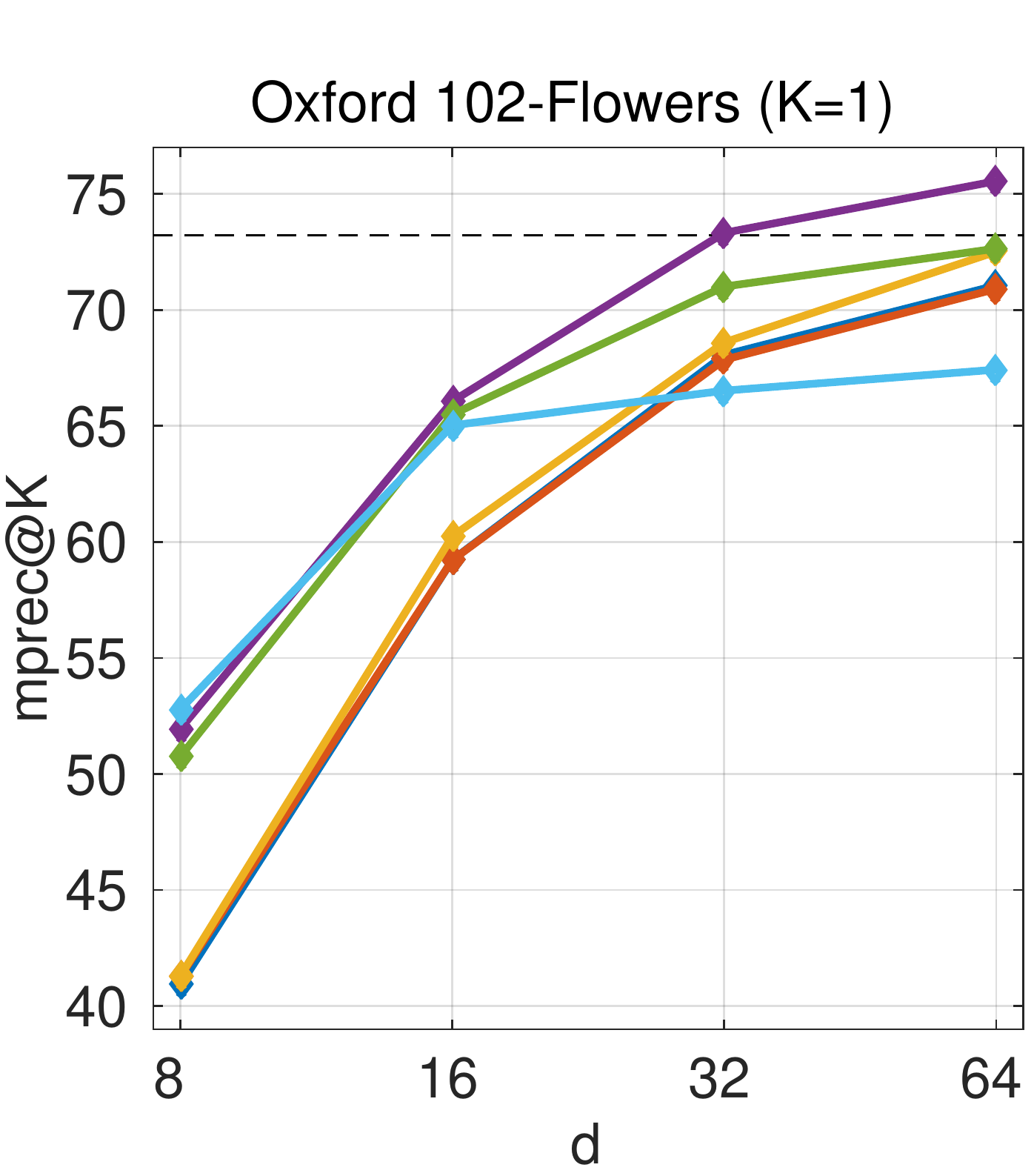}   
\includegraphics[width=0.245\textwidth,trim=20 180 190 155,clip]{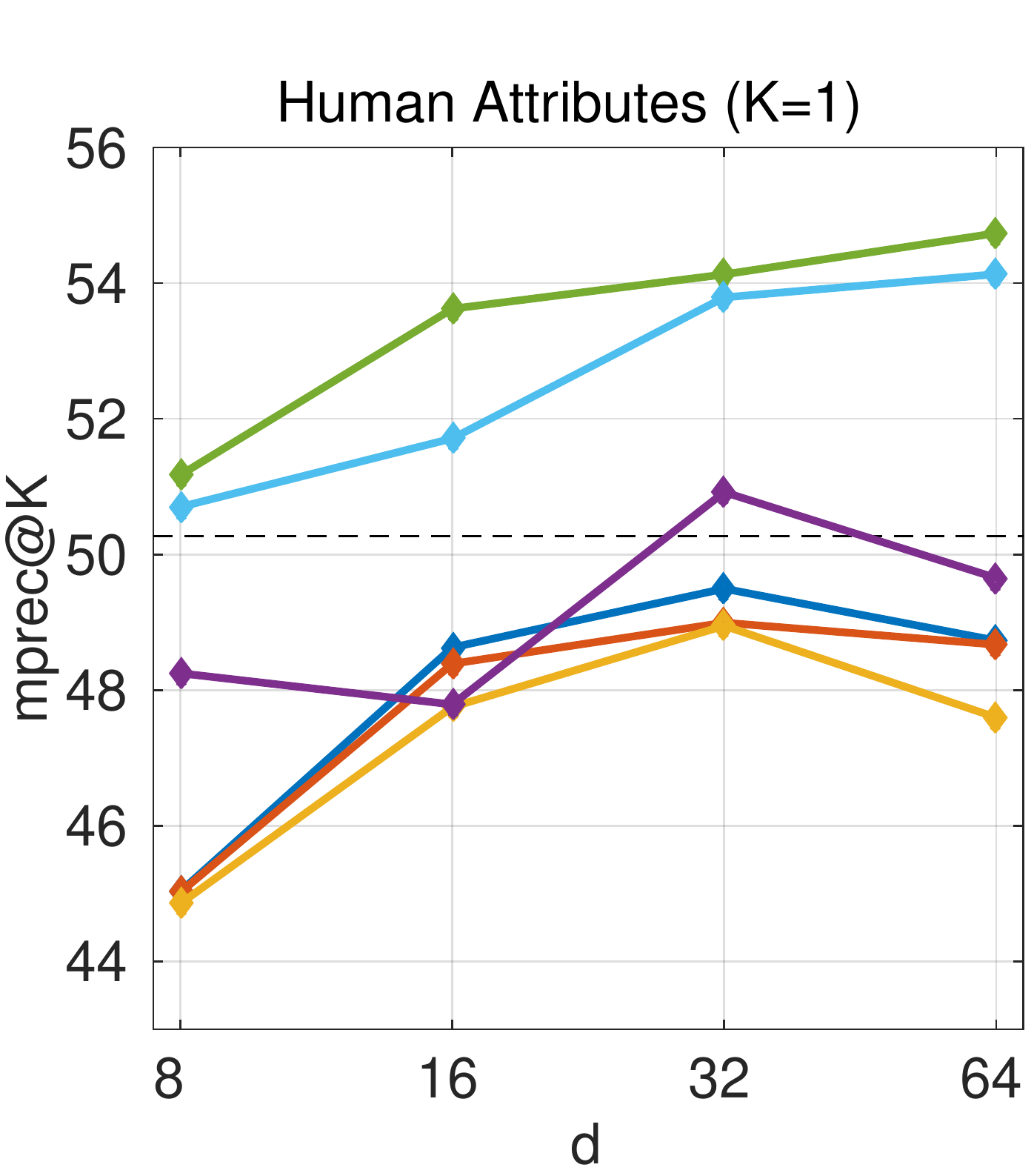}   
\includegraphics[width=0.335\textwidth,trim=20 180  50 155,clip]{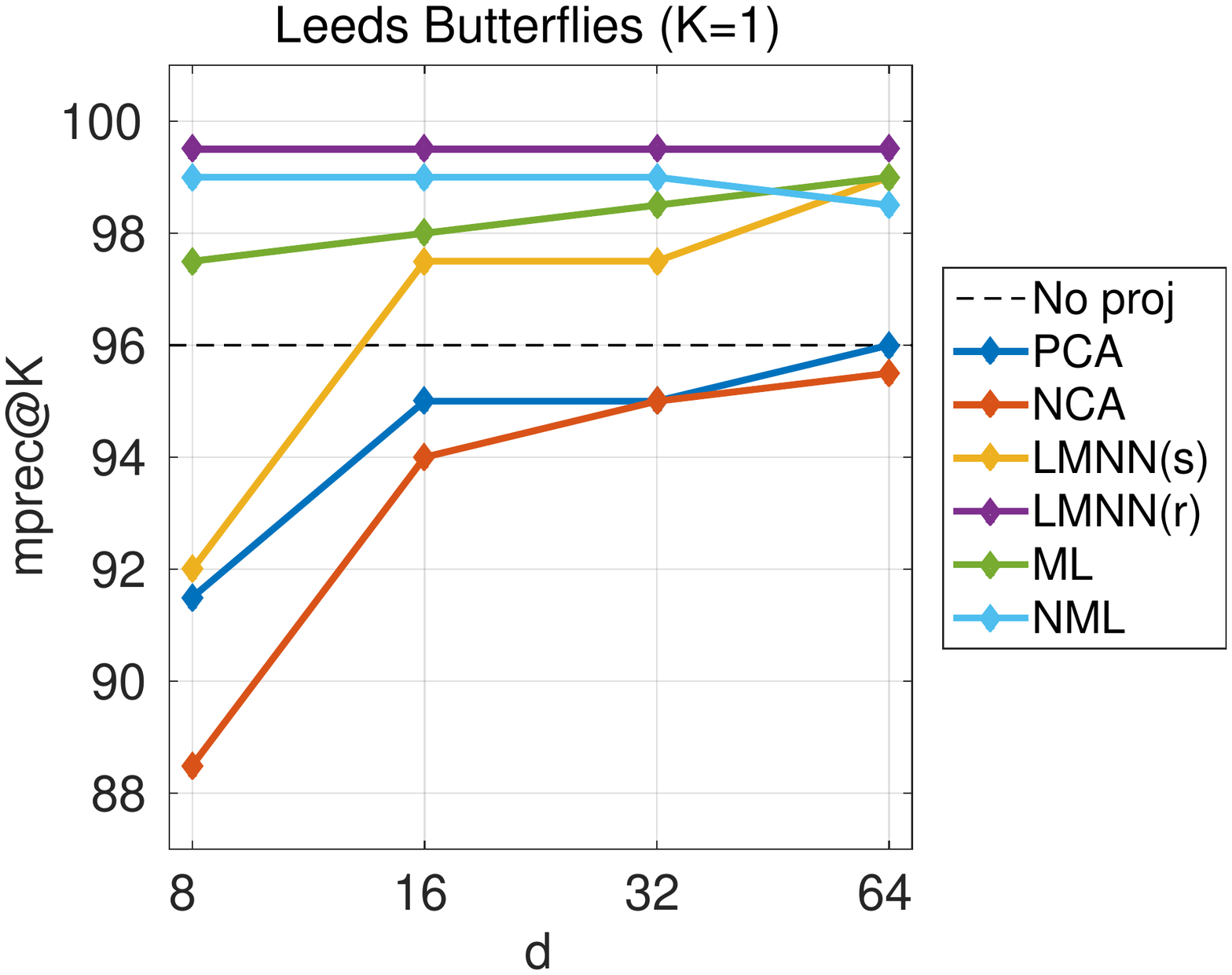}   
\caption{
Comparisons of methods on the different datasets for $K=1$ and $d \in \{8,16,32,64\}$.
}
\label{supp_fig:prec_d_k1}
\end{figure*}

\begin{figure*}
\centering
\includegraphics[width=0.245\textwidth,trim=20 180 190 155,clip]{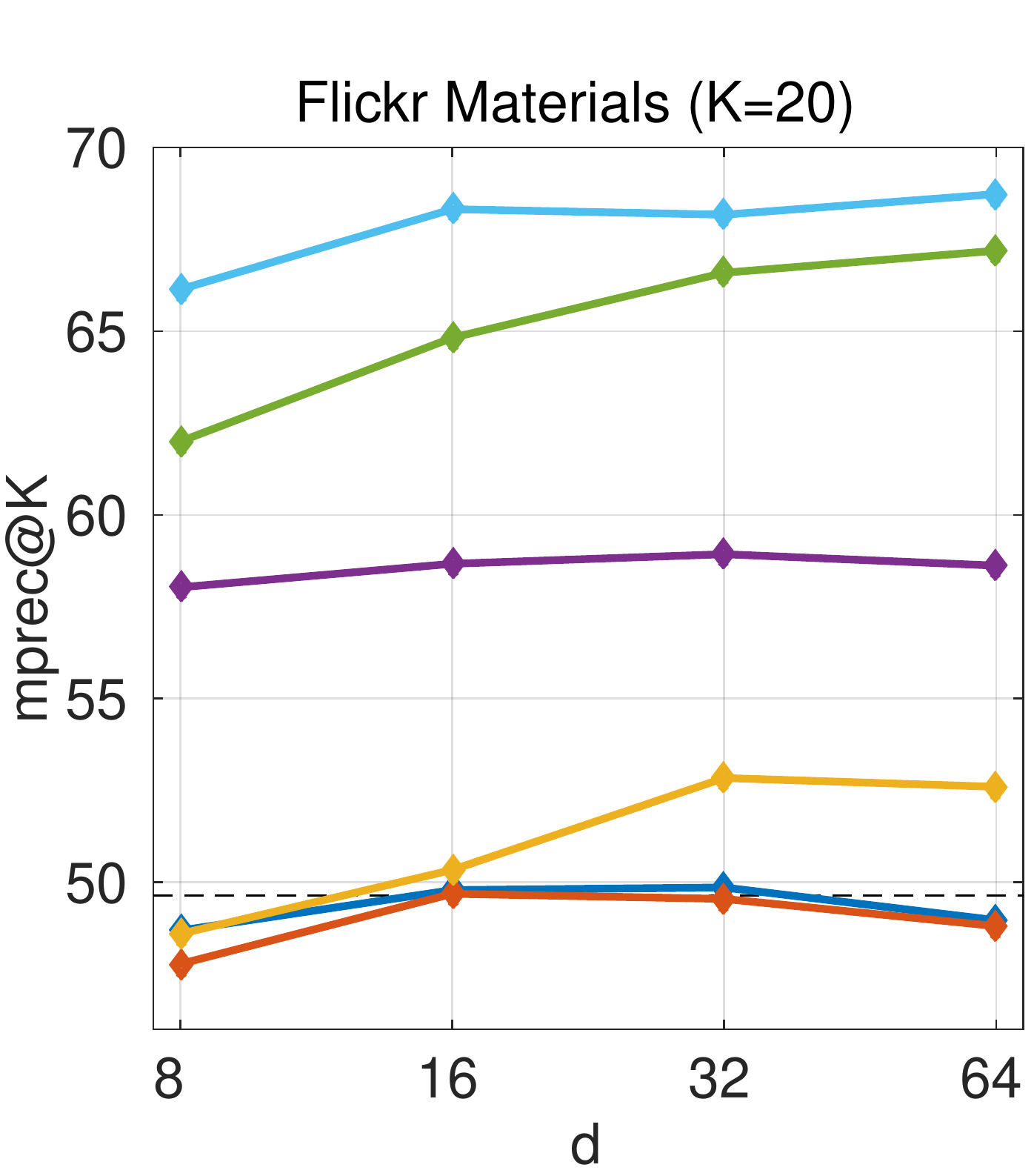}   
\includegraphics[width=0.245\textwidth,trim=20 180 190 155,clip]{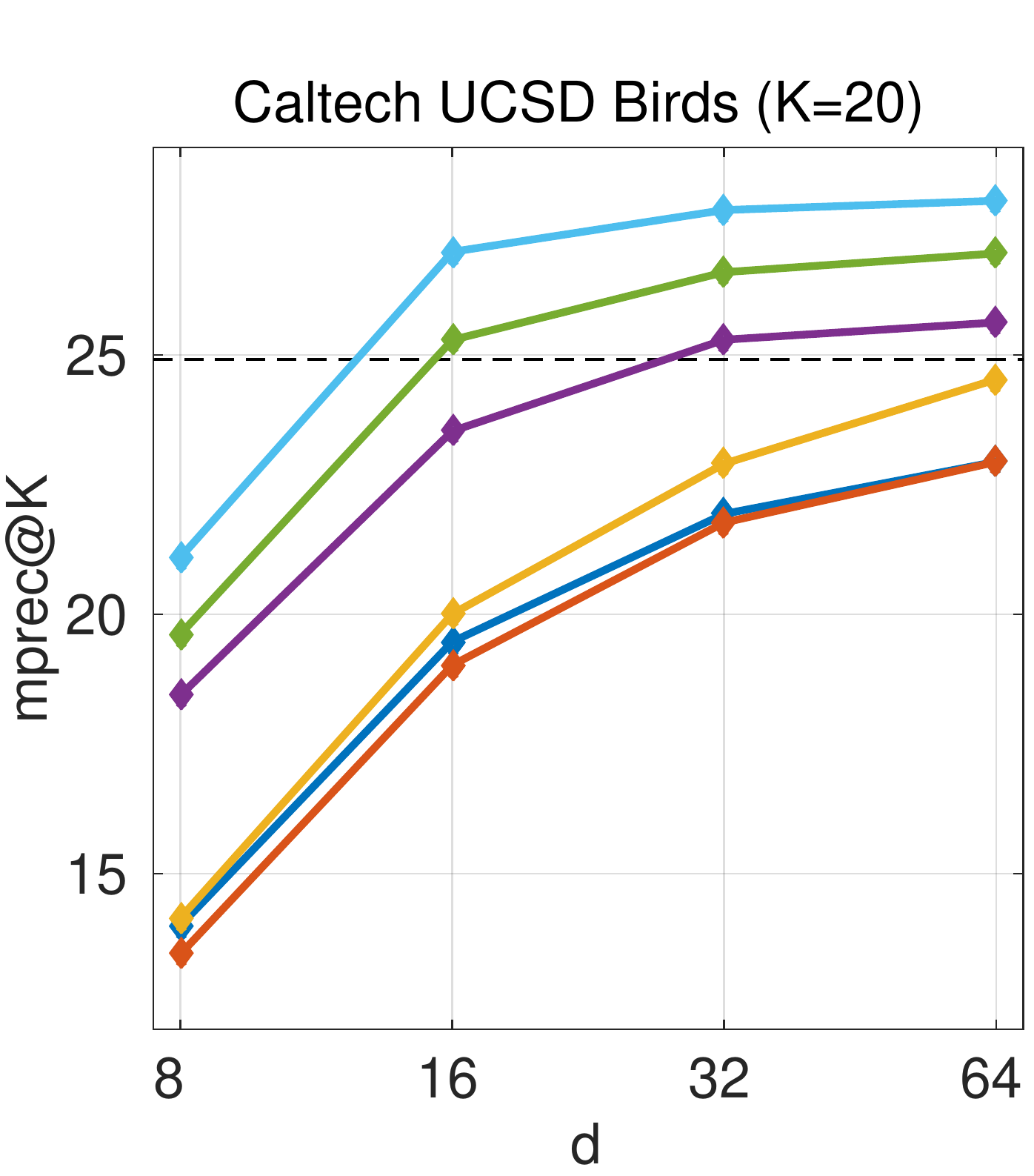}   
\includegraphics[width=0.245\textwidth,trim=20 180 190 155,clip]{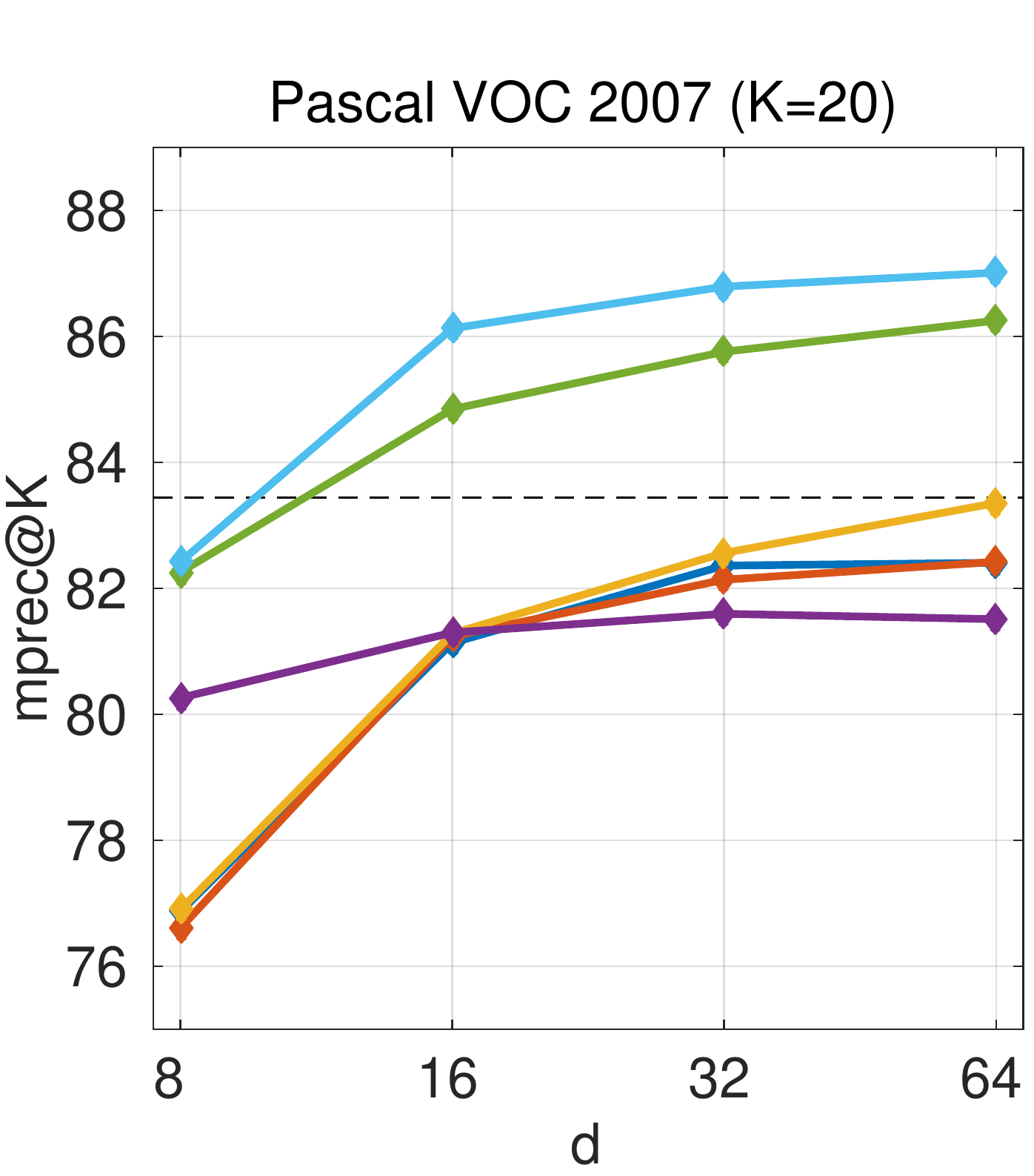}   
\includegraphics[width=0.245\textwidth,trim=20 180 190 155,clip]{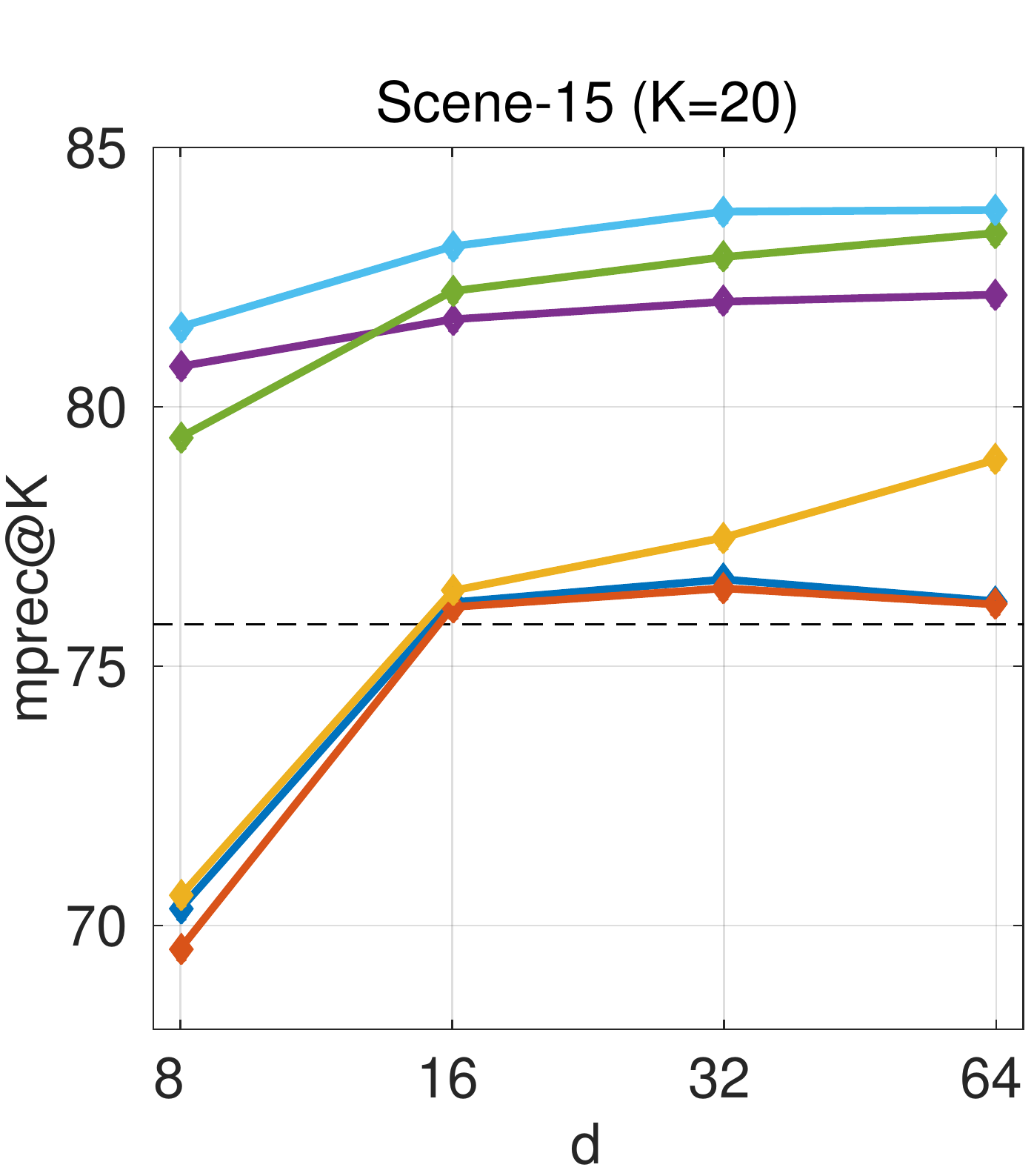}   
\includegraphics[width=0.245\textwidth,trim=20 180 190 155,clip]{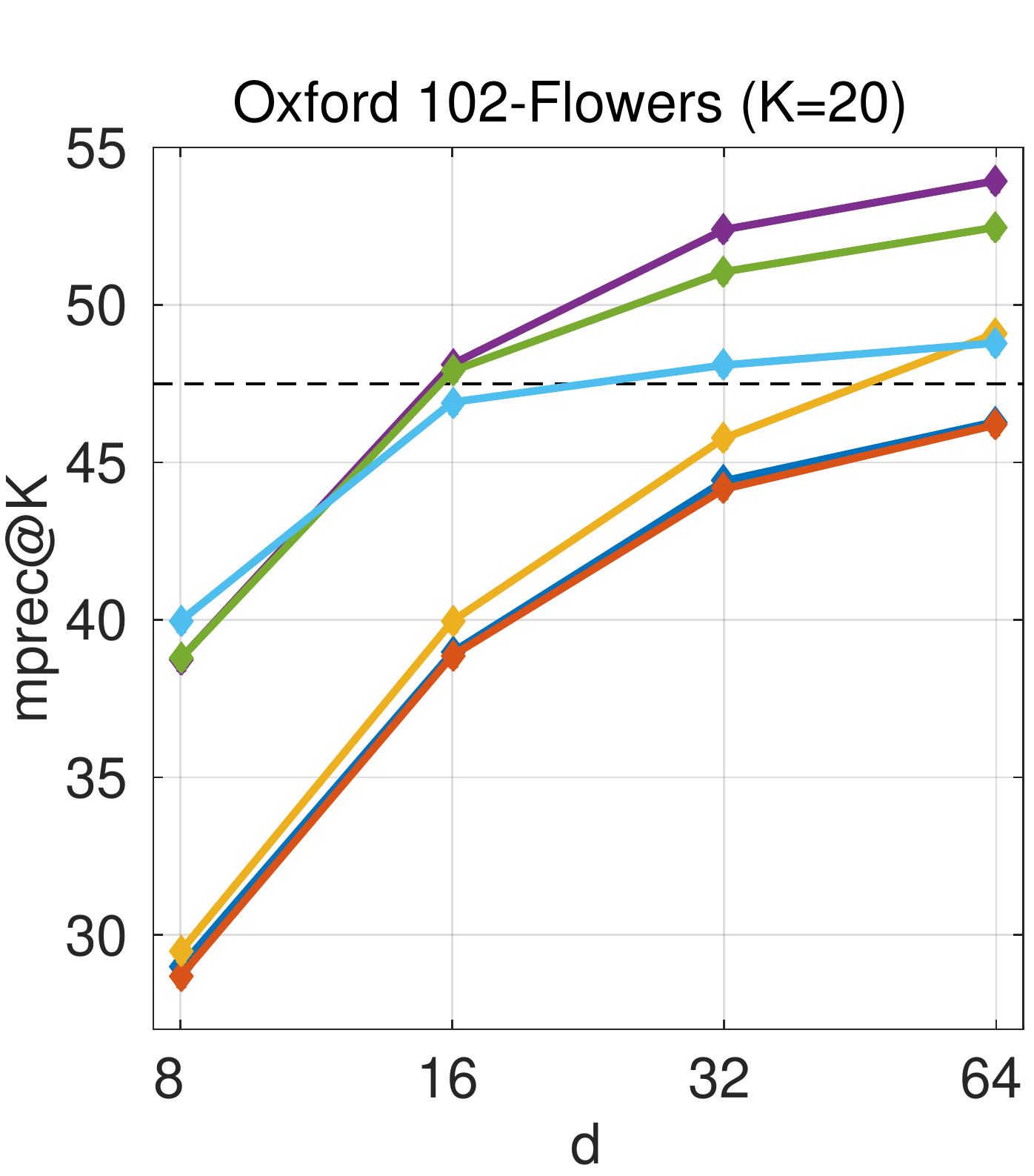}   
\includegraphics[width=0.245\textwidth,trim=20 180 190 155,clip]{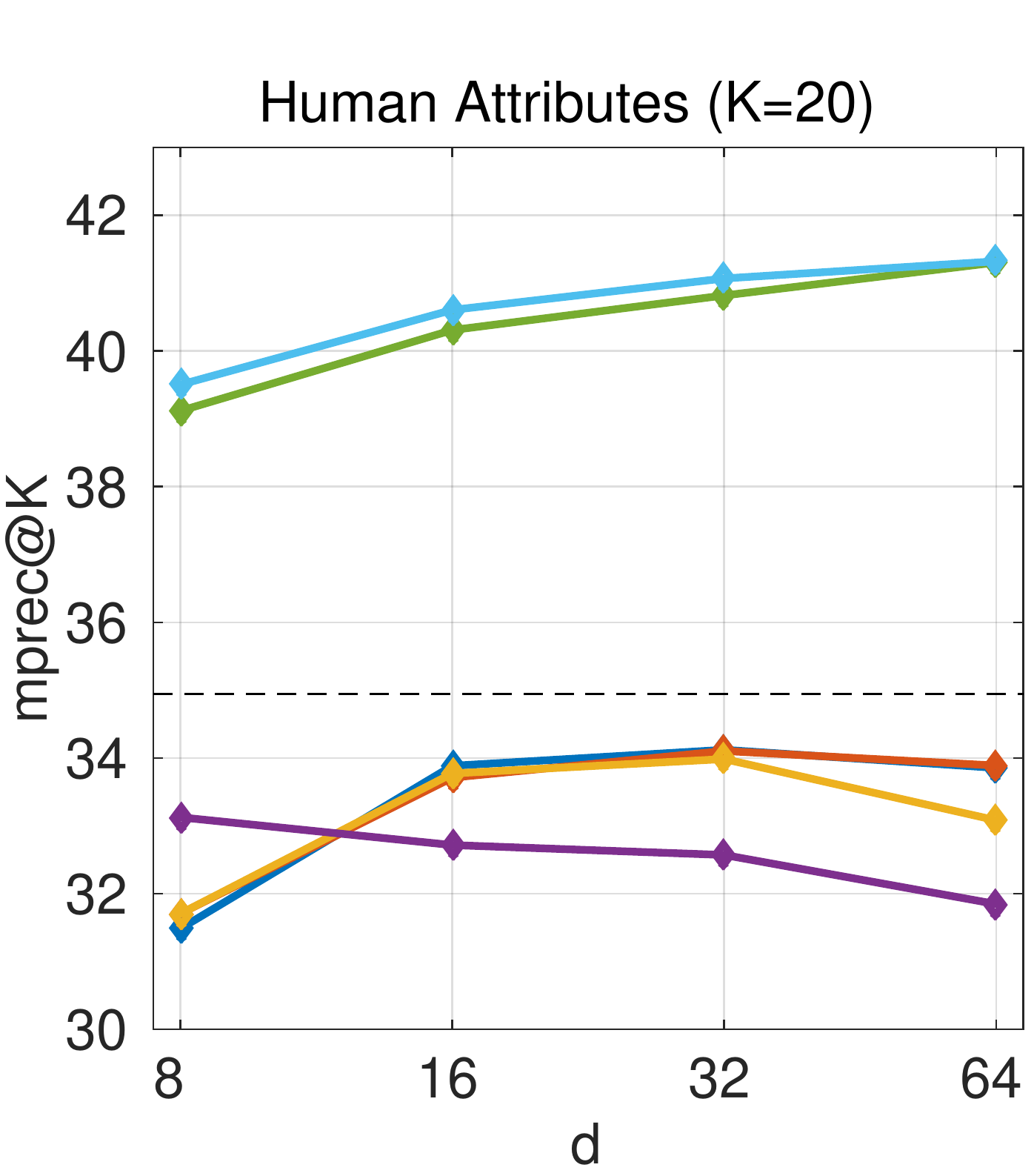}   
\includegraphics[width=0.335\textwidth,trim=20 180  50 155,clip]{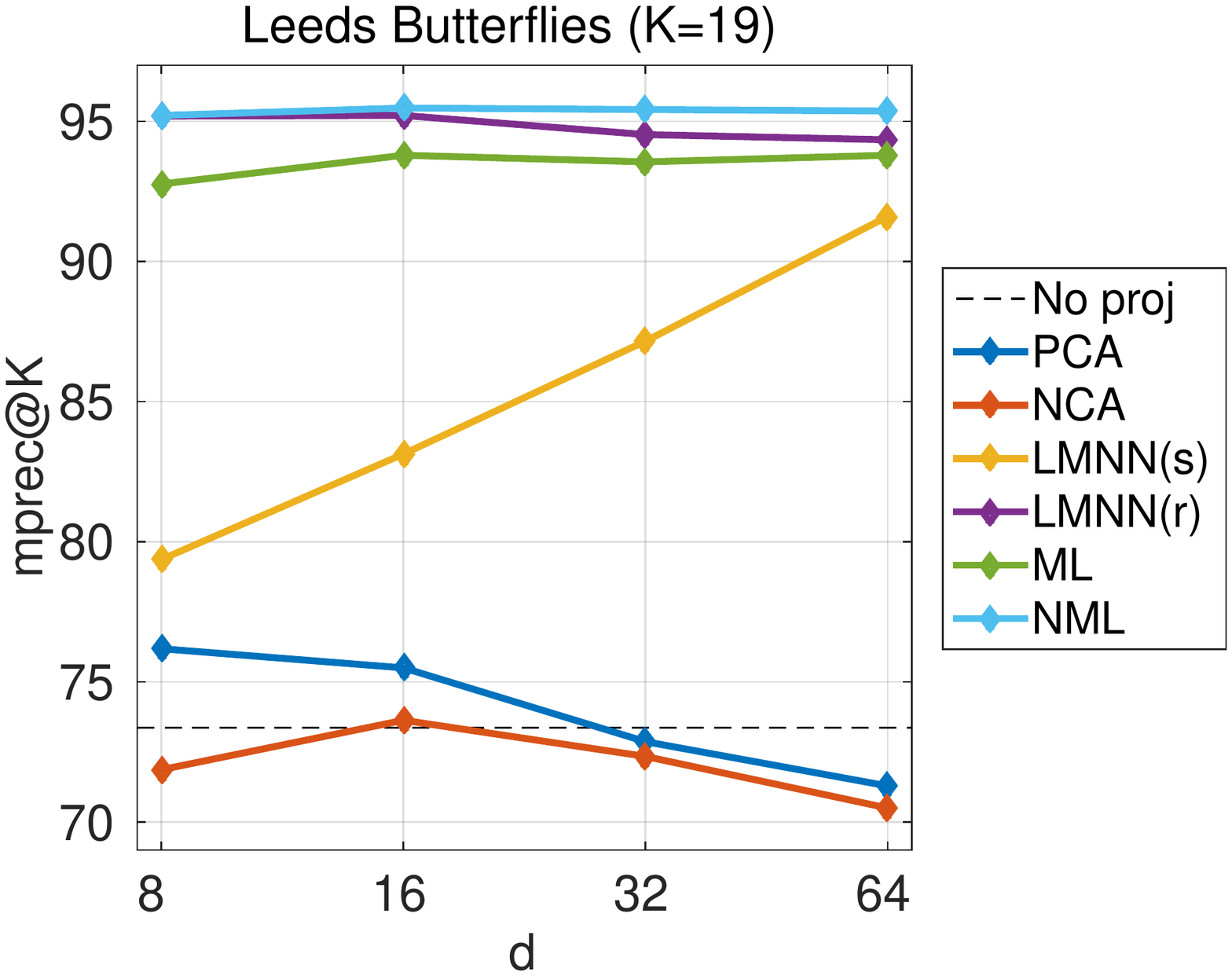}   
\caption{
Comparisons of methods on the different datasets for $K=20$ (19, for datasets with only 20 positive images
per class) and $d \in \{8,16,32,64\}$.
}
\label{supp_fig:prec_d_k20}
\end{figure*}

Fig.~\ref{supp_fig:prec_k_d8}--\ref{supp_fig:prec_k_d64} show the performance vs.\ $K$ for different $d\in
\{8,16,32,64\}$ and 
Fig.~\ref{supp_fig:prec_d_k1}--\ref{supp_fig:prec_d_k20} show the performance of the
method vs.\ the projection dimension $d$ for different values of top retrieved examples $K \in
\{1,10,20,30\}$.
The results support the discussion in the paper -- the proposed method (NML) improves the
performance over the baselines in most of the datasets for most of the settings.  Otherwise, it is
as good as the best baseline and only in very few cases, is worse than any baseline.

\begin{figure*}
\centering
\includegraphics[width=0.245\textwidth,trim=20 180 190 155,clip]{prec_k_fmd_d16.pdf}   
\includegraphics[width=0.245\textwidth,trim=20 180 190 155,clip]{prec_k_cub_d16.pdf}   
\includegraphics[width=0.245\textwidth,trim=20 180 190 155,clip]{prec_k_voc07_d16.pdf}   
\includegraphics[width=0.245\textwidth,trim=20 180 190 155,clip]{prec_k_scene15_d16.pdf}   
\includegraphics[width=0.245\textwidth,trim=20 180 190 155,clip]{prec_k_flowers_d16.pdf}   
\includegraphics[width=0.245\textwidth,trim=20 180 190 155,clip]{prec_k_hatdb_d16.pdf}   
\includegraphics[width=0.335\textwidth,trim=20 180  50 155,clip]{prec_k_butterfly_d16.pdf}   
\caption{
Comparisons of methods on the different datasets for $K \in [1,\max(50,n^+-1)]$ and $d=16$.
}
\label{supp_fig:prec_k_d16}
\end{figure*}

\begin{figure*}
\centering
\includegraphics[width=0.245\textwidth,trim=20 180 190 155,clip]{prec_d_fmd_k10.pdf}   
\includegraphics[width=0.245\textwidth,trim=20 180 190 155,clip]{prec_d_cub_k10.pdf}   
\includegraphics[width=0.245\textwidth,trim=20 180 190 155,clip]{prec_d_voc07_k10.pdf}   
\includegraphics[width=0.245\textwidth,trim=20 180 190 155,clip]{prec_d_scene15_k10.pdf}   
\includegraphics[width=0.245\textwidth,trim=20 180 190 155,clip]{prec_d_flowers_k10.pdf}   
\includegraphics[width=0.245\textwidth,trim=20 180 190 155,clip]{prec_d_hatdb_k10.pdf}   
\includegraphics[width=0.335\textwidth,trim=20 180  50 155,clip]{prec_d_butterfly_k10.pdf}   
\caption{
Comparisons of methods on the different datasets for $K=10$ and $d \in \{8,16,32,64\}$.
}
\label{supp_fig:prec_d_k10}
\end{figure*}

\begin{figure*}
\centering
\includegraphics[width=0.245\textwidth,trim=20 180 190 155,clip]{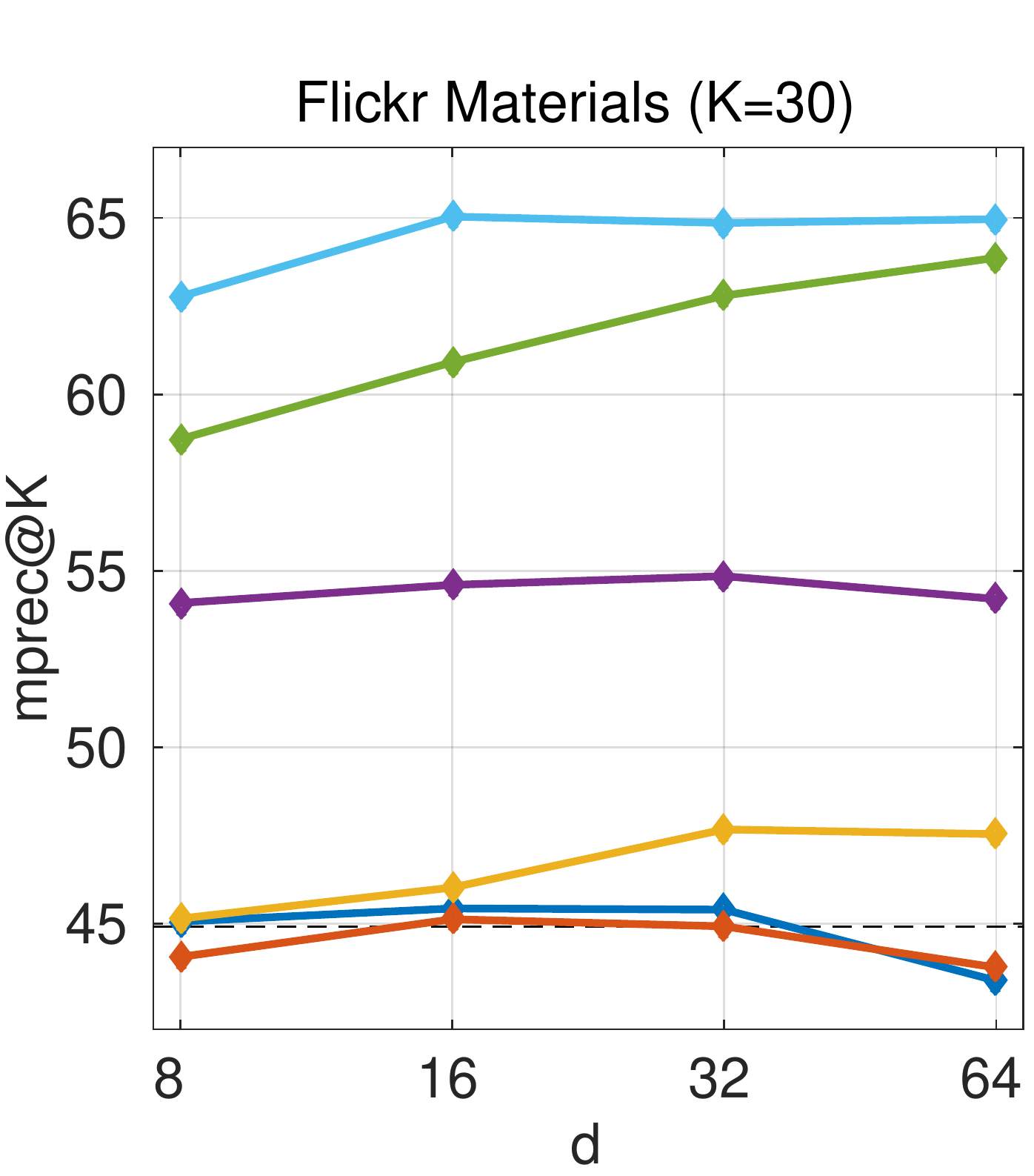}   
\includegraphics[width=0.245\textwidth,trim=20 180 190 155,clip]{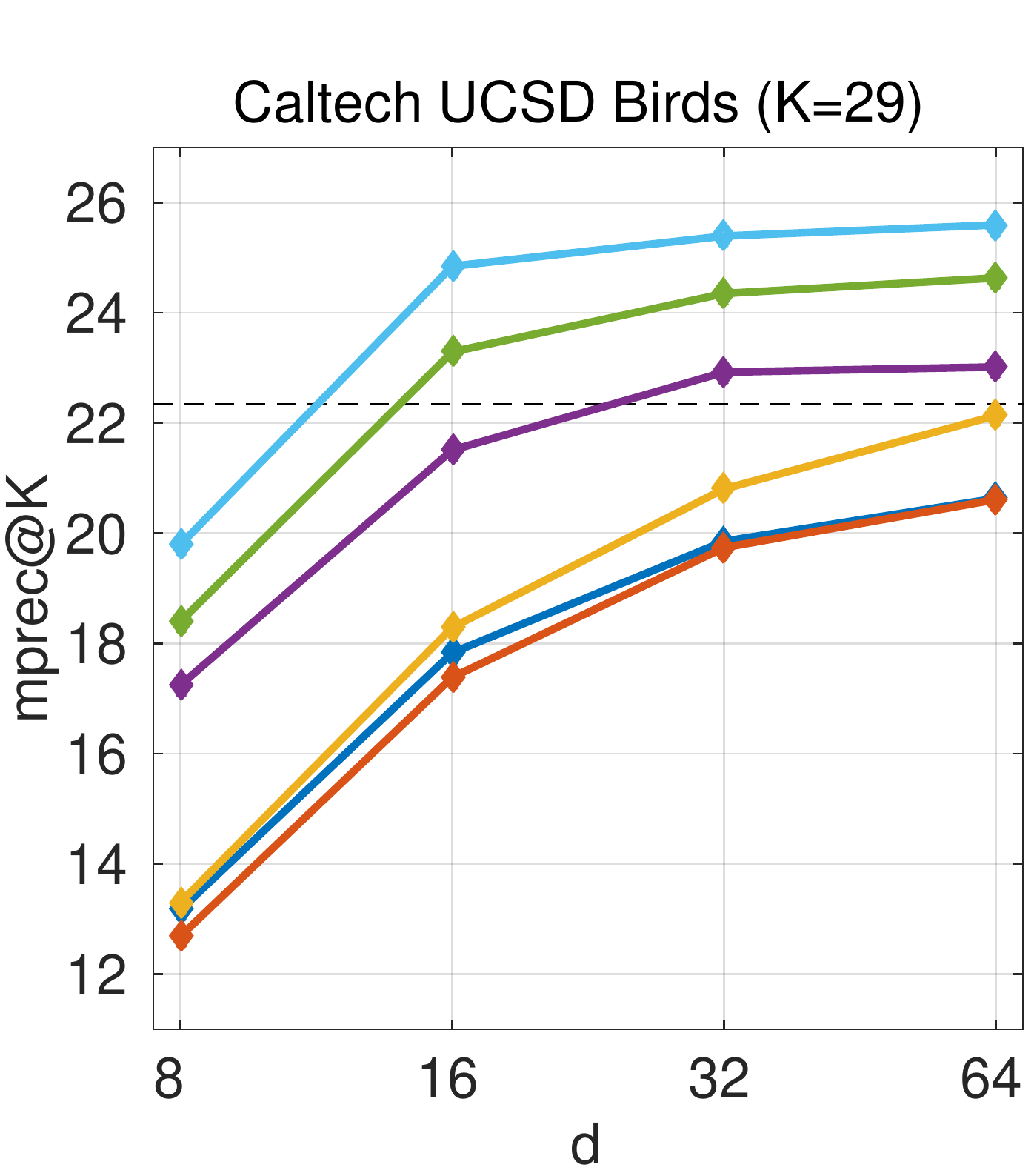}   
\includegraphics[width=0.245\textwidth,trim=20 180 190 155,clip]{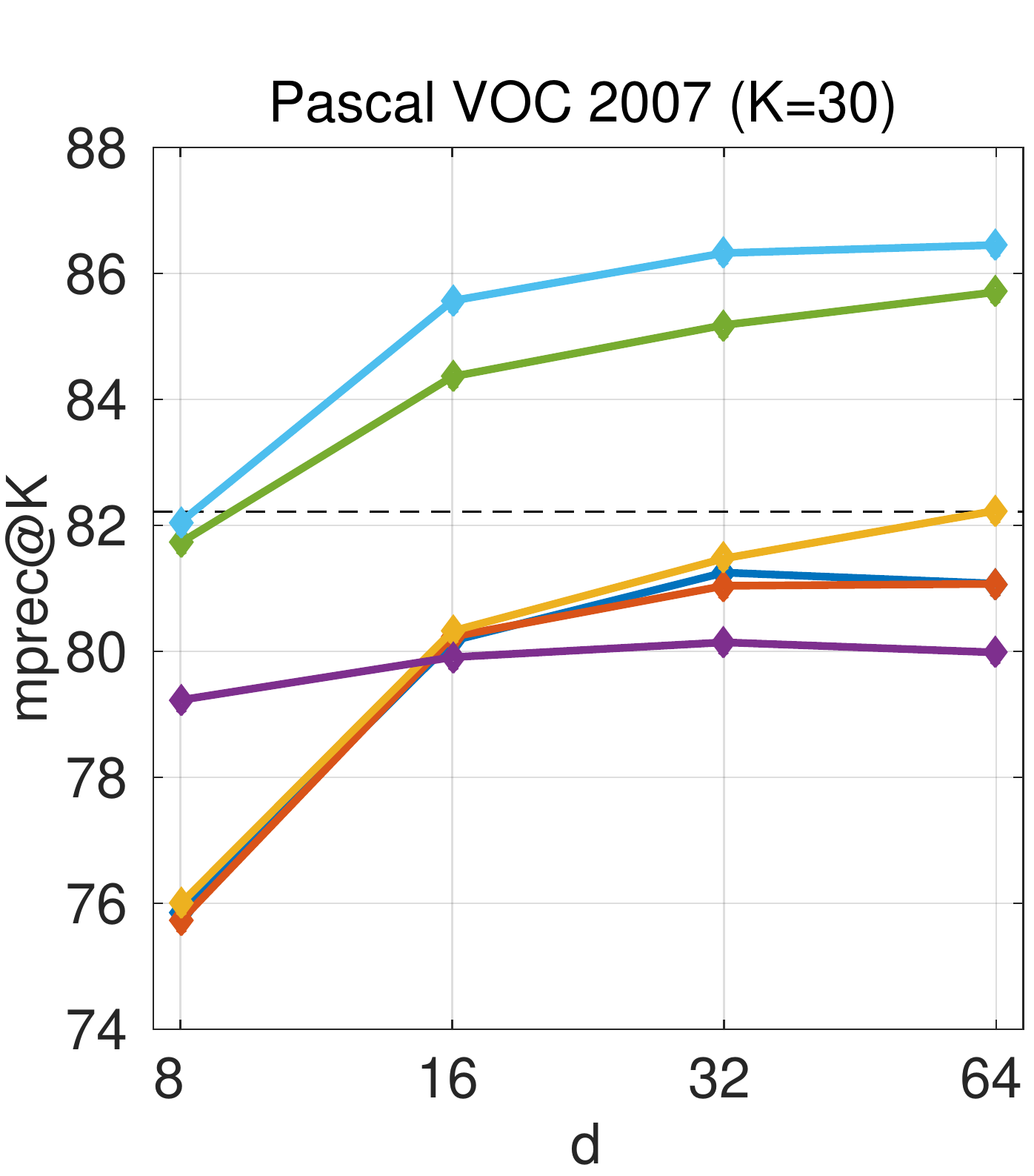}   
\includegraphics[width=0.245\textwidth,trim=20 180 190 155,clip]{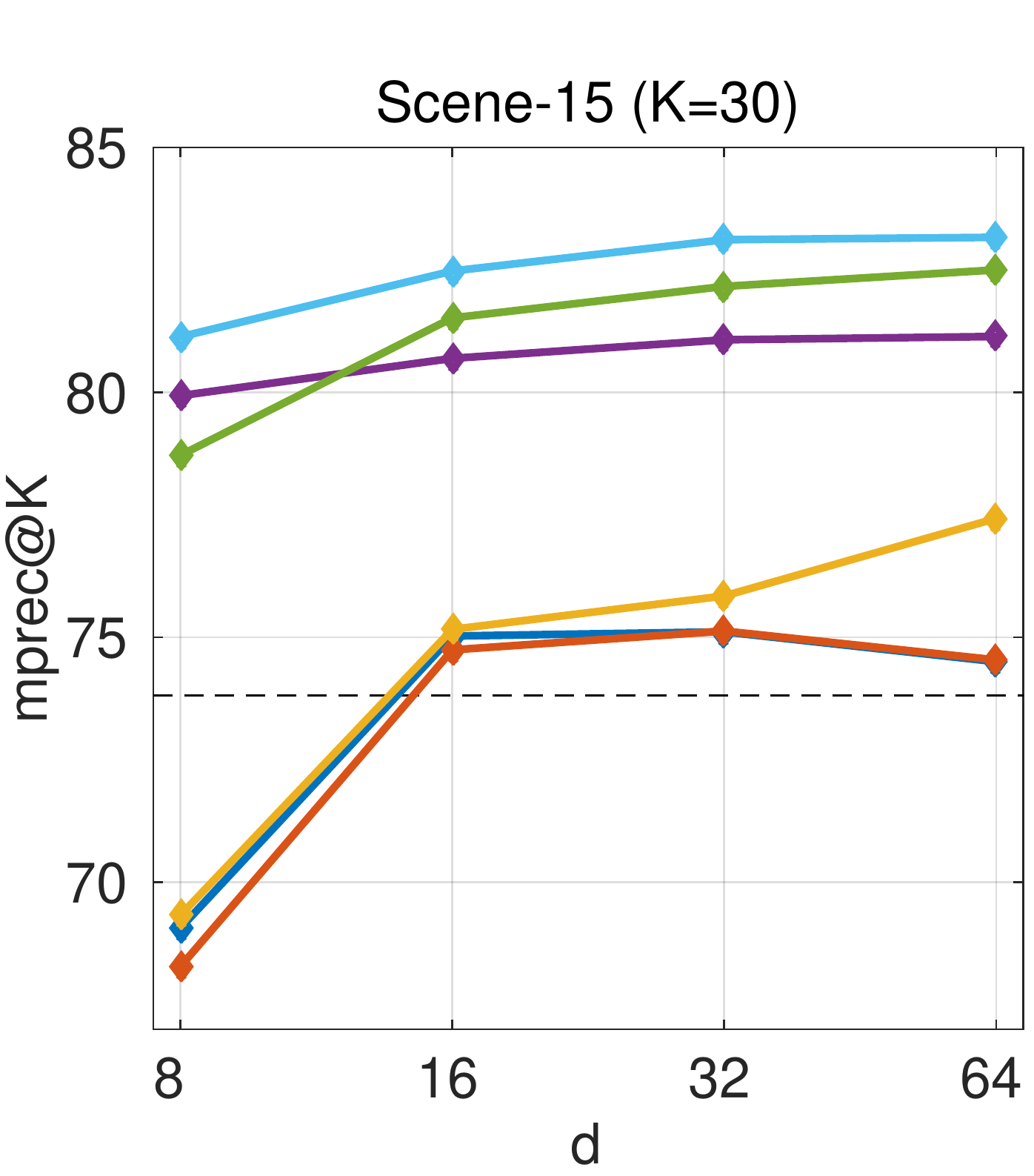}   
\includegraphics[width=0.245\textwidth,trim=20 180 190 155,clip]{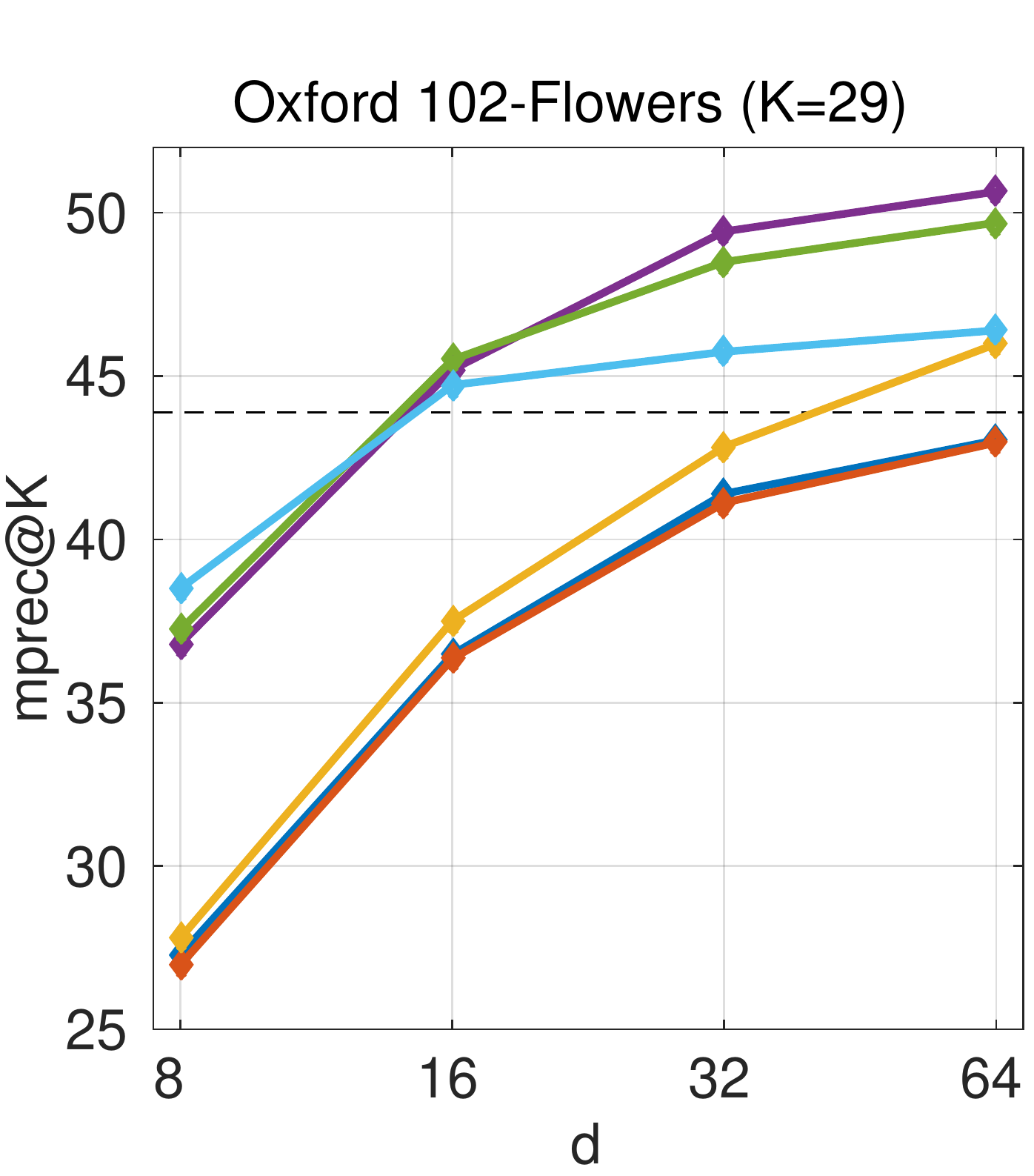}   
\includegraphics[width=0.335\textwidth,trim=20 180  50 155,clip]{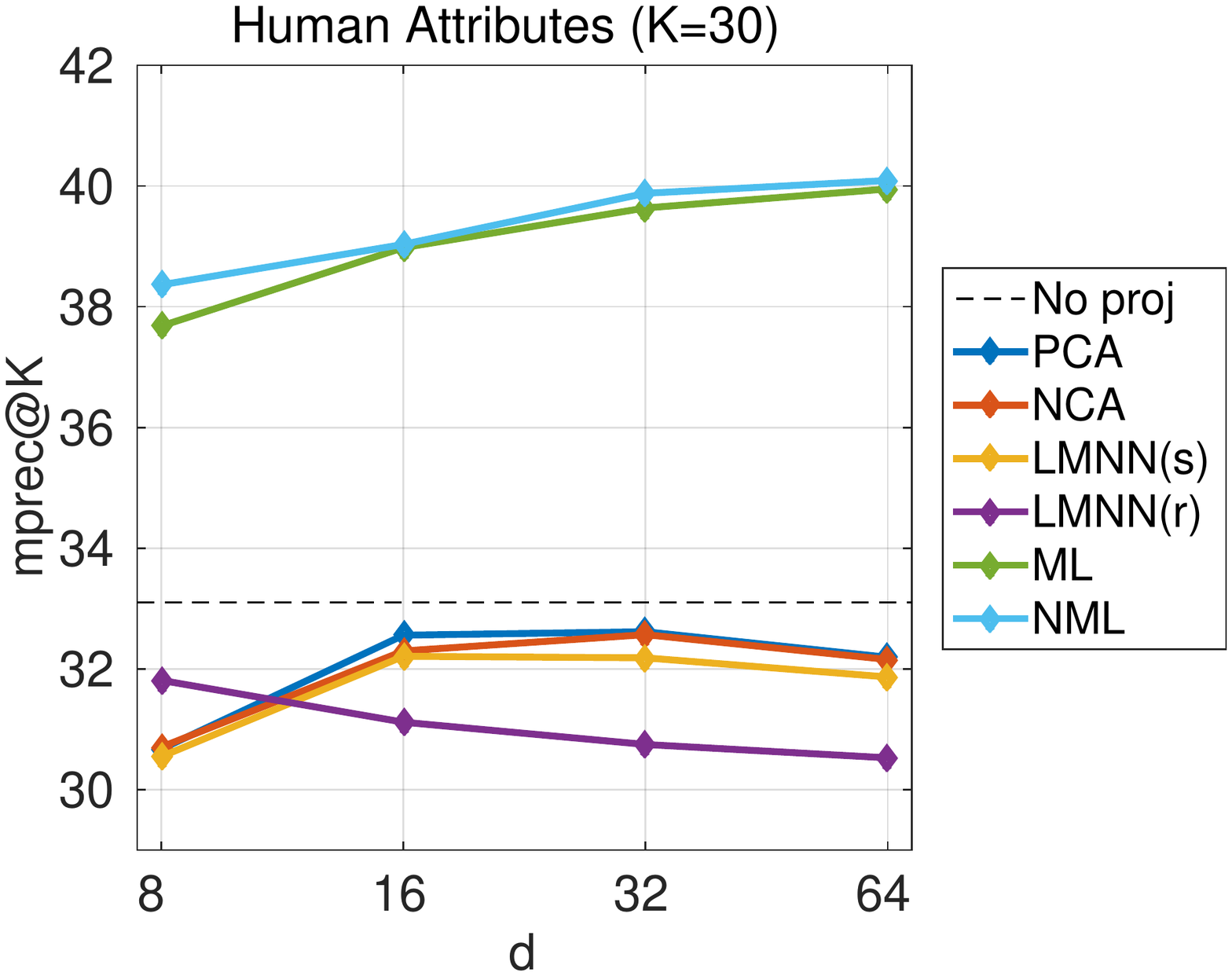}   
\caption{
Comparisons of methods on the different datasets for $K=30$ (29, for datasets with only 30 positive images
per class) and $d \in \{8,16,32,64\}$.
}
\label{supp_fig:prec_d_k30}
\end{figure*}

\end{document}